\documentclass[sigconf]{acmart}

\usepackage{bm}
\usepackage{xcolor}
\colorlet{RED}{red}
\usepackage{soul}
\usepackage{flafter}
\usepackage{float}

\AtBeginDocument{%
  \providecommand\BibTeX{{%
    \normalfont B\kern-0.5em{\scshape i\kern-0.25em b}\kern-0.8em\TeX}}}

\copyrightyear{2022} 
\acmYear{2022} 
\setcopyright{rightsretained} 
\acmConference[CHI '22]{CHI Conference on Human Factors in Computing Systems}{April 29-May 5, 2022}{New Orleans, LA, USA}
\acmBooktitle{CHI Conference on Human Factors in Computing Systems (CHI '22), April 29-May 5, 2022, New Orleans, LA, USA}
\acmDOI{10.1145/3491102.3517522}
\acmISBN{978-1-4503-9157-3/22/04}

\acmSubmissionID{9355}

\begin{document}

%%
%% The "title" command has an optional parameter,
%% allowing the author to define a "short title" to be used in page headers.
% \title{Debiased-CAM for bias-agnostic faithful visual explanations of machine learning}
% \title{Debiased-CAM to mitigate systematic error with faithful visual explanations of machine learning}
\title{Debiased-CAM to mitigate image perturbations with\\
faithful visual explanations of machine learning}

%%
%% The "author" command and its associated commands are used to define
%% the authors and their affiliations.
%% Of note is the shared affiliation of the first two authors, and the
%% "authornote" and "authornotemark" commands
%% used to denote shared contribution to the research.
\author{Wencan Zhang}
\affiliation{%
  \institution{National University of Singapore}
  \streetaddress{COM1, 13, Computing Dr}
  \city{}
  \country{Singapore}
  \postcode{117417}
}
\email{wencanz@u.nus.edu}

\author{Mariella Dimiccoli}
\affiliation{%
  \institution{Institut de Robòtica i Informàtica Industrial, CSIC-UPC}
  \streetaddress{Carrer de Llorens i Artigas, 4}
  \city{Barcelona}
%   \state{State}
  \country{Spain}
  \postcode{08028}
}
\email{maria.dimiccoli@upc.edu}

\author{Brian Y. Lim}
\affiliation{%
  \institution{National University of Singapore}
  \streetaddress{COM1, 13, Computing Dr}
  \city{}
  \country{Singapore}
  \postcode{117417}
}
\email{brianlim@comp.nus.edu.sg}

%%
%% By default, the full list of authors will be used in the page
%% headers. Often, this list is too long, and will overlap
%% other information printed in the page headers. This command allows
%% the author to define a more concise list
%% of authors' names for this purpose.
% \renewcommand{\shortauthors}{First Author’s Name, Initials, and Last Name}

%%
%% The abstract is a short summary of the work to be presented in the
%% article.
\begin{abstract}
Model explanations such as saliency maps can improve user trust in AI by highlighting important features for a prediction.
However, these become distorted and misleading when explaining predictions of images that are subject to systematic error (bias) by perturbations and corruptions. Furthermore, the distortions persist despite model fine-tuning on images biased by different factors (blur, color temperature, day/night).
We present Debiased-CAM to recover explanation faithfulness across various bias types and levels by training a multi-input, multi-task model with auxiliary tasks for explanation and bias level predictions.
In simulation studies, the approach not only enhanced prediction accuracy, but also generated highly faithful explanations about these predictions as if the images were unbiased. 
In user studies, debiased explanations improved user task performance, perceived truthfulness and perceived helpfulness. 
Debiased training can provide a versatile platform for robust performance and explanation faithfulness for a wide range of applications with data biases.
\end{abstract}

%%
%% The code below is generated by the tool at http://dl.acm.org/ccs.cfm.
%% Please copy and paste the code instead of the example below.
%%
\begin{CCSXML}
<ccs2012>
   <concept>
       <concept_id>10003120.10003121.10011748</concept_id>
       <concept_desc>Human-centered computing~Empirical studies in HCI</concept_desc>
       <concept_significance>500</concept_significance>
       </concept>
   <concept>
       <concept_id>10010147.10010178.10010224</concept_id>
       <concept_desc>Computing methodologies~Computer vision</concept_desc>
       <concept_significance>500</concept_significance>
       </concept>
   <concept>
       <concept_id>10010147.10010257.10010282.10011305</concept_id>
       <concept_desc>Computing methodologies~Semi-supervised learning settings</concept_desc>
       <concept_significance>300</concept_significance>
       </concept>
   <concept>
       <concept_id>10002978.10003029.10011150</concept_id>
       <concept_desc>Security and privacy~Privacy protections</concept_desc>
       <concept_significance>300</concept_significance>
       </concept>
 </ccs2012>
\end{CCSXML}

\ccsdesc[500]{Human-centered computing~Empirical studies in HCI}
\ccsdesc[500]{Computing methodologies~Computer vision}
\ccsdesc[300]{Computing methodologies~Semi-supervised learning settings}
\ccsdesc[300]{Security and privacy~Privacy protections}

%%
%% Keywords. The author(s) should pick words that accurately describe
%% the work being presented. Separate the keywords with commas.
\keywords{Explainable AI, Misleading explanations, Class activation map, Robust machine learning, Image perturbations, User studies}

%%
%% This command processes the author and affiliation and title
%% information and builds the first part of the formatted document.
\maketitle

\section{Introduction}

\begin{figure*}[!t]
    \centering
    \hspace*{-0.6cm}
    \includegraphics[width=14.0cm]{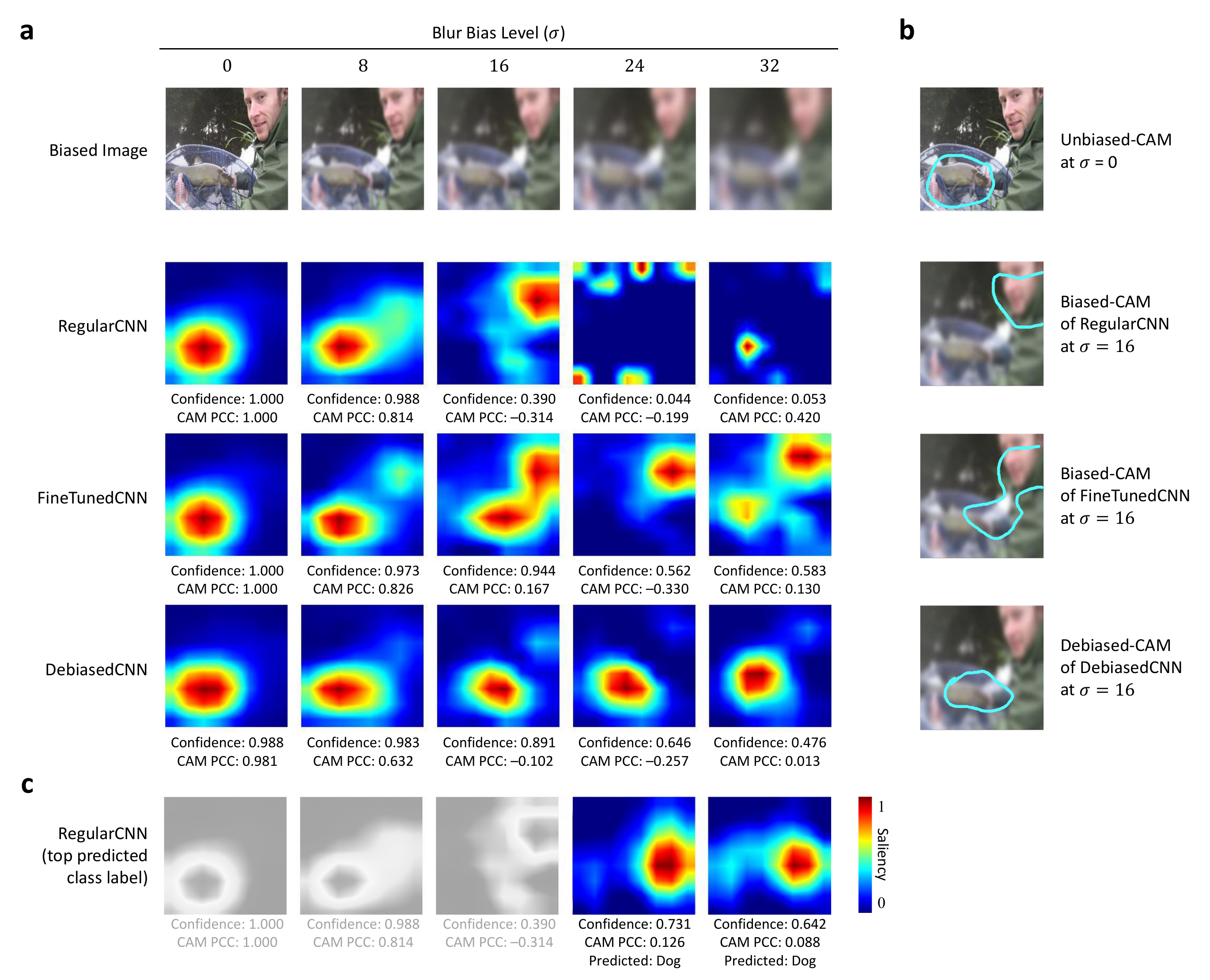}
    \hspace*{-0.5cm}
    \vspace{-0.3cm}
    \caption{
    Deviated and debiased CAM explanations for prediction label "Fish". 
    a) Debiased-CAMs (from DebiasedCNN) were most faithful to the Unbiased-CAM (from RegularCNN at $\sigma=0$) as blur bias increased. In contrast, Biased-CAMs from RegularCNN and FineTunedCNN became very deviated with a much lower CAM Pearson Correlation Coefficient (PCC). 
    The wrong CAMs can mislead users to think the predictions were wrong even if they were correct.
    b) Debiased-CAM selected similar important pixels of the Fish as Unbiased-CAM, while Biased-CAMs selected irrelevant pixels of the person or background instead. 
    % Important pixels shown within contour lines (cyan) were overlaid  on the actual unblurred scene ($\sigma=0$) for reference and the blurred input image at $\sigma=16$. 
    c) CAM of the top predicted class label with only RegularCNN at $\sigma=32$ predicting the wrong prediction label “Dog”.  
    }
    \vspace{-0.3cm}
    \label{fig:biasedCAMs}
    \Description{
    Demos of images and different types of GradCAMs.
    a) The 1st row shows the same photo blurred with 5 different levels (sigma = 0, 8, 16, 24, 32), where a man is holding a fish. The 2nd row shows the GradCAM visual explanations of RegularCNN on corresponding blurred images. The red region indicates a higher importance while the blue region indicates a lower importance. The text below the image indicates the confidence score of the image instance and the CAM truthfulness score (via PCC). The 3rd row and 4th row indicate the GradCAM visual explanations of FinetunedCNN and DebiasedCNN respectively.
    b) The 1st figure shows the same unblurred image instance, and the cyan contour lines marked the salient region according to the unbiasedCAM. The following 3 figures show the blurred image (sigma = 16) and corresponding cyan contour lines based on BiasedCAM from RegularCNN, FinetunedCNN and DebiasedCAM from DebiasedCNN.
    c) Similiar to the 2nd row in subfigure a). Only the last 2 CAM figures differ since the classifier's prediction labels are changed from fish to dog.
    }
    
\end{figure*}

Machine learning models are increasingly capable to achieve impressive performance in many prediction tasks, such as image recognition \cite{krizhevsky2012imagenet}, medical image diagnosis \cite{esteva2017dermatologist}, captioning \cite{vinyals2015show} and dialog systems \cite{das2017visual}. 
% Their prevalence in society requires responsible use, particularly in supporting decision explainability. Specifically, the need to improve user trust and understanding \cite{lipton2018mythos, miller2019explanation, wang2019designing} has propelled the development of a wide variety of explainable artificial intelligence (XAI) and interpretable machine learning methods \cite{guidotti2018survey, hohman2018visual, wang2019designing, zhang2018visual}. 
% {\color{orange} In practical applications, data usually suffers from corruption or biasing due to deliberate or unintentional reasons. For example, obfuscation has been proved an effective privacy-preservation scheme for both images shared on social media platform \cite{padilla2015visual,hasan2018viewer} and photos captured by wearable cameras \cite{ dimiccoli2018mitigating, alharbi2019mask}. On the other hand, images may also be biased with shifted color temperature \cite{afifi2019else} due to mis-set white balance. Prior works have shown that biased data leads to poor model performance \cite {dimiccoli2018mitigating, vasiljevic2016examining}, and finetuning model on biased data can mitigate performance drop. However, none of them investigated the influence of biased data on model's explanation.}
Despite their superior performance, deep learning models are complex and unintelligible; this limits user trust and understanding \cite{lipton2018mythos, miller2019explanation, wang2019designing}. This has driven the development of myriad explainable artificial intelligence (XAI) and interpretable machine learning methods \cite{guidotti2018survey, hohman2018visual, wang2019designing, zhang2018visual}. 
Saliency maps \cite{selvaraju2017grad, Simonyan15, zhou2016learning} can provide intuitive explanations of Convolutional Neural Networks (CNN) for image prediction tasks by indicating which pixels or neurons were used for model inference. Amongst these, class activation map (CAM) \cite{zhou2016learning}, Grad-CAM \cite{selvaraju2017grad} and extensions \cite{chattopadhay2018grad, wang2020score} are particularly useful by identifying pixels relevant to specific class labels. Users can verify the prediction correctness by checking whether expected pixels are highlighted. Models would be considered more trustworthy if their CAMs matched what users believe as salient. 

% {\color{orange} From a preliminary experiment, we observed that biased images tended to produced deviated, less faithful explanations that are misaligned with people’s expectations \cite{posner2004attention}, which impedes human verification and trust \cite{du2019techniques} of the model prediction.} For example, saliency maps \cite{selvaraju2017grad, Simonyan15, zhou2016learning}, which explain image task predictions may highlight irrelevant targets and mislead users to interpret that the model is focusing on the wrong concept. (Fig. \ref{fig:biasedCAMs}a (Biased-CAMs) demonstrates how explanations deviate more with increasing image bias.)

Despite the fidelity of CAMs on clean images, real-world images are typically subjected to systematic error, such as image blurring, color-distortion or lighting changes, which can affect what CAMs highlight. 
We call this systematic error \textit{bias}\footnote{Note that this does not refer to \textit{social bias} that is presently popularly studied in AI fairness and algorithmic bias. We are using the word as defined in engineering and physics regarding measurements.} since it is directional based on a contextual factor or confound, and contrast it with \textit{noise} that is based on non-directional random error.
Also note that we are not referring to societal bias or discrimination (e.g., racism, sexism)~\cite{dodge2019explaining}.
Blurring can be due to accidental motion \cite{kupyn2018deblurgan} or defocus blur \cite{vasiljevic2016examining}, or deliberate obfuscation for privacy protection \cite{dimiccoli2018mitigating}. 
Unlike ~\cite{zhao2021exploiting} which found explanation harms privacy, we find that privacy can harm explanation.
Images may also be biased with shifted color temperature \cite{afifi2019else} due to mis-set white balance, or 
biased with daylight changes (e.g., day to night, sunrise/sunset). 
These biases decrease model prediction performance \cite{afifi2019else, dimiccoli2018mitigating, vasiljevic2016examining} and we further show that they also lead to deviated or distorted CAM explanations that are less faithful to the original scenes. For different bias types (image blur, and color temperature shift, day/night lighting), we found that CAMs deviated more as image bias increased (Fig. \ref{fig:biasedCAMs} and Fig. \ref{fig:biasedCAMsAcrossDatasets}: Biased-CAMs from RegularCNN for $\sigma$ > 0). Although Biased-CAM represents what the CNN considers important in a biased image, 
% {\color{blue}the deviation in CAM explanation reflects the instability of model's output given imperfect data~\cite{Zheng2016ImprovingTR}, which impedes the model's robustness in practical scenarios.} Moreover, Biased-CAM
it is misaligned with people’s expectations \cite{posner2004attention}, misleads users to irrelevant targets, and impedes human verification and trust \cite{du2019techniques} of the model prediction. For example, when explaining the inference of the “Fish” label for an image prediction, Biased-CAMs select pixels of the man instead of the fish (Fig. \ref{fig:biasedCAMs}). 

To align with user expectations, models should not only have the right predictions but also have the right reasons \cite{ross2017right}; however, current approaches face challenges in achieving this goal, particularly for biased data. First, while fine-tuning the model on biased data can improve its performance \cite{dimiccoli2018mitigating, vasiljevic2016examining}, this does not necessarily produce explanations aligned with human's understanding. Indeed, we found that explanations remain deviated and unfaithful (Fig. \ref{fig:biasedCAMs}, FineTunedCNN Biased-CAMs). 
% {\color{orange} Conversely, attention transfer \cite{komodakis2017paying, li2018tell} helps calibrate model's explanation faithfulness on image related tasks, while existing methods are mainly designed to handle clean data instead of biased data.} 
Conversely, retraining the model with attention transfer~\cite{komodakis2017paying, li2018tell} only improves explanation faithfulness for clean images, but cannot handle biased images.
Finally, evaluating the human interpretability of explanations requires deep inquiry into user perception, understanding and usage \cite{abdul2018trends, alqaraawi2020evaluating, doshi2017towards}, but typical evaluations of XAI involve only data simulations \cite{bach2015pixel, fong2017interpretable, ross2017right, sundararajan2017axiomatic, zhang2018interpretable} or simple surveys \cite{bau2017network, chattopadhay2018grad, schramowski2020making, selvaraju2017grad, zhou2018interpretable}. 

Inspired by how people can “see through the blur” to recognize blurred images due to prior experiences with unblurred but unrelated images, we propose a debiasing approach such that models are trained to faithfully explain the event despite biased sources. Using CNNs with Grad-CAM saliency map explanations \cite{selvaraju2017grad}, we developed DebiasedCNN that interprets biased images as if predicting on the unbiased form of images and produces explanations, Debiased-CAMs, that are more human-relatable and robust. The approach has a modular design: 
1) it is self-supervised which does not require additional human annotation for training; 
% {\color {orange} 2) it learns explanations in a secondary prediction task via a differentiable loss;} 
2) it produces explanations as a secondary prediction task, so that they are retraininable to be debiased;
3) it models the bias level as a tertiary task to support bias-aware predictions. 
The approach not only enhances prediction performance on biased data, 
% {\color{orange}but also generates highly faithful explanations aligned with user's understanding regarding the semantics in image} as if the data were unbiased
but also produces highly faithful explanations about these predictions as if the data were unbiased 
(Fig. \ref{fig:biasedCAMs}: DebiasedCNN CAMs).

To evaluate the developed model, we conducted simulation and user studies to address the research questions on 
1) how bias decreases explanation faithfulness and how well debiasing mitigates this, and 
2) how sensitive people are to perceiving explanation deviations and how well debiasing improves perceived explanation truthfulness and helpfulness. For generality, the simulation studies spanned different image prediction tasks object recognition, activity recognition with egocentric cameras, image captioning, and scene understanding), bias types (blur, color shift, and night vision interpolation) and various datasets. Across all studies, we found that while increasing bias led to poorer prediction performance and worse explanation deviation, Debiased-CAM showed the best improvement in task performance as well as explanation faithfulness. Instead of trading off task performance for explanation faithfulness, our debiasing training improved both. We further demonstrated the usability and usefulness of Debiased-CAMs in two controlled user studies. Quantitative statistical and qualitative thematic analyses validated that users can perceive the improved truthfulness and helpfulness of Debiased-CAMs on biased images. 
In summary, this paper made the following \textbf{contributions}:
\begin{enumerate}
    \item Assessed the deviations in model explanations due to bias in data across different bias types and levels.
    \item Proposed a technical approach to accurately predict and faithfully explain inferences under data bias.
    \item Validated the improvements in perceived truthfulness and helpfulness of debiased explanations.    
\end{enumerate}
 
\section{Related Work}
We review explainable AI methods for image predictions, how images get biased, how misleading explanations harm user experience and performance, and methods to improve explanation faithfulness.

\subsection{Explainable AI for visual CNN models} 
Many explainable AI (XAI) techniques have been proposed to understand the predictions of CNNs. These include 
saliency maps \cite{chattopadhay2018grad,ramaswamy2020ablation,selvaraju2017grad,simonyan2014deep,zhou2016learning}, 
feature visualization \cite{bau2017network,hohman2019s,olah2017feature}, neuron activations \cite{kahng2017cti} and 
concept variables \cite{kim2018interpretability,koh2020concept}. 
Saliency maps are intuitive to interpret deep CNN models, where important pixels are highlighted to indicate their importance towards the model prediction. 
Computing the prediction gradient \cite{simonyan2014deep, sundararajan2017axiomatic, zeiler2014visualizing} can identify sensitive pixels. 
Another approach divides prediction outcome across features by Taylor series approximation \cite{bach2015pixel} or Shapley values \cite{lundberg2017unified}. Specific to CNNs, coarser saliency maps can be generated by aggregating activation maps as a weighted sum across convolutional kernels \cite{chattopadhay2018grad, ramaswamy2020ablation, selvaraju2017grad, wang2020score, zhou2016learning}. 
% While CAM \cite{zhou2016learning} requires model retraining, Grad-CAM \cite{selvaraju2017grad} can be read from any CNN without retraining. 
% Extensions improved Grad-CAM for robustness \cite{ramaswamy2020ablation} and multiple objects \cite{chattopadhay2018grad}. 
For this work, we evaluated Grad-CAM \cite{selvaraju2017grad} to test if users can perceive truthful, biased, or debiased explanations, and expect our findings to be generalizable. 
% generalize to its extensions~\cite{chattopadhay2018grad,ramaswamy2020ablation}.
% CAM saliency maps are intuitive for lay end-users to interpret with basic perception skills.
% Hence, we chose Grad-CAM for testing if users can perceive truthful, biased, or debiased explanations.

\subsection{Systematic error and corruptions in images} 
Although many models are trained on clean curated images, real-world images are subject to systematic errors (biases), perturbations and corruptions.
Contextual or incidental biases include blurring, color distortions, or lighting changes.
Blurring may be due to accidental motion blur \cite{kupyn2018deblurgan}, defocus blur \cite{vasiljevic2016examining} or deliberate obfuscation for privacy protection \cite{dimiccoli2018mitigating}. 
Image color shift \cite{afifi2019else} may be due to mis-set white balance. 
These biases can degrade model performance \cite{afifi2019else, dimiccoli2018mitigating, vasiljevic2016examining}, and limit their usefulness in real-world applications.
Images of outdoor scenes regularly change by time of day and seasons due to sunlight or weather changes~\cite{laffont2014transient}.
Images can also be corrupted due to data processing, such as JPEG compression artifacts, Gaussian noise, brightness or contrast levels~\cite{hendrycks2019benchmarking}. 
% However, fewer applications are likely to use such images of low quality.
Mitigation strategies to handle such data errors include model fine-tuning with images at known blurred levels \cite{dimiccoli2018mitigating}, or data augmentation with images blurred at multiple levels \cite{vasiljevic2016examining}.
However, these approaches only aimed to improve prediction performance and not explanation faithfulness.
In this work, we found that explanations remain deviated and we propose methods to debias them.

Such data errors are related to the problem of model robustness, where small changes to data should not cause large changes in model behavior. This is an active area of research~\cite{Zheng2016ImprovingTR, hendrycks2019benchmarking, hendrycks2020augmix}, but methods typically focus on improving performance by increasing decision boundary smoothness. 
In this work, we aim to make explanations more robust.
Recent work by Dombrowski et al.~\cite{Dombrowski2022TowardsRE} improved explanation robustness by similarly increasing decision smoothness relative to explanations, but this assumes clean data, and learns average explanations under bias. Instead, we debias explanations away from deviations due to biased data.
Also, other than focusing on explanation robustness or stability towards the impression of global trustworthiness, we focus on faithful explanations that are verifiable per instance.

\subsection{Risk of misleading model explanations}

% Although explanation can help improve algorithm transparency and enhance user's trust on black box models, there is a growing trend in building human-understandable explanations that communicate knowledge with human and improve their decision making \cite{bansal2021does, lage2019evaluation, wang2021show}. However, recent studies have posed a risk that model's explanation can be erroneous when inputs are manipulated by adversarial noises \cite{dombrowski2019explanations, ghorbani2019interpretation} or contain uncertainty \cite{wang2021show}. Those erroneous explanation would conversely decrease user's trust on system \cite{de2020case, lakkaraju2020fool}, or mislead users with irrelevant information \cite{lakkaraju2020fool}. Therefore, robust model explanation with human-relatbale information to reflect intrinsic patterns of data \cite{hancox2020robustness} is a desirable property in read-world applications. In this paper, we mainly investigated the misleading effect from visual explanations in image-based CNN models, and proposed generalizable training framework for calibration.  
% }

User studies of model explanations aim to show that explanations can improve user understanding and trust~\cite{lim2009and, kizilcec2016much, nourani2019effects, zhang2020effect, wang2021explanations}.
% These studies tend to use cases where the model predicts correctly and explanations are pre-selected to be good. This measures the upper bound of how well explanations can help.
These tend to study scenarios of correct model predictions and ideal explanations, but
% In contrast, 
models can make prediction errors or may not be confident in their decisions. Studies have explored how this may lead to distrust, mistrust and over-trust~\cite{lim2011investigating,yin2019understanding, poursabzi2021manipulating}.
For such cases, explanations can be avoided when there is a high chance of model error.
However, explanations can still be wrong despite the model predicting correctly.
For example, explanations may highlight spurious pixels~\cite{zhang2020interpretable}, be adversarially manipulated~\cite{dombrowski2019explanations, ghorbani2019interpretation}, or subject to input error~\cite{wang2021show}.
These cases are harder to detect, pose a serious risk to decrease user trust~\cite{de2020case, lakkaraju2020fool}, or mislead users ~\cite{lakkaraju2020fool}.
Unlike works that explore how different explanation formats affect trust~\cite{yang2020visual}, we investigate how slight data variations affect user performance and trust.
% Note that the data bias we refer to is due to sensory perceptual error or statistical variance (e.g., noise, lighting), and not societal bias or discrimination (e.g., racism, sexism) investigated by Dodge et al.~\cite{dodge2019explaining}.
Since data bias and corruption are prevalent in the real-world, it is tantamount to identify the severity of the problem and mitigate it with more robust explanations~\cite{hancox2020robustness}.
In this work, we quantify the extent of explanation deviation due to data bias, and evaluated how sensitive users are to these deviations.

% \subsection{Fixing explanations with attention transfer} 
\subsection{Attention transfer to correct explanations} 
While explanation techniques are primarily designed to improve human understanding of model behavior, they can be used to guide model training. 
One approach is to use transfer learning to regularize attention from a better model to the model under training, such as with student-teacher networks \cite{komodakis2017paying}.
Another approach indirectly trains attention by ablating salient pixels from input images and maximizing the classification loss between the ablated and original images \cite{li2018tell}.
However, these approaches only train on clean data and will reinforce biased explanations if trained on obfuscated or biased data.
% Instead, we propose a self-supervised training approach to transfer explanations from an unbiased source to the model predicting on biased data.
Unlike conventional self-supervised learning with data augmentation and contrastive learning to improve feature learning~\cite{chen2020simple}, we use the unbiased explanation as a surrogate "label" to train the debiased model to predict a more faithful explanation.
% {\color{blue}\cite{chen2020simple, noroozi2018boosting, watanabe2017student}}

\section{Technical Approach}
\begin{figure*}[!ht]
    \centering    
    \includegraphics[width=16.0cm]{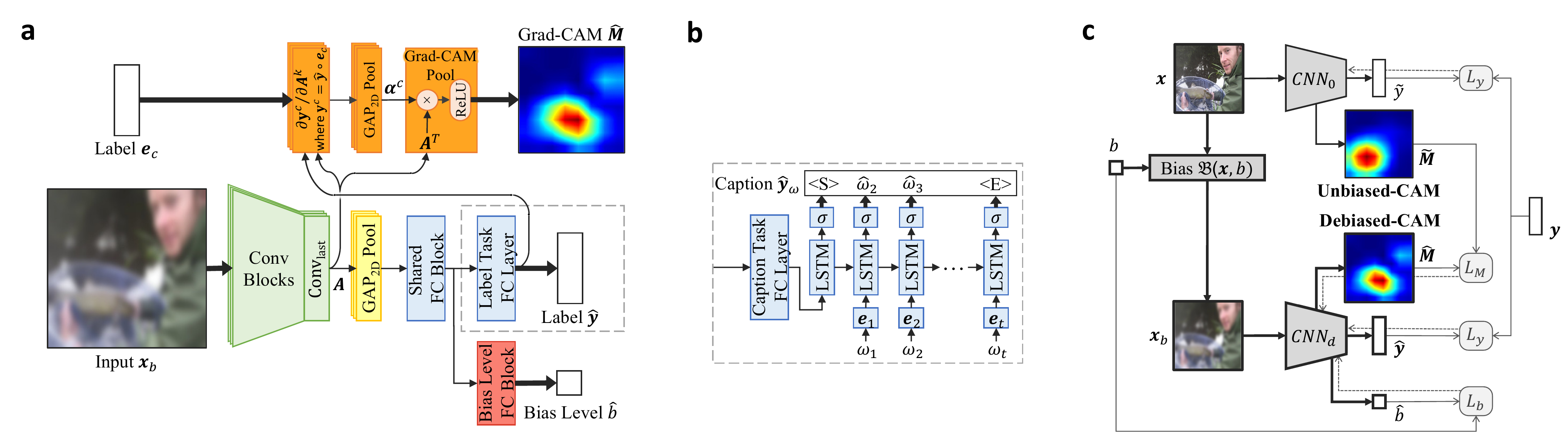}
    \vspace{-0.3cm}
    \caption{Architecture of DebiasedCNN. a) DebiasedCNN is a multi-input, multi-task convolutional neural network with two inputs image $\bm{x}_b$ and label $\bm{e}_c$ for class $c$, and three tasks for primary prediction task $\widehat{\bm{y}}$, CAM explanation task $\widehat{\pmb{M}}$, and bias level prediction task $\widehat{b}$. b) DebiasedCNN can be trained for different primary tasks, such as image captioning. c) Meta-architecture with self-supervised learning to minimize the CAM loss $L_M$ between Unbiased-CAM $\widetilde{\pmb{M}}$ from RegularCNN ($CNN_0$) predicting on unbiased image $\bm{x}$ and Debiased-CAM $\widehat{\pmb{M}}$ from DebiasedCNN ($CNN_d$) predicting on biased image $\bm{x}_b$ at bias level $b$. %\textbf{d}, A baseline FineTuneCNN ($CNN_f$) that is only trained to minimize the primary task loss $L_y$, will generate a deviated Biased-CAM $\breve{\pmb{M}}$ on a biased image $\bm{x}_b$.
    } 
    \label{fig:architecture}
    \Description[]{
    a) The architecture of the multi-input, multi-task convolutional
	neural network. The input image $\bm{x}_b$ is fed into a feature
	extractor (green) to get activation maps $\bm{A}$. The activation
	maps are further fed into a global average pooling layer (yellow)
	and fully-connected layers to support the primary classification task and bias level estimation task. Meanwhile, the activation maps and the label $\bm{e}_c$ are fed into three Grad-CAM relevant layers (orange) to produce the Grad-CAM saliency map {{M}}.
    b) The decoder structure in the image captioning task. The embedding is firstly passing through a fully connected layer and then serve as the initial hidden state. At each time step, the current word embedding and the hidden state vector is fed into LSTM to obtain the caption output.
    c)  Meta-architecture of self-supervised learning strategy. The unbiased image passes through the RegularCNN model ($CNN_0$) to get its prediction and Unbiased-CAM. The biased image, which is obtained by applying bias transform on the same unbiased instance, passes through the DebiasedCNN model ($CNN_d$) to get its prediction, bias estimation, and Unbiased-CAM. During this process, Unbiased-CAM serve as the supervision signal for calibration. 
    }
    \vspace{-0.3cm}
\end{figure*}

We first describe baseline RegularCNN and FineTunedCNN approaches to predict on unbiased and biased image data, then our proposed DebiasedCNN architectures to predict on biased image data with debiased explanations.

% \subsection{Deviated Explanation on Biased Image}
% \subsection{Regular and Fine-tuned single-task models}
\subsection{Regular and Fine-tuned Models}
% % % % % % % % % % % % % % % % % % % % % % % % % % % % % % % % % % % %
% * Reorganized From RESULTS: Self-Supervised Multi-Bias, Multi-Input, Multi-Task Model for Debiased-CAM
% % % % % % % % % % % % % % % % % % % % % % % % % % % % % % % % % % % %
A regularly trained CNN model (RegularCNN) can generate a truthful CAM $\widetilde{\pmb{M}}$ (Unbiased-CAM) of an unbiased image $\bm{x}$, but will produce a deviated CAM $\check{\pmb{M}}$ (Biased-CAM) for the image under bias $\bm{x}_b$, i.e., $\widetilde{\pmb{M}}\left(x\right) \neq \widehat{\pmb{M}}(\bm{x}_b)$,
%. This happens since the model is not trained on any biased images and thus cannot learn spurious correlations with blurred pixels. 
due to the model not training on any biased images and learning spurious correlations with blurred pixels.
A fine-tuned model %(FineTunedCNN, Fig. \ref{fig:architecture}d) 
trained on biased images can improve the prediction performance on biased images, but will still generate a deviated CAM $\check{\pmb{M}}$ (Fig. \ref{fig:biasedCAMs}a and Fig. \ref{fig:biasedCAMsAcrossDatasets}a-c: CAMs of FineTunedCNN), as it was only trained with the classification loss and not explanation loss. 
% {\color{orange}Our proposed method mitigates this issue by teaching CNN to pay attention more correctly such that it can produce less deviated CAM explanation even on the biased image $\bm{x}_b$.}
While these models can be explained with Grad-CAM, they are not retrainable to improve their CAM faithfulness.

% \subsection{DebiasedCNN Model to explain with Debiased-CAM}
\subsection{DebiasedCNN Model with Debiased-CAM Explanations}

\subsubsection{Trainable CAM as secondary prediction task}
\label{subsection:GradCAM}
% % % % % % % % % % % % % % % % % % % % % % % % % % % % % % % % % % % %
% * Reorganized from Method 1: Debiasing CAM Prediction Task with Differentiable CAM Loss
% % % % % % % % % % % % % % % % % % % % % % % % % % % % % % % % % % % %
We enable CAM retraining by redefining Grad-CAM as a prediction task.
Grad-CAM \cite{selvaraju2017grad} computes a saliency map explanation of an image prediction with regards to class $c$ as the weighted sum of activation maps in the final convolutional layer of a CNN. 
Each activation map $\bm{A}^k$ indicates the activation $\bm{A}_{ij}^k$ for each grid cell $(i,j)$ of the $k$th convolution filter $k \in K$ (set of all filters).
The importance weight $\alpha_k^c$ for the $k$th activation map is calculated by back-propagating gradients from the output $\hat{\bm{y}}$ to the convolution filter, i.e.,
% $\alpha_k^c=\frac{1}{HW}\sum_{i=1}^{H}\sum_{j=1}^{W} \frac{\partial \bm{y}^c}{\partial \bm{A}_{ij}^k} = {GAP}_{ij}\left(\frac{\partial \bm{y}^c}{\partial \bm{A}^k}\right),$
\begin{equation}
    \alpha_k^c=\frac{1}{HW}\sum_{i=1}^{H}\sum_{j=1}^{W} \frac{\partial \bm{\hat{y}^c}}{\partial \bm{A}_{ij}^k} \equiv {GAP}_{ij}\left(\frac{\partial \bm{\hat{y}}^c}{\partial \bm{A}^k}\right)
\label{eq:GradCAM_alpha}
\end{equation}
where $H$ and $W$ are the height and width of activation maps, respectively; $\bm{y}^c$ is a one-hot vector indicating only the probability of class $c$; $GAP_{ij}(\cdot)$ is the global average pooling operation. 
The class activation map (CAM) is the weighted combination of activation maps, followed by a ReLU transform to only show positive activations for class $c$, i.e.,
% $M^c\left(i,j\right)=ReLU{\left(\sum_{k}{\alpha_k^c \bm{A}^k}\right)} \equiv \widehat{\pmb{M}}=ReLU\left(\alpha^c \bm{A}^T\right),$
\begin{equation}
    M^c=\text{ReLU}{(\sum_{k}{\alpha_k^c \bm{A}^k})} \equiv \widehat{\pmb{M}}=\text{ReLU}\left(\alpha^c \bm{A}^T\right)
    \vspace{-0.2cm}
\label{eq:GradCAM_heatmap}
\end{equation}
which we rewrite as a matrix multiplication of all $K$ %$K=\left|\mathcal{K}\right|$ 
importance weights $\bm{\alpha}^c=\left\{\alpha_k^c\right\}^K$ and the transpose of activation maps $\bm{A}$ along the $k$th axis, i.e., $\bm{A}^T=\left\{\bm{A}_{ij}^k\right\}^{K \times H \times W}$. 

Therefore, the CAM prediction task can be redefined as three non-trainable layers (computational graph) in the neural network (orange in Fig. \ref{fig:architecture}a) to compute $\frac{\partial \bm{y}^c}{\partial \bm{A}^k}$, $\alpha_k^c$, and $\widehat{\pmb{M}}$.
By reformulating Grad-CAM as a \textit{secondary prediction task}, we can train the model with faithful CAM based on differentiable CAM loss by backpropagating through this task.
% Instead of treating Grad-CAM as a post-hoc explanation, we explicitly formulate Grad-CAM as a differentiable prediction task. More concretely, three layers are stacked in the neural network (orange in Fig. \ref{fig:architecture}a) to compute $\frac{\partial \bm{y}^c}{\partial \bm{A}^k}$, $\alpha_k^c$, and $\widehat{\pmb{M}}$, respectively. Given a reference label on Grad-CAM, we can train the model with the CAM loss. While the backpropagation pass through these layers, they contain no trainable parameters, hence the CAM explanations still explain the model with its activation maps. 
This task takes $\bm{e}_c$ as the second input to the CNN architecture to specify the target class label for the CAM. $c$ is set as the ground truth class label at training time, and chosen by the user at run time. 
We call the aforementioned approach Multi-Task DebiasedCNN, and call the conventional use of Grad-CAM as Single-Task DebiasedCNN.
For single-task DebiasedCNN, the loss is
% backpropagated through the primary classification task
added as a simple sum to the primary classification task,
% , i.e., CAM loss is calculated as a weighted sum number 
rather than predicted with secondary task. This will limit its learning since weights are not updated with gradient descent.

% \subsubsection{Self-supervised DebiasedCNN}
\subsubsection{Training CAM debiasing with Self-Supervised Learning}
% To debias explanations from biased images, 
To debias CAMs $\widehat{\pmb{M}}$ of biased images $\bm{x}_b$ toward truthful Unbiased-CAMs $\widetilde{\pmb{M}}$ of clean images $\bm{x}$, 
i.e., $\widehat{\pmb{M}}\left(\bm{x}_b\right)\approx\widetilde{\pmb{M}}(\bm{x})$,
we train DebiasedCNN with self-supervised learning 
% {\color{blue}\cite{chen2020simple, noroozi2018boosting, watanabe2017student}}
to transfer knowledge of corresponding unbiased images in RegularCNN into DebiasedCNN.
% % % % % % % % % % % % % % % % % % % % % % % % % % % % % % % % % % % %
% * Some new description about the rationale that we construct the self-supervised learning setting.
% % % % % % % % % % % % % % % % % % % % % % % % % % % % % % % % % % % %
% {\color{orange}One key challenge of learning with CAM loss is the lack of a reference CAM for each image instance. However, we notice that most of the unbiased explanations (i.e. unbiased images evaluated on RegularCNN) are more truthful and less likely to be misleading. We can leverage knowledge from unbiased explanations to guide the model's training on biased images. To this end, we develop a self-supervised training framework where biased images are paired with their clean versions, such that unbiased explanations can be used to improve the biased explanations.}
%
% % % % % % % % % % % % % % % % % % % % % % % % % % % % % % % % % % % %
% * Reorganized From RESULTS: Self-Supervised Multi-Bias, Multi-Input, Multi-Task Model for Debiased-CAM
% % % % % % % % % % % % % % % % % % % % % % % % % % % % % % % % % % % %
%
We aim to minimize the difference between Unbiased-CAM $\widetilde{\pmb{M}}$ and Debiased-CAM $\widehat{\pmb{M}}$.
The training involves the following steps (see Fig. \ref{fig:architecture}c): 
1) Given a dataset with clean images $\bm{x} \in X$ and labels $y$, apply a bias transformation (e.g., blur) to create biased variants of each image $\bm{x}_b \in X_b$. 
2) Train a RegularCNN to predict label $\tilde{y}$ on clean image $\bm{x}$. We assume that its Grad-CAM explanations $\widetilde{\pmb{M}}$ are correct and serve as a good oracle for Unbiased-CAMs. 
3) Train a DebiasedCNN to predict label $\hat{y}$ on corresponding biased image $\bm{x}_b$, and explain with CAM $\widehat{\pmb{M}}$. DebiasedCNN is trained with loss function:
\begin{equation}
    L=
    {L_y\left(y,\widehat{y}\right)}+
    \omega_M{L_M\left(\widetilde{\pmb{M}},\widehat{\pmb{M}}\right)}
\label{eq:debiasCAM_loss}
\end{equation}
% where the classification loss $L_y$ and CAM loss $L_M$ are differentiable 
% with respect to their corresponding prediction tasks 
where $L_y$ is the classification loss, $L_M$ is the CAM loss, and $\omega_M$ is a hyperparameter. 
% We compute $L_M$ with mean square error (MSE) loss
% and $L_y$ depending on the primary task (described later). 
The training can be interpreted as attention transfer from an unbiased model to the new model.
DebiasedCNN can be generalized to image prediction other tasks (e.g., image captioning: Fig. \ref{fig:biasedCAMsAcrossDatasets}b), 
other bias types (e.g., color temperature, lighting: Fig. \ref{fig:biasedCAMsAcrossDatasets}c,d),
different base CNN models (e.g., VGG16, Inception v3, ResNet50, Xception),
and for privacy-preserving machine learning. 
% (Supplementary Fig. \ref{sfig:architecturePrivacy})

% {\color{orange} 
% Before training, we construct biased images by applying bias transformation (e.g. blur) on clean images. 
% RegularCNN plays the role of an oracle, which is pretrained on unbiased images. 
% During the training of DebiasedCNN, when each biased image $\bm{x}_b$ is fed into DebiasedCNN, we query RegularCNN with the corresponding unbiased image instance $\bm{x}$ to obtain unbiased-CAM $\widetilde{\pmb{M}}$, and then used it to calibrate Debiased-CAM $\widehat{\pmb{M}}$. During this process, we force DebiasedCNN to pay attention more correctly by transferring attention knowledge from RegularCNN.}   
%$\widehat{\mathbf{M}}$ $\widehat{\pmb{M}}$ $\widehat{\boldsymbol{M}}$ $\widehat{\pmb{M}}$ 

% % % % % % % % % % % % % % % % % % % % % % % % % % % % % % % % % % % %
% * Reorganized From RESULTS: Self-Supervised Multi-Bias, Multi-Input, Multi-Task Model for Debiased-CAM
% % % % % % % % % % % % % % % % % % % % % % % % % % % % % % % % % % % %
% The proposed multi-task model is trained without human annotation using self-supervised learning \cite{noroozi2018boosting, watanabe2017student}; this is unlike model fine-tuning which only trains with classification loss.%(Fig. \ref{fig:architecture}d)

% \subsection{Model Variants and Training Loss Functions}
% \subsubsection{Multi-bias DebiasedCNN}
\subsubsection{Bias-agnostic, Multi-bias predictions with tertiary task}
Image biasing can happen sporadically at run time, so the image bias level $b$ may be unknown at training time. Instead of training on specific bias levels \cite{dimiccoli2018mitigating} or fine-tuning with data augmentation on multiple bias levels \cite{vasiljevic2016examining}, we added a \textit{tertiary prediction task} --- bias level regression --- to DebiasedCNN to leverage supervised learning (Fig. \ref{fig:architecture}a: salmon-colored layers). This enables DebiasedCNN to be bias-aware (can predict bias level) and bias-agnostic (predict under any bias level).
% In summary, DebiasedCNN has multiple capabilities to predict Debiased-CAMs at various bias levels. 
With the bias level prediction task, the training loss function for multi-bias, multi-task DebiasedCNN is: 
% \begin{equation}
%     L\left({w}\right)={L_y\left(y,\widehat{y}\left({w}\right)\right)}+\omega_M{L_M\left(\widetilde{\pmb{M}},\widehat{\pmb{M}}\right)}+\omega_b{L_b\left(b,\widehat{b}\right)}
% \label{eq:debiasCAM_loss}
% \end{equation}
\begin{equation}
    L={L_y\left(y,\widehat{y}\right)}+\omega_M{L_M\left(\widetilde{\pmb{M}},\widehat{\pmb{M}}\right)}+\omega_b{L_b\left(b,\widehat{b}\right)}
\label{eq:debiasCAM_mb_loss}
\end{equation}
% where the classification loss $L_y$, CAM loss $L_M$ and bias prediction loss $L_b$ are differentiable with respect to corresponding tasks, $\omega_M$, 
where $L_b$ is the bias prediction loss, and
$\omega_b$ is a hyperparameter.

\subsubsection{Training loss terms}
% The specific loss terms are defined as follows: 
In all, there are three loss terms:
primary task loss $L_y$ as cross-entropy loss for standard classification tasks, and as the sum of negative log likelihood for each word \cite{vinyals2015show} in image captioning tasks; bias level loss $L_b$ as the mean squared error (MSE), common for regression tasks; CAM loss $L_M$ as the mean squared error (MSE), since CAM prediction can be considered a 2D regression task, and this is common for visual attention tasks \cite{komodakis2017paying}. 
% Other suitable metrics for the CAM loss include: mean absolute error (MAE) which penalizes large differences less than MSE; Kullback-Leibler Divergence (KLD) or Jensen-Shannon Distance (JSD) which compare the distribution of pixel saliency between CAMs, but are more expensive to calculate; and Pearson’s Correlation Coefficient (PCC) which compares the pixel-wise correlation between CAMs, but is also computationally expensive for training.

% % % % % % % % % % % % % % % % % % % % % % % % % % % % % % % % % % % %
% * Reorganized From Method 3: Model Variants and Training Loss Functions, 
% * Some description about the alternative metrics are cut.
% % % % % % % % % % % % % % % % % % % % % % % % % % % % % % % % % % % %
\subsubsection{Summary of DebiasedCNN Model Variants}
DebiasedCNN has a modular design: 1) single-task (st) or multi-task (mt) to improve model training; and 2) single-bias (sb) or multi-bias (mb) to support bias-aware and bias-agnostic predictions. 
We denote the four DebiasedCNN variants as (sb, st), (mb, st), (sb, mt), (mb, mt), and conducted ablation studies to compare between them.
Supplementary Fig. \ref{sfig:modelVariants} and Supplementary Table \ref{table:table-modelVariants} show details of %the model variant architectures and training losses.
each variant.
\section{Simulation Studies}
To evaluate how much CAMs deviate with biased images and how well DebiasedCNN recovers CAM Faithfulness, we conducted five simulation studies with varying datasets, prediction tasks (classification, captioning), bias types (blur, color temperature, day/night lighting), and bias levels.
% For all studies, we measured model Task Performance as the area under the precision-recall curve (PR AUC, Method \ref{5}) and CAM Faithfulness as the Pearson’s Correlation Coefficient (PCC) and with the Jensen-Shannon Divergence (JSD) between the CAM and Unbiased-CAM (Method \ref{6}; JSD results in Supplementary Fig. \ref{3}). DebiasedCNN showed improvements across all simulation studies with some differences which we highlight.
These studies inform which applications explanation biasing is problematic, and show that our debiased training can successfully mitigate these deviations.

\subsection{Evaluation Metrics}
% % % % % % % % % % % % % % % % % % % % % % % % % % % % % % % % % % % % 
% * From Method 5: Simulation Studies Model Task Performance Metrics. 
% * From Method 6: Simulation Studies CAM Faithfulness Metrics, 
% * Details of the metrics are cut. 
% % % % % % % % % % % % % % % % % % % % % % % % % % % % % % % % % % % % 

We evaluated \textit{prediction performance} and \textit{CAM explanation faithfulness} to compare model variants. For classification, we measured the area under the precision-recall curve (PR AUC) as it is robust against imbalanced data \cite{saito2015precision}, and calculated the class-weighted macro average to aggregate across multiple classes. For image captioning, we calculated the BLEU-4 \cite{papineni2002bleu} score that measures how closely 4-grams in the predicted and actual captions matched. 
For bias level regression, we calculated accuracy with $R^2$. 
% (Supplementary Fig. \ref{sfig:simulationPlot_R2}).
%
We define the correctness of CAM explanations by their similarity or \textit{faithfulness} to the original Unbiased-CAMs from RegularCNN that infers on unbiased data.
To better compare CAMs beyond simple residual differences (e.g., MAE, MSE), we calculated CAM Faithfulness as the Pearson’s Correlation Coefficient (PCC) \cite{li2015data, bylinskii2018different} of pixel-wise saliency as it closely matches the human perception to favor compact locations and match the number of salient locations \cite{li2015data}, and it fairly weights between false positive and false negatives \cite{bylinskii2018different}. 
% See Appendix \ref{subsection:jsd} and Supplementary Fig. \ref{sfig:simulationPlots_JSD} for CAM faithfulness calculated with another saliency metric, JSD.

\subsection{Results}
\begin{figure*}[ht!]
    \centering   
    % \hspace*{-0.3cm}
    \vspace{-0.2cm}
    \includegraphics[width=18.0cm]{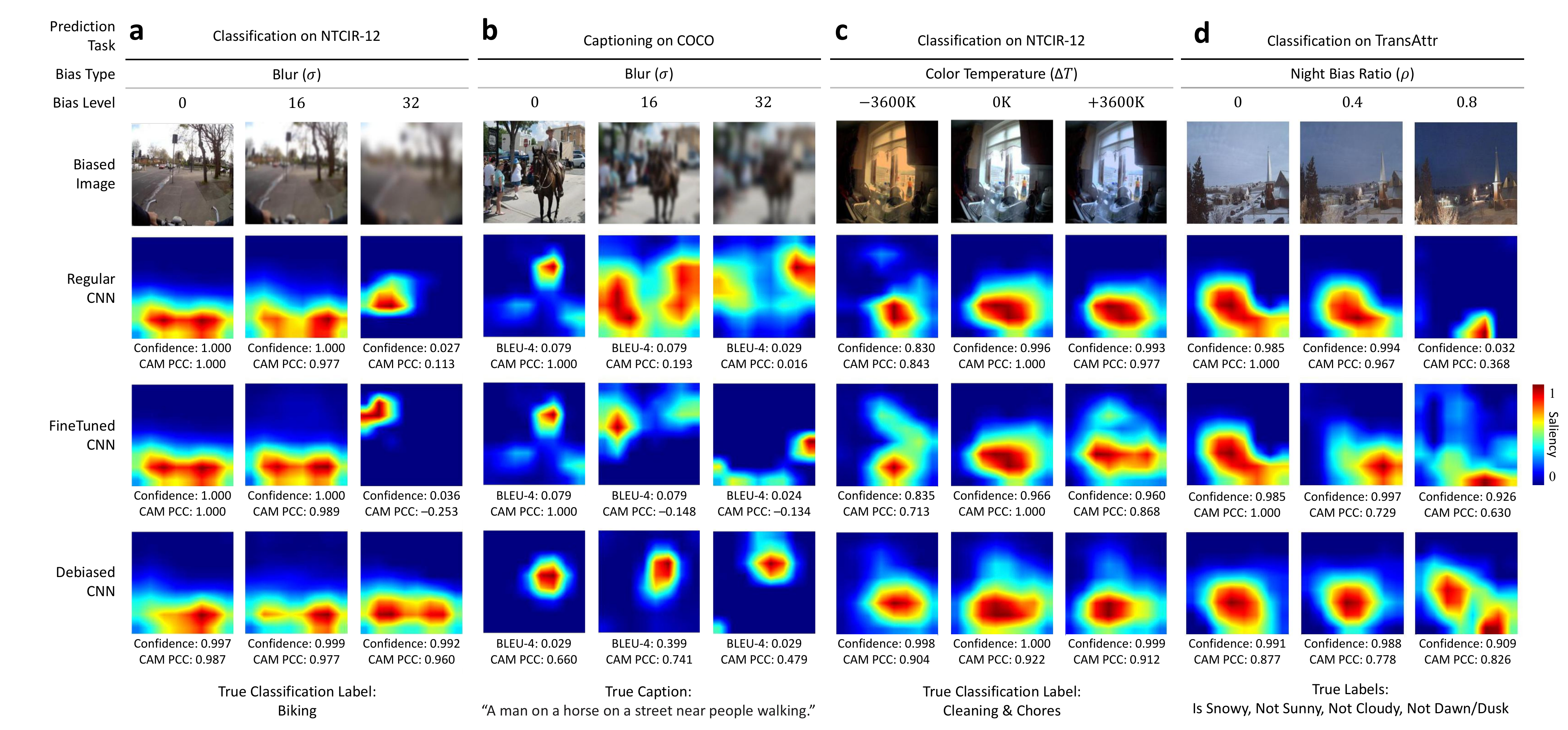}
    \vspace{-0.6cm}
    \caption{
    Deviated and debiased CAM explanations from models trained on different prediction tasks (a-d) with varying bias levels.
    In general, RegularCNN and FineTunedCNN had deviated CAMs that missed selecting important pixels, while DebiasedCNN had CAMs similar to Unbiased-CAMs.
    At no bias, all CAMs from RegularCNN and FineTunedCNN are unbiased.}
    
    % a) %For wearable camera activity recognition, 
    % RegularCNN and FineTunedCNN had deviated CAMs where bicycle handlebars became less salient, unlike DebiasedCNN. 
    % b) %For image captioning, 
    % % Regular/FineTunedCNN had wildly deviated CAMs as blur bias increased, where background people and other areas were highlighted; while DebiasedCNN had less deviated CAMs that still selected the horse and rider. c) %For wearable camera activity recognition, 
    % Regular/FineTunedCNN had CAMs which were more deviated for orange-bias than for blue-bias; whereas DebiasedCNN had less deviated CAMs, highlighting the kitchen sink. 
    % d) %For wearable camera activity recognition, 
    % Regular/FineTunedCNN had CAMs which deviated to not highlight snowy regions on the scene; whereas DebiasedCNN had less deviated CAMs, still highlighting that region.
    % a-d) At no bias, all CAMs from RegularCNN and FineTunedCNN are unbiased.
    % }
    \label{fig:biasedCAMsAcrossDatasets}
    \Description{
    Demos of images and different types of GradCAMs on different tasks and bias types. subfigures a), b), c), d) share the same layout.
    a) The 1st row shows the same photo (an egocentric photo of biking activity) blurred with 3 different levels (sigma = 0, 16, 32), where a man is holding a fish. The 2nd row shows the GradCAM visual explanations of RegularCNN on corresponding blurred images. The red region indicates a higher importance while the blue region indicates a lower importance. The text below the image indicates the confidence score of the image instance and the CAM truthfulness score (via PCC). The 3rd row and 4th row indicate the GradCAM visual explanations of FinetunedCNN and DebiasedCNN respectively.
    b) shows the same photo (with a man riding a horse) blurred with 3 levels.
    c) shows the same photo (kitchen from an egocentric view) with 3 levels of color temperature.
    d) shows the same photo (with an snowy outdoor scene) capture on 3 different daytime.
    }
    \vspace{-0.3cm}
\end{figure*}

\begin{figure*}[ht!]
    \centering    
    \includegraphics[width=17.2cm]{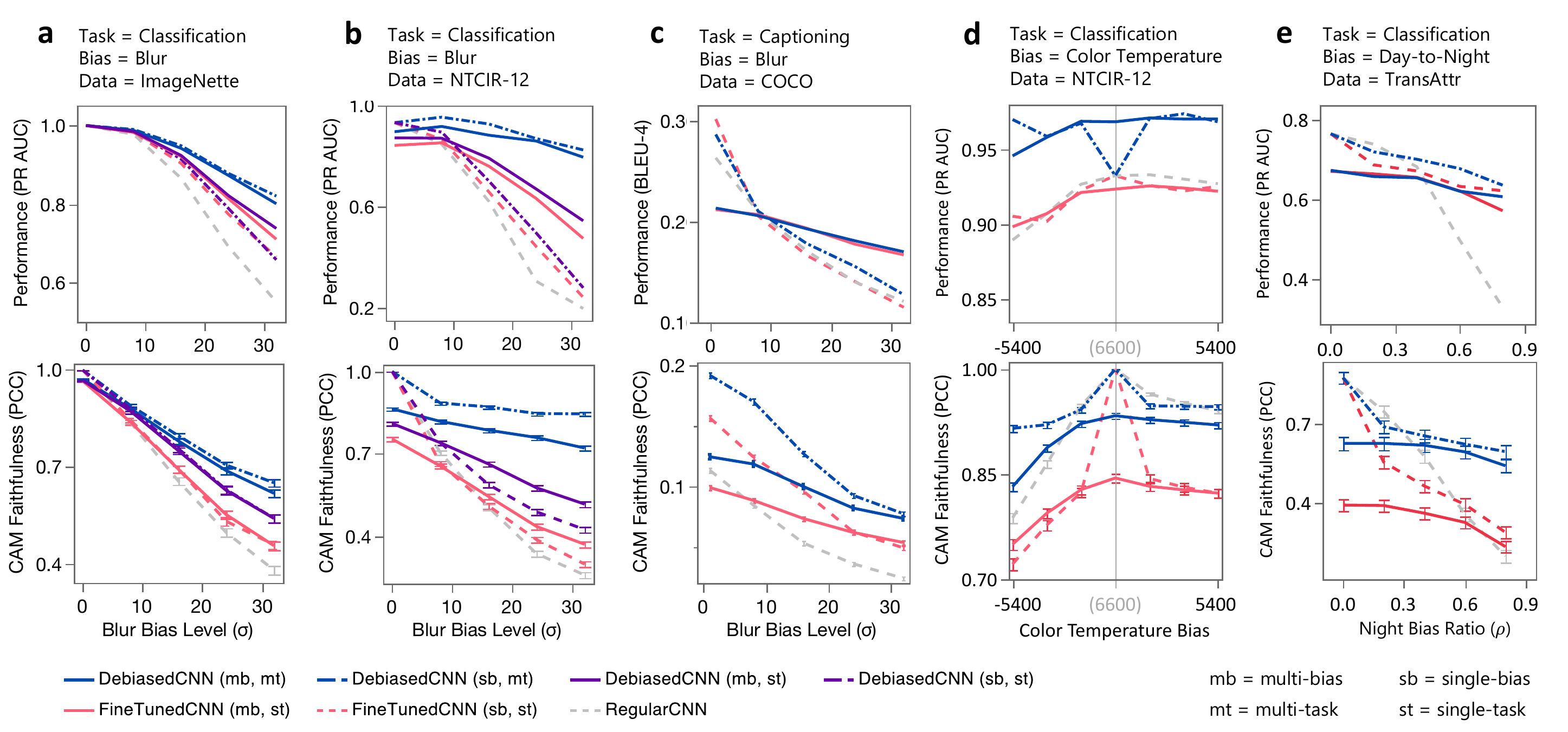}
    \vspace{-0.2cm}
    \caption{
    Task Performance and CAM Faithfulness for different prediction tasks with increasing bias levels. 
    a)-e) All models' Task Performance and CAM Faithfulness decreased with increasing blur, while DebiasedCNN decreased the least. 
    % DebiasedCNN improved Task Performance over RegularCNN and FineTunedCNN variants due to attention transfer with higher CAM Faithfulness. 
    For DebiasedCNN variants, multi-task had the highest CAM Faithfulness and Task Performance that is higher than single-task. 
    % \textbf{a}, Task Performance for no blur ($\sigma = 0$) was almost perfect ($\text{PR AUC} \approx 1$) because models were pre-trained on the superset ImageNet dataset. 
    % \textbf{b}, Task Performance and CAM Faithfulness decreased more sharply than a, since ImageNette is a cleaner dataset than NTCIR-12 with fewer label classes. In contrast, multi-bias DebiasedCNN maintained high Task Performance and CAM Faithfulness across all blur levels, indicating very effective debiasing. 
    % \textbf{c}, Task Performance and CAM Faithfulness were lower for image captioning than for classification since captioning is more complex. 
    % \textbf{d}, DebiasedCNN had the best Task Performance and CAM Faithfulness even as they decreased asymmetrically about 6600K. 
    % \textbf{a-d}, Model variants annotated as st = single-task, mt = multi-task, sb = single-bias, mb = multi-bias. Error bars indicate 90\% confidence interval. 
    %\textbf{a-c}, Image blur bias was varied with the standard deviation $\sigma$ of the Gaussian blur for the normalized image. d, Color temperature bias (in Kelvin) was varied with respect to neutral cloudy daylight at 6600K.
    % \textbf{e}, Similar results as \textbf{a} with DebiasedCNN out-performing other models with increasing bias ratio.
    }
    \vspace{-0.4cm}
    \Description{
    Barcharts of evaluation result on different prediction tasks, bias types and datasets. 5 subfigrues share the same layout.
    a) Two plots show the PRAUC (for classification performance) and PCC (for CAM faithfulness measurement) across different bias levels for various CAM conditions on blur biased ImageNette dataset. Generally, PRAUC and PCC decrease with a stronger blur bias level, while DebiasedCNN > FinetunedCAM > RegularCNN.
    b) Two plots show the PRAUC (for classification performance) and PCC (for CAM faithfulness measurement) on blur biased NTCIR dataset. Generally, PRAUC and PCC decrease with a stronger blur bias level, while DebiasedCNN > FinetunedCAM > RegularCNN.
    c) Two plots show the BLUE-4 (for image captioning performance) and PCC (for CAM faithfulness measurement) on blur biased COCO dataset. Generally, PRAUC and PCC decrease with a stronger blur bias level, while DebiasedCNN > FinetunedCAM > RegularCNN. The trends are relatively flat since the captioning task is more chanlleging.
    d) Two plots show the PRAUC (for classification performance) and PCC (for CAM faithfulness measurement) on color temperature biased NTCIR dataset. Generally, PRAUC and PCC decrease with a stronger blur bias level (far from 6600K), while DebiasedCNN > RegularCNN > FinetunedCAM.
    d) Two plots show the PRAUC (for classification performance) and PCC (for CAM faithfulness measurement) on day-night-conversion biased transAttr dataset. Generally, PRAUC and PCC decrease with a stronger blur bias level, while DebiasedCNN > RegularCNN > FinetunedCAM.
    }
    \label{fig:simulationPlots}
\end{figure*}

In general, CAMs deviate more from Unbiased-CAMs as bias levels increased, but DebiasedCNN reduces this deviation. Debiased retraining also improved model prediction performance, which suggests that DebiasedCNN indeed "sees through the bias".
Fig. \ref{fig:simulationPlots} shows our evaluation Task Performance and CAM Faithfulness in ablation studies across increasing bias levels for different prediction tasks and datasets (Supplementary Table \ref{table:table-datasetDescription}). 
% Supplementary Table \ref{table:simulationResult} describes in detail these improvements. 
Fig. \ref{fig:biasedCAMsAcrossDatasets} shows some examples of deviated and debiased CAMs.
Next, we describe the experiment method and results for each simulation study.

\subsubsection{Simulation Study 1 (Blur Bias)}
We evaluated CAMs for blur biased images of the object recognition dataset ImageNette~\cite{imagenette}.
% We varied images to bias them with three types, blur, color temperature and night vision at different bias levels.
% \footnote{{\color{blue}We assume DebiasedCNN(sb, mt) = DebiasedCNN(sb, st) = FineTunedCNN(sb, st) = RegularCNN when $\sigma=0$. This is because DebiasedCNN or FineTunedCNN do not exist when no bias occurs in images.}}
% We blurred images by applying a uniform Gaussian blur filter \cite{dimiccoli2018mitigating} using opencv-python v4.2.0. 
We scaled images to a standardized maximum size of 1000×1000 pixels and applied uniform Gaussian blur at various standard deviations $\sigma$. 
We found that Task Performance and CAM Faithfulness decreased with increasing blur level for all CNNs, but DebiasedCNN mitigated these decreases (Fig. \ref{fig:simulationPlots}a). This indicates that model training with additional CAM loss improved model performance rather than trading-off explainability for performance \cite{rudin2019stop}. 
RegularCNN had the worst Task Performance and the lowest CAM Faithfulness for all blur levels ($\sigma>8$). 
In comparison, trained with differentiable CAM loss, DebiasedCNN (sb, mt) showed marked improvements to both metrics, up to 2.33x and 6.03x over FineTunedCNN’s improvements, respectively. 
Trained with non-differentiable CAM loss, DebiasedCNN (sb, st) improved both metrics to a lesser extent than DebiasedCNN (sb, mt), confirming that separating the CAM task from the classification task 
% in the latter variant 
enabled better weights update 
% in model training
. 
Trained with an additional bias-level task, multi-bias DebiasedCNN (mb, mt) achieved high Task Performance and CAM Faithfulness for all bias levels that is only marginally lower than single-bias DebiasedCNN (sb, mt), because of the former's good regression performance for bias level prediction (Supplementary Fig. \ref{sfig:simulationPlot_R2}). 
% Multi-bias DebiasedCNN generalizes across bias levels better than single-bias DebiasedCNN when evaluated at non-specific bias levels (Supplementary Fig. \ref{sfig:simulationPlots_singlebiasEvaluation}). 
% Finally, all models generated more faithful CAMs when they had a higher Prediction Confidence (Supplementary Fig. \ref{sfig:simulationPlots_faithfulnessvsConfidence}).

\subsubsection{Simulation Study 2 (Blur Bias, Egocentric)}
\label{subsubsection:simulation_study2}
We evaluated the impact of blur biasing with a more ecologically realistic task --- wearable camera activity recognition (NTCIR-12 \cite{gurrin2016ntcir}). This task\footnote{Note that we mean that the use case could have blurred images, not that the NTCIR-12 dataset has blurred images. Blurring or biasing ImageNet photos (which may include curated stock photos) is an unrealistic use case, but there are more ecologically legitimate reasons for egocentric photos to be biased.} represents a real-world use case where egocentric cameras may capture blurred images accidentally due to motion or defocus, or deliberately for privacy protection. 
We found the same trends as for the ImageNette classification task with some differences due to the increased task difficulty (Fig. \ref{fig:simulationPlots}b). In particular, the differences between RegularCNN and DebiasedCNN in Task Performance and CAM Faithfulness were amplified, indicating that debiasing is more useful for this application. 
Task Performance and CAM Faithfulness decreased steeply for RegularCNN with increasing blur bias, while DebiasedCNN significantly recovered both metrics, demonstrating marginal decreases with increasing bias. 
FineTunedCNN marginally increased CAM Faithfulness from RegularCNN ($<44\%$), while DebiasedCNN achieved a much larger improvement by up to 229\%. 
We verified these trends for different CNN backbones and found that more accurate models produced more faithful CAMs even for stronger blur (Supplementary Figs. \ref{sfig:biasedCAMs_acrossBackbones} and \ref{sfig:simulationPlot_acrossBackbones}). 
Hence, Debiased-CAM enables privacy-preserving wearable camera activity recognition with improved performance and faithful explanations.

\subsubsection{Simulation Study 3 (Blur Bias Captioning)}
We evaluated the influence of blur on a different prediction task --- image captioning (COCO \cite{chen2015microsoft}). 
We found similar trends in Task Performance and CAM Faithfulness as before, though all models performed poorly at all blur levels (Fig. \ref{fig:simulationPlots}c). Furthermore, CAM Faithfulness was low for all models, even for RegularCNN at a small blur bias ($\sigma=1$). 
This could be because captioning is much harder than classification, and CAM retraining is weakened by vanishing gradients due to the long LSTM recurrence.
%is affected by the gradient calculation inside LSTM blocks, such that CAM explanations were inaccurate even for barely biased images. 
Yet, DebiasedCNN improved CAM Faithfulness for all blur levels by up to 224\% from RegularCNN.
%, indicating the importance of attention transfer at the model’s convolution layers from Unbiased-CAM to retain CAM faithfulness.

\subsubsection{Simulation Study 4 (Color Temperature Biased)}
We evaluated color temperature bias on wearable camera images in NTCIR-12. 
This represents another realistic problem for the wearable camera use case, where the white balance may be miscalibrated. 
We set the neutral color temperature $t$ to 6600K (cloudy/overcast) and perturbed the color temperature bias by applying Charity’s color mapping function to map a temperature to RGB values \cite{blackbody}.
Color temperature can be bidirectionally biased towards warmer (more orange, lower values) or cooler (more blue, higher values) temperatures from neutral 6600K. 
Furthermore, image pixel values deviate asymmetrically with larger deviations for orange than for blue biases. Consequently, we found that orange bias led to a larger decrease in Task Performance and CAM Faithfulness than blue bias (Fig. \ref{fig:simulationPlots}d). 
% More details in Appendix \ref{subsubsection:color_bias}.
Notably, CAM deviation was smaller across all color temperature biases than for blur biases, as indicated by the smaller decrease in CAM Faithfulness (compare Fig. \ref{fig:simulationPlots}b, d); hence, Task Performance also did not decrease as much as blur bias. FineTunedCNN had similar Task Performance but lower CAM Faithfulness than RegularCNN; this suggests that color-biased images were too similar to improve model training with classification fine-tuning, and yet this significantly degraded explanation quality. In contrast, DebiasedCNN improved Task Performance and CAM Faithfulness compared to RegularCNN. Furthermore, due to bidirectional bias, multi-bias training enabled DebiasedCNN (mb, mt) to have significantly higher Task Performance even for unbiased images ($\Delta t_b=0$).

\subsubsection{Simulation Study 5 (Lighting Bias)}
We evaluated lighting bias for outdoor scenes for a multi-label scene attribute recognition task (transient attribute database, TransAttr~\cite{laffont2014transient}).
Lighting in outdoor scenes regularly change across hours or seasons due to transient attributes, such as sunlight or weather changes. Hence, models trained on images captured in one lighting condition may predict and explain differently under other conditions. 
Specifically, for the multi-label prediction task of classifying whether a scene is Snowy, Sunny, Foggy, or Dawn/Dusk, we biased whether the scene was daytime or nighttime. 
We performed a pixel-wise interpolation with ratio $\rho$ to simulate interstitial periods between day and night (details in Appendix \ref{subsubsection:lighting_bias}).
% This represents a realistic circumstance that scenes change dramatically by hours and seasons, and this can influence the model behavior.
We found similar trends in Task Performance and CAM Faithfulness as with previous blur-biased classification tasks.
The image prediction training was biased towards day-time photos, and as photos became darker to represent dusk or night time, all models generated more deviated, but least so for DebiasedCNN.
Given the regularity and frequency of outdoor scenes changes, this study demonstrates the prevalence of biasing in model predictions and explanations, and emphasizes the need for Debiased-CAMs.

\section{User Studies}
\label{section:user_studies}
% In addition to simulation, we conducted two user studies to measure user perception of CAM truthfulness and helpfulness for Unbiased, Debiased and Biased CAMs.
Having found that DebiasedCNN improves CAM faithfulness, we next evaluated how well Debiased-CAM improves human interpretability over Biased-CAM. We conducted user studies to evaluate their perceived truthfulness (User Study 1) and helpfulness (User Study 2) in an AI verification task for a hypothetical smart camera with privacy blur filters, label predictions and CAM explanations, i.e., the Simulation Study 1 prediction task. 
Both studies had a 3$\times$3 factorial design with two independent variables — Blur Bias level (None $\sigma=0$, Weak $\sigma=16$, Strong $\sigma=32$) and CAM type (Unbiased, Debiased, and Biased). 
Unbiased-CAM is the CAM from RegularCNN predicting on the unbiased image regardless of blur bias level; Debiased-CAM is the CAM from DebiasedCNN (mb, mt) and Biased-CAM is the CAM from RegularCNN predicting on the biased image at corresponding Blur Bias levels. At the None blur level, Biased-CAM is identical to Unbiased-CAM.
The user studies were approved by our university Institutional Review Board.

\subsection{User Study 1 (CAM Truthfulness)}
The first study evaluated the perceived truthfulness of Unbiased, Debiased, and Biased CAMs.

\subsubsection{Experiment Procedure}

Participants: 
1) read the introduction and gave consent; 
2) studied a tutorial about automatic image labeling, privacy blurring, heatmap explanations, and how to interpret the survey questions; 
3) answered four screening questions to test their labeling of an unblurred and a weakly blurred image and their selection of important locations in an image and a CAM; 
4) if screening was passed (all correct answers), answered background questions on technology savviness and image comprehension, 
% otherwise was directed to the survey end; 
performed the main study with 10 trials; and ended with demographic questions.
See Supplementary Figs. \ref{sfig:userStudy1_screening}-\ref{sfig:userStudy1_Trial} for questionnaire details.

\begin{figure}[!t]
    \centering    
    \vspace{-0.1cm}
    \includegraphics[width=8.0cm]{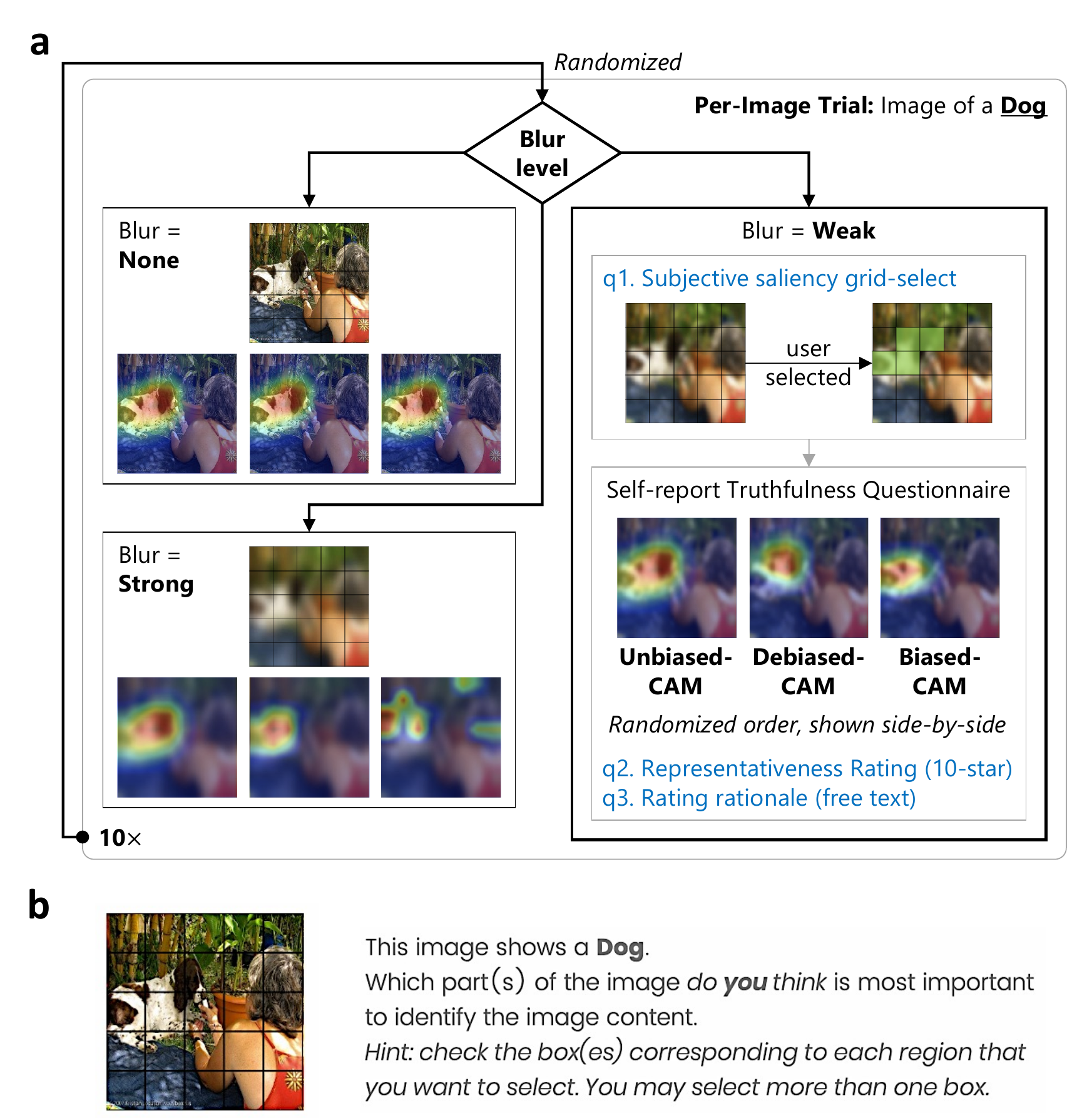}
    \vspace{-0.1cm}
    \caption{
    User Study 1 main study procedure (a) and grid selection UI to measure CAM Truthfulness Selection (b).
    }
    \label{fig:userStudy1Procedure}
    \Description{
    a) The flowchart of User study 1. Each participant will be randomly allocated to 1 out of 3 blur-level conditions. In each condition, the participant will view an image (e.g. a human play with dog) overlaid with grids, and 3 Grad-CAM explanations. She has to grid salient grids, rate the representativeness of CAMs and indicate rationale of rating.
    b) The UI of grid-select question: The photo overlaid with grids shown on the left, the question description on the right.
    }
    \vspace{-0.5cm}
\end{figure}

In the main study (Fig. \ref{fig:userStudy1Procedure}a), each participant viewed 10 repeated image trials, where each trial was randomly assigned to one of the three Blur Bias levels (within-subjects). All participants viewed the same 10 images (selection criteria described in Appendix \ref{section:image_selection}) in random order. For each trial, the participant: 
viewed a labeled unblurred image, indicated the most important locations on the image regarding the label with a “grid selection” UI (q1); and 
in the next page, viewed the blurred image, viewed CAMs of all 3 types generated from that and arranged randomly side-by-side, rated how well each CAM represented the image label on a 10-star scale (q2), and wrote her rating rationale (q3).

\subsubsection{Experiment Apparatus and Measures}
% We employed objective measures of human perception and opinion, where appropriate, to mitigate poor estimation of perceptions
% % Although self-reported ratings are common in human-subjects studies of system and explanation usage \cite{Simonyan15, ross2017right}, participants may poorly estimate their perceptions 
% ~\cite{angel1988use, avrahami2007biases, goldstein2014lay}. Specifically, 
We used a “grid selection” user interface (UI) to measure objective truthfulness (Fig. \ref{fig:userStudy1Procedure}b) to mitigate poor estimation of perceptions~\cite{angel1988use, avrahami2007biases, goldstein2014lay}. 
% , and, in User Study 2 (described later), the “balls and bins” question \cite{goldstein2014lay} to elicit user labeling {\color{blue}(Fig. \ref{fig:uiBallsAndBins})}. 
%
It overlays a clickable grid on the image for selecting important cells regarding the label. For usability, we limited the grid to 5×5 cells that can be selected or unselected (binary values). In the surveys, we referred to CAMs as “heatmaps”, which is a more familiar term. 
% We define User-CAM as the participant’s grid selection response, and CAM as the heatmap shown. 
To compare the participant’s grid selection (User-CAM) with the heatmap shown (CAM), we aggregated CAM by averaging the pixel saliency in each cell and calculated \textbf{CAM Truthfulness Selection Similarity} as the Pearson’s Correlation Coefficient (PCC) between User-CAM and CAM. 
We also measured the \textbf{CAM Truthfulness Rating} as a subjective, self-reported rating on a uni-polar 10-point star scale (1 to 10). We collected the rationale of ratings as open-ended text.
We measured the task time (per trial) as \textbf{Task Time Level} as low ($<$33 percentile), high ($>$66), medium, to account for response thoughtfulness.
We tracked the \textbf{Image Label} of each image, since some types are easier to recognize even if blurred.

\subsubsection{Participants}
We recruited 36 participants from Amazon Mechanical Turk (AMT) with high qualification ($\geq5000$ completed HITs with \textgreater 97\% approval rate). 
32 participants passed screening, and completed the survey in a median time of 15.9 minutes and were compensated US\$2.00.
They were 41.7\% female and 23-69 years old (Median = 35).

\subsubsection{Statistical Analysis and Quantitative Results}
% {\color{blue}We excluded 40/320 responses from analysis based on the exclusion criterion of taking \textgreater 200 seconds to complete each page per trial.} 
For all dependent variables, we fit a multivariate linear mixed effects model with Blur Bias Level, CAM Types, Image Label and Task Time Level as fixed effects, Blur Bias Level × CAM Type, Image Label × Blur Bias Level, Image Label × CAM Type, Task Time Level × Blur Bias Level and Task Time Level × CAM Type as fixed interaction effects, and Participant as a random effect. 

\begin{figure}[!t]
    \centering    
    \vspace{-0.1cm}
    \includegraphics[width=8.1cm]{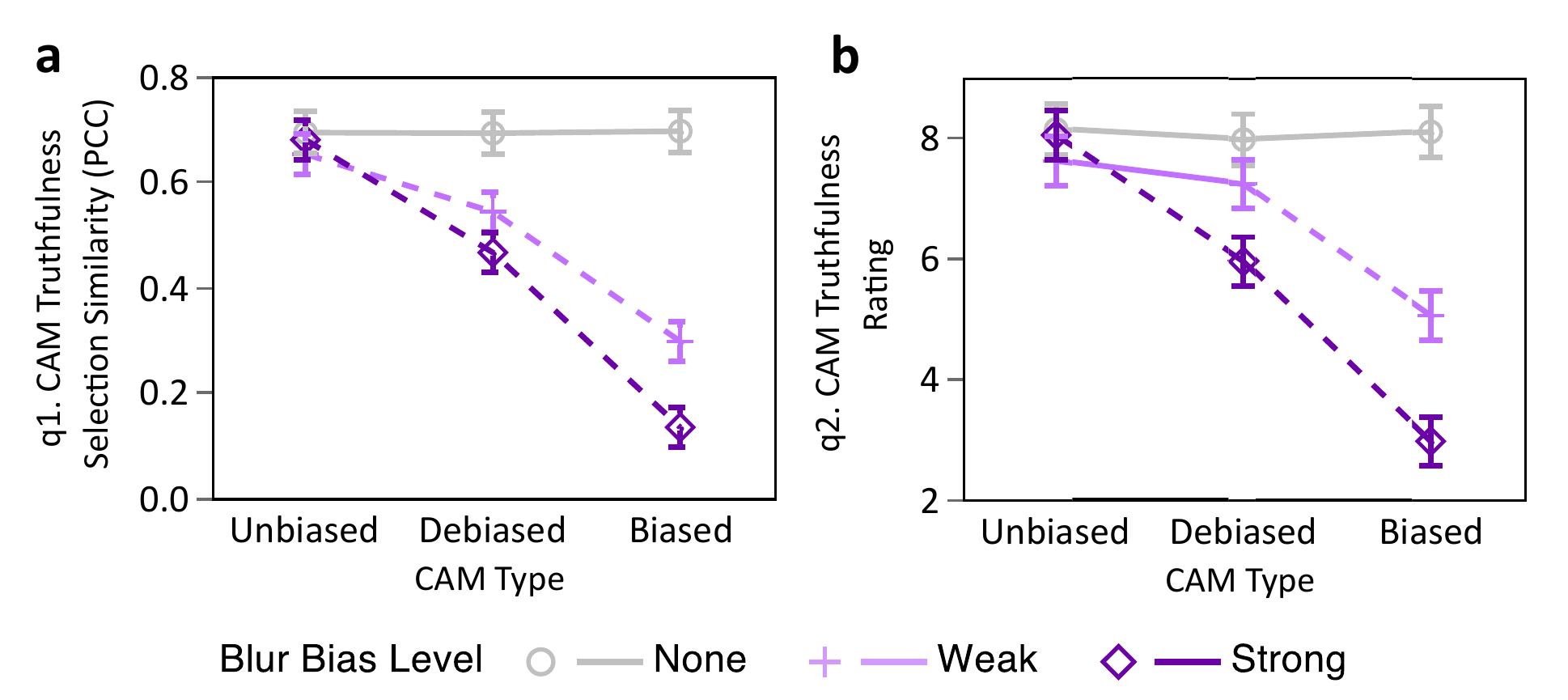}
    \vspace{-0.2cm}
    \caption{User Study 1 results.
    CAM Truthfulness decreased with blur, but was improved with Debiased-CAM.
    % both a) Selection Similarity (PCC) and b) Perceived Rating. 
    Dotted lines indicate very significant $p<.0001$ comparisons; solid lines indicate no significance at $p>.01$. Error bars indicate 90\% confidence interval.
    }
    \label{fig:userStudy1Plots}
    \Description{
    a) Barchart of CAM Truthfulness (PCC via grid selection) for 3 CAM conditions and 3 blur bias levels. Generally, None Blur > Weak Blur > Strong Blur, while Unbiased CAM > Debiased CAM > Biased CAM.
    b) Barchart of CAM Truthfulness rating for 3 CAM conditions and 3 blur bias levels. Generally, None Blur > Weak Blur > Strong Blur, while Unbiased CAM > Debiased CAM > Biased CAM.
    }
    \vspace{-0.4cm}
\end{figure}

Supplementary Table \ref{table:table-study1ModelDetails} reports the model fit ($R^2$) and significance of ANOVA tests for each fixed effect. 
Due to the large number of comparisons in our analysis, we consider differences with $p<.001$ as significant. This is sufficiently strict for a Bonferroni correction for 50 comparisons ($\alpha$ = .05/50). Furthermore, all results reported were significant at $p<.0001$, unless otherwise stated. We performed post-hoc contrast tests for specific differences described. All statistical analyses were performed using JMP (v14.1.0).

% Fig. \ref{fig:userStudy1Plots} shows results of the statistical analyses on {\color{blue}320} trials from {\color{blue}32} participants recruited from Amazon Mechanical Turk with significant findings at $p<.0001$. 
% The distribution of computed CAM Faithfulness (PCC) for different CAM type and Blur Bias levels (Fig. \ref{fig:InstanceFaithfulness}) guided the hypotheses for the results. 
%
Fig. \ref{fig:userStudy1Plots} summarizes our results.
Unbiased-CAM had the highest CAM Truthfulness Selection Similarity, while Biased-CAM the lowest Similarity that was only 21.3-43.7\% of the truthfulness of Unbiased-CAM. Debiased-CAM had significantly higher CAM Truthfulness Selection Similarity than Biased-CAM at 69.4-79.0\% of the truthfulness of Unbiased-CAM. Similarly, for blurred images, participants rated Unbiased-CAM as the most truthful (M = 7.83 out of 10, standard error = 0.12), followed by Debiased-CAM (M = 6.00 $\pm 0.21$ to 7.21 $\pm 0.18$), and Biased-CAM as the least truthful (M = 3.05 $\pm 0.21$ to 4.98 $\pm 0.26$). 
% Our qualitative analysis of participant rating rationales found that Debiased-CAM and Unbiased-CAM were rated as more truthful because they 1) highlighted semantically relevant targets while avoiding irrelevant ones, 2) did not highlight regions that were too narrow or wide for expected objects in the domain, and 3) had accurate shapes and edge boundaries for salient regions. Detailed quotes and interpretations in supplementary materials.
In summary, Debiased-CAM improved CAM truthfulness, despite stronger blur that reduced CAM truthfulness by highlighting wrong or unexpected regions, sizes, and shapes.

\subsubsection{Thematic Analysis and Qualitative Findings}
We analyzed the rationale of participant ratings to better understand how participants interpreted different CAMs as truthful or untruthful, and what visual features they perceived in images and CAMs. We performed a thematic analysis with open coding \cite{muller2010grounded}. Two authors independently coded the rationales and discussed the coding until themes converged. Next, we first describe rationales for different blur levels, then describe themes spanning all blur levels. Note that all CAM types were shown anonymously (labeled A, B, and C) with randomly orders; we quote them specifically by type for clarity. 
% Supplementary Fig. \ref{sfig:userStudyInstances1} shows the images and CAMs that participants viewed.

For None blur, as expected, most participants perceived CAMs as identical, e.g., \textit{“all 3 images are the same and mostly representative”} (Participant P23, “Fish” image); though some participants could perceive the slight decrease in the CAM truthfulness of Debiased-CAM, e.g., for the “Church” image, P1 wrote that Unbiased-CAM and Biased-CAM \textit{“had the most focus on *all* the crosses on the roof of the church and therefore I thought they were the most representative. [Debiased-CAM] gives less importance to the leftmost cross on the roof and therefore was rated lower.”} For Weak blur, participants felt Unbiased-CAM was very truthful, Debiased-CAM was slightly less truthful, and Biased-CAM was untruthful; e.g., P29 felt that Biased-CAM \textit{“doesn't show anything but blackness, [other CAMs] are much better in the way the heatmap shows details.”} For Strong blur, participants perceived Debiased-CAM as moderately truthful, but Biased-CAM as very untruthful, e.g., P18 felt that \textit{“[Biased-CAM] is totally off, nothing there is a garbage truck. [Unbiased-CAM] shows the best and biggest area, and [Debiased-CAM] is good too but I'm thinking not good enough as [Unbiased-CAM].”}

Across blur conditions, we found that participants interpreted whether a CAM was truthful based on several criteria — primary object, object parts, irrelevant object, coverage span, and shape. Participants checked whether the primary object in the label was highlighted (e.g., \textit{“That heatmap that focuses on the chainsaw itself is the most representative.”} P20, Chain Saw), and also checked whether specific parts of the primary object were included in the highlights (e.g., \textit{“[Unbiased-CAM and Debiased-CAM] correctly identify the fish though [Unbiased-CAM] also gives importance to the fish's rear fin.”} P1, Fish, Weak blur). P15 noted differences between the CAMs for the \textit{“French Horn” image: “[Unbiased-CAM] places the emphasis over the unique body of the French horn, and it places more well-defined, yellow and green emphasis on the mouthpiece and the opening of the horn itself. [Biased-CAM] is too vertical to completely capture the whole horn, and [Debiased-CAM]’s red area is too small to capture the body of the horn, and does not capture the opening of the horn or the mouthpiece.”} Participants rated a CAM as less truthful if it highlighted irrelevant objects, e.g., \textit{“[Debiased-CAM] is quite close to capturing the entire church. (But) [Unbiased-CAM] captures more of the tree.”} (P26, Church). Much discussion also focused on the coverage of salient pixels. Less truthful CAMs had coverages that were either too wide (e.g., \textit{“[Debiased and Biased CAMs] are inaccurate. They are too wide.”} P22, Garbage Truck), covering the background or other objects to get \textit{“less representative when it misleads you into the background or surroundings of the focus. It needs to only emphasize the critical area.”} (P23, Church); or too narrow, not covering enough of the key object such that it \textit{“is very small and does not highlight the important part of the image. It is too narrow.”} (P30, Fish). Finally, participants appreciated CAMs that highlighted the correct shape of the primary object, e.g., \textit{“[Debiased-CAM] perfectly captures the shape of the ball and all of its quadrants. [Unbiased-CAM] is a little more oblong than the golf ball itself, so it's not as perfect. [Biased-CAM] is almost a vertical red spot and does not really capture the shape of the golf ball at all.”} (P15, Golf Ball).

In summary, we found that Debiased-CAM and Unbiased-CAM were perceived as truthful, because they: 1) highlighted semantically relevant targets while avoiding irrelevant ones, so concept or object-aware CNN models are important \cite{bau2017network, kim2018interpretability}; 2) had salient regions that were neither too wide nor narrow for the image domain; and 3) had accurate shape and edge boundaries for salient regions, which can be obtained from gradient explanations \cite{rudin2019stop}.

\subsection{User Study 2 (CAM Helpfulness)}
% Having shown that Debiased-CAM was perceived as more truthful than Biased-CAM, we next investigated how helpful Debiased-CAM was to verify predictions of blur biased images. 
% We recruited participants to view one of 3 CAM types of images at 3 blur levels in a 3×3 factorial within-subjects experiment. 
% Having validated the perceived truthfulness for each CAM type, we next evaluated their helpfulness to verify predictions of blur biased images.
The second study evaluated the perceived helpfulness of each CAM type to verify predictions of blur biased images.

\begin{figure}[ht]
    \centering    
    \vspace{-0.1cm}
    \includegraphics[width=8.4cm]{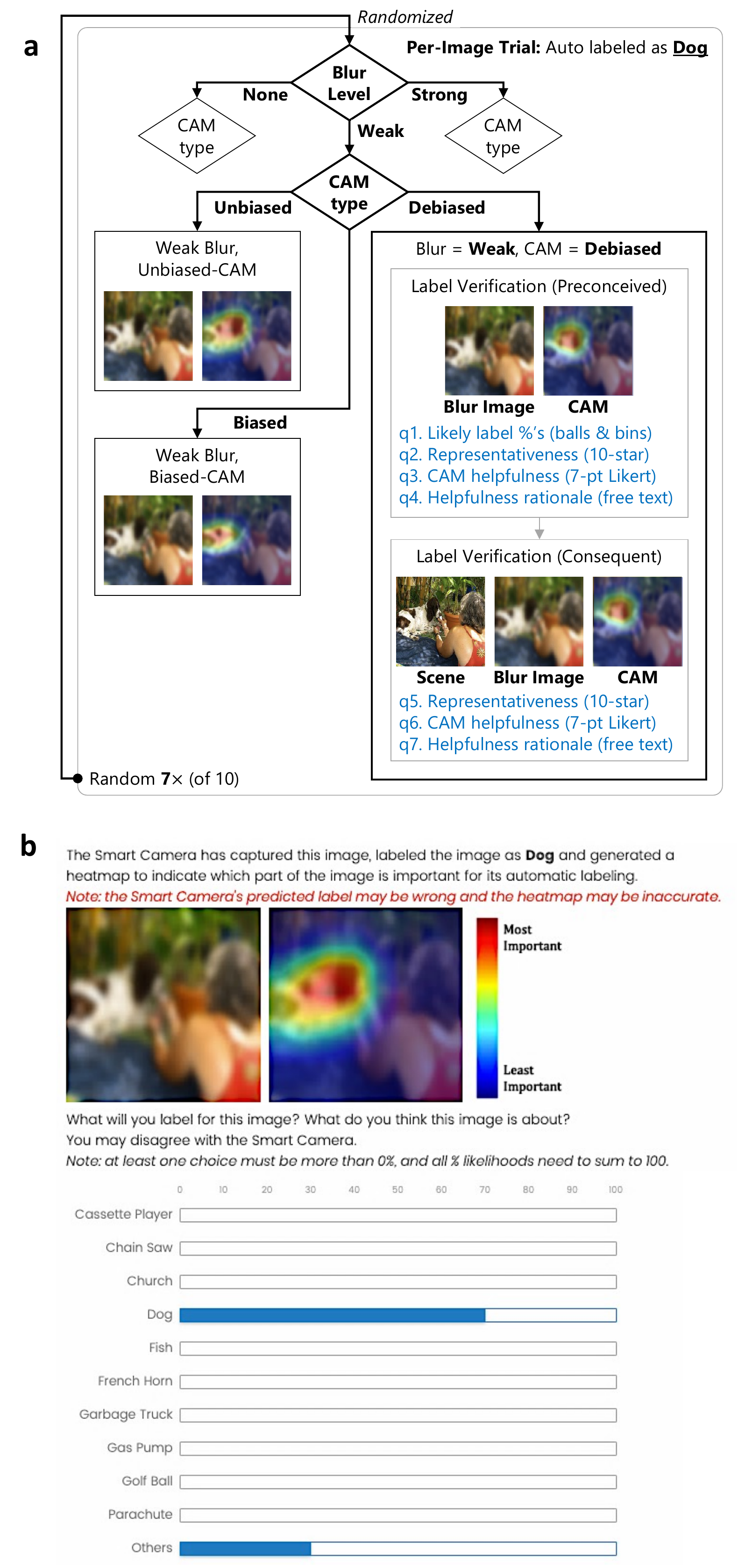}
    \vspace{-0.2cm}
    \caption{
    User Study 2 main study procedure (a) and “balls and bins” UI to elicit user labeling (b).
    }
    \label{fig:userStudy2Procedure}
    \Description{
    a) The flowchart of User study 2. Each participant will be randomly allocated to 1 out of 3 blur-level conditions and 1 out of 3 CAM conditions. In each condition, the participant will view an blurred image (e.g. a human play with dog), and view 1 Grad-CAM explanation. She has to label the image from candidate options, rate the representativeness and helpfulness of the CAM and indicate rationale of rating. After that, the participant will be expose to the unblurred version of the image and answer the questions again. 
    b) The UI of balls and bins question: The text description of the question come first, followed by an blurred image and the corresponding CAM explanation. The user is asked to elicit the image label by dragging the slider on a group of possible options.
    }
    \vspace{-0.5cm}
\end{figure}

\subsubsection{Experiment Procedure}
The procedure is the same as User Study 1, except for the main study section.
% To investigate the helpfulness of CAMs on blurred images, we modified the sequence in the survey of User Study 1. 
User Study 1 focused on CAM Truthfulness to obtain the participant’s saliency annotation of the unblurred image before revealing CAMs. 
In User Study 2, showing the unblurred image first will invalidate the use case of verifying predictions on blurred images, since the participant would have foreknowledge of the image. Hence, participants needed to see the blurred image and model prediction first, answer perception questions, then see the  image unblurred. 
% Fig. \ref{fig:userStudy2Procedure}a illustrates the experiment procedure of User Study 2. 
% Participants began with the same procedure as User Study 1, including the same introduction, tutorial, screening quiz, and background and demographics questions, but experienced a different main study section. 

In the main study (Fig. \ref{fig:userStudy2Procedure}a), each participant viewed 7 repeated image trials, each randomly assigned to one of 9 conditions (3 Blur Bias levels × 3 CAM types) in a within-subjects experiment design. Participants viewed 7 randomly chosen images from the same 10 images of User Study 1, instead of all 10, so that they could not easily conclude the class label for the remaining images by eliminating previous classes. For each trial, the participant performed the common explainable AI task to verify the label prediction of the model. 
On the first page, the participant viewed a labeled image at the assigned Blur Bias level with corresponding CAM for the assigned CAM type, indicated her likelihood choice(s) for the image label with the “balls and bins” question \cite{goldstein2014lay} to elicit user labeling (Fig. \ref{fig:userStudy2Procedure}b) (q1); rated how well each CAM represented the image label (q2); rated how helpful the CAM was for verifying the label (q3), and wrote the rationale for her rating (q4). 
% Supplementary Fig. \ref{sfig:userStudy2_Trial}a shows the first questionnaire page {\color{blue}(Preconceived)}. 
On the next page, participants saw the image unblurred and answered questions q2-4 again as questions q5-7. 
See Supplementary Fig. \ref{sfig:userStudy2_Trial} for questionnaire details.
% This allowed us to compare preconceived and consequent ratings and rationale (Fig. \ref{fig:userStudyPlots_preconceivedVsConsequent}).

\subsubsection{Experiment Apparatus and Measures}
For q1, we asked the participant to indicate likelihoods of 10 possible image labels with the “balls and bins” question \cite{goldstein2014lay, sharpe2000distribution, delavande2008eliciting} to elicit her probability distribution $\mathbf{p}=\left\{p_c\right\}^T$ over label classes $c\in C$. This question is reliable in eliciting probabilities from lay users \cite{goldstein2014lay, sharpe2000distribution} and avoids priming participants with the actual label $c_0$, since it asks about all labels. We calculated the participant’s selected label $\acute{c}$ as the class with the highest probability, i.e., $\acute{c}={\rm argmax}_c{\left(p_c\right)}$, \textbf{Labeling Confidence} as the indicated likelihood for the actual label $p_{c_0}$, and \textbf{Label Correctness} as $\left[\acute{c}=c_0\right]$, where $\left[\cdot\right]$ is the Iverson bracket notation. 
We measured the perceived \textbf{CAM Helpfulness} and \textbf{CAM Truthfulness Ratings} on a bipolar 7-point Likert scale (–3 = Strongly Disagree, 
% –2 = Disagree, –1 = Somewhat Disagree, 0 = Neither Agree nor Disagree, +1 = Somewhat Agree, +2 = Agree, 
+3 = Strongly Agree). We collected rating rationale as open-ended text. We used different formats for CAM Truthfulness and CAM Helpfulness to mitigate repetitive or copied responses and to allow for a more precise measurement of CAM Truthfulness.
We also measured \textbf{Task Time Level} and \textbf{Image Label} per trial.

\subsubsection{Participants}
We recruited 191 new participants from AMT with the same qualification criteria as User Study 1. 
162 participants passed screening, completed the survey in a median time of 18.4 minutes and were compensated US\$2.00. 
They were 46.0\% female and between 21 and 74 years old (Median = 37).
We excluded 7 participants who gave wrong labels
% \footnote{i.e., the participant’s label with the highest probability was not the actual label; in practice, only 1 mistake allowed} 
for >60\% of encountered unblurred images, which indicated the participant's poor recognition ability.

\subsubsection{Statistical Analysis and Quantitative Results}

For each dependent variable, we fit a multivariate linear mixed effects model with the same fixed, interaction, and random effects as in User Study 1.
We further analyzed CAM Truthfulness and Helpfulness ratings with fixed main and interaction effects regarding whether users rated before or after seeing the unblurred version of the image, i.e., Unblurred Disclosure: preconceived or consequent. Supplementary Table \ref{table:table-study2ModelDetails} reports the model fit ($R^2$) and ANOVA tests for each fixed effect, and report significant results similarly to User Study 1.

\begin{figure}[!ht]
    \centering   
    % \hspace*{-0.2cm}
    \includegraphics[width=8.5cm]{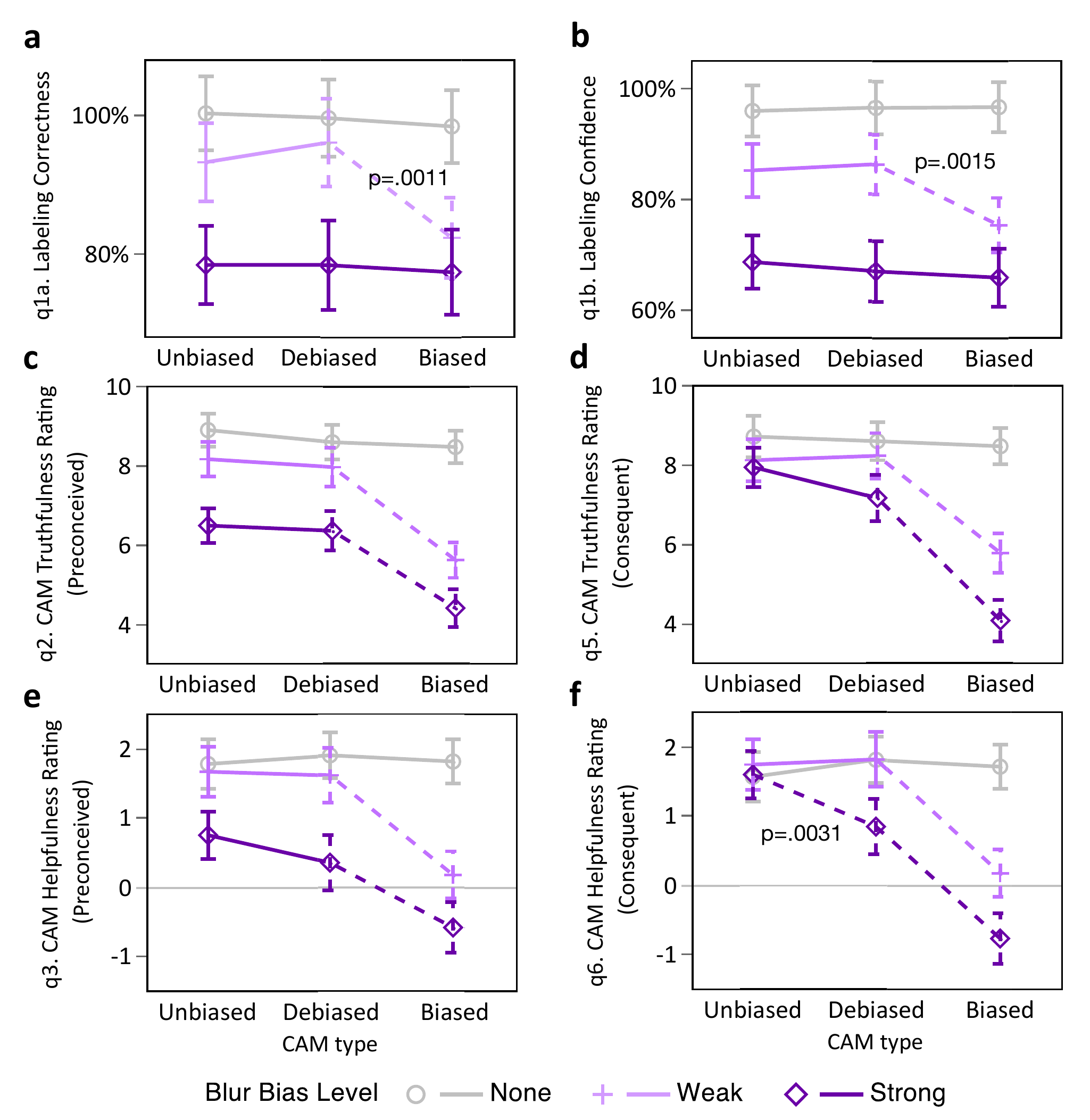}
    \vspace{-0.5cm}
    \caption{
    Quantitative results for User Study 2.
    Decision correctness (a,b) and perceived ratings (c-f) decreased with stronger blur, but Debiased-CAM improved them to be similar as with Unbiased-CAM.
    }
    \label{fig:userStudy2Plots}
    \vspace{-0.3cm}
    \Description{
    a) Barchart of labeling correctness for 3 CAM conditions and 3 blur bias levels. Generally, None Blur > Weak Blur > Strong Blur. For weak blur, Unbiased CAM ~= Debiased CAM > Biased CAM.
    b) Barchart of labeling confidence for 3 CAM conditions and 3 blur bias levels. Generally, None Blur > Weak Blur > Strong Blur. For weak blur, Unbiased CAM ~= Debiased CAM > Biased CAM.
    c) Barchart of preconceived CAM truthfulness rating for 3 CAM conditions and 3 blur bias levels. Generally, None Blur > Weak Blur > Strong Blur. For strong and weak blur, Unbiased CAM ~= Debiased CAM > Biased CAM.
    d) Barchart of consequent CAM truthfulness rating for 3 CAM conditions and 3 blur bias levels. Generally, None Blur > Weak Blur > Strong Blur. For strong and weak blur, Unbiased CAM ~= Debiased CAM > Biased CAM.   
    e) Barchart of preconceived CAM helpfulness rating for 3 CAM conditions and 3 blur bias levels. Generally, None Blur > Weak Blur > Strong Blur. For strong and weak blur, Unbiased CAM ~= Debiased CAM > Biased CAM.
    f) Barchart of preconceived CAM helpfulness rating for 3 CAM conditions and 3 blur bias levels. Generally, None Blur > Weak Blur > Strong Blur. For strong blur, Unbiased CAM ~= Debiased CAM > Biased CAM. For weak blur, Unbiased CAM > Debiased CAM > Biased CAM.
    }
\end{figure}

Fig. \ref{fig:userStudy2Plots} summarizes our results from 1,085 trials of 155 included participants. Differences in decision quality (Labeling Correctness and Labeling Confidence) across CAM types depended on blur bias level. For None blur, the decision quality was high for all CAM types (confidence M = 95.6\% $\pm 0.6\%$, correctness M = 99.0\% $\pm 0.5\%$) due to the ease of the tasks, while for Strong blur, the decision quality was low for all CAM types (confidence M = 68.5\%$\pm 1.8\%$, correctness M = 79.9\%$\pm 2.2\%$), suggesting that blurring was too strong even for truthful CAMs to be useful. However, for Weak blur, Debiased-CAM reduced labeling error by 1.92x (1 – Correctness: from 18.2\% $\pm 3.5\%$ to 9.5\% $\pm 2.9\%$) and improved confidence from 75.4\% $\pm 2.9\%$ to 82.8\% $\pm 2.7\%$ compared to Biased-CAM. We found stronger differences in preconceived ratings of CAM types. For Weak blur, participants rated Debiased-CAM as more truthful (M = 7.7 $\pm 0.2$ vs. 5.6 $\pm 0.3$ out of 10) and more helpful (M = 1.56 $\pm 0.13$ vs. 0.15 $\pm 0.19$ on a 7-point Likert scale from –3 to 3) than Biased-CAM. Moreover, for Strong blur, although their decision quality did not improve, participants perceived Debiased-CAM as more truthful (M = 6.4 $\pm 0.2$ vs. 4.4 $\pm 0.3$) and helpful (M = 0.60 $\pm 0.16$ vs. –0.49 $\pm 0.19$) than Biased-CAM. These effects were similar and slightly amplified for consequent ratings (Fig. \ref{fig:userStudy2Plots}), indicating that users more strongly appreciated Debiased-CAM and disliked Biased-CAM if they had foreknowledge of the unblurred scenes. 
In summary, Debiased-CAM recovered the usefulness of CAMs for moderately blurred images, and were perceived as helpful even for strong blur.

\subsubsection{Thematic Analysis and Qualitative Findings}
To understand why participants rated CAMs as helpful or unhelpful, 
% we analyzed the rationale of both their preconceived ratings when seeing the blurred image and consequent ratings after seeing the unblurred image scene. We 
we performed a thematic analysis on rationales, similarly to User Study 1. 
% These results elucidate the mental model of how truthful and debiased CAMs were useful even for blurred images. 
Rationale depended much on image Blur Bias level, and we identified how truthful and helpful debiased CAMs were even for blurred images.

For unblurred images (None blur level), participants mostly felt that CAMs were helpful, because CAMs helped to: 
1) focus their attention \textit{“on the most important part of the image, which helps me to quickly identify and label the image.”} (Participant P106, Garbage Truck); 
2) ignore irrelevant targets to \textit{“let me know I can disregard the person in the foreground”} (P89, Dog), \textit{“It helps hone in on what the content is, and helps to ignore the extra things in the frame.”} (P14, Chain Saw); 
3) matched their expectations since CAMs \textit{“did a solid job of identifying the garbage truck.”} (P36) and was \textit{“highly correlated to where the fish is in this image.”} (P38). Conversely, as expected, many participants considered CAMs unhelpful since \textit{“I could easily identify the object in the image without the heatmap”} (P32, Church).

For images with Weak blur, a truthful CAM: 
4) \textit{“helps focus my attention to that area on the blurry picture”} (P105, Debiased-CAM), \textit{“clearly give hint on what was needed to notice in the photo”} (P140, Unbiased-CAM); and 
5) helped to confirm image labels, e.g., P3 felt that \textit{“the heatmap gives me the idea that the object might be a fish, I could not tell otherwise”} and wrote after seeing the unblurred image that \textit{“I wouldn't have known what the object was without the heatmap.”} P118 described how Unbiased-CAM \textit{“pointed to the steeple and it helped me realize that it was indeed a picture of a church. I had trouble recognizing it on my own.”} Debiased-CAMs helped to locate suspected objects in unexpected images, e.g., P96 felt that \textit{“based on what the heatmap is marking, that's the exact spot where someone would hold a french horn”}, and P67 noted \textit{“that is not an area where I would expect to find a fish, so it's helpful to have this guide.”}

For images with Strong blur, many participants felt that the CAMs were very unhelpful, because 
6) the task was too difficult such that they had \textit{“NO idea what image is and heatmap doesn't help.”} (P68, Biased-CAM), felt the task \textit{“was very hard, i could not figure it out”} (P71, Debiased-CAM), did not have much initial trust as \textit{“I feel that the heatmap could be wrong because of the clarity of the image.”} (P62, Unbiased-CAM). Some participants would 
7) blindly trust the CAM due to a lack of other information such that \textit{“without the heatmap and the suggestion, I would have no guess for what this is. I am flying a bit blind. So, I concur with the recommendation (french horn) until I see more.”} (P92, Unbiased-CAM) and due to the trustful expectation that CAM \textit{“enables me to know the most useful part in the camera.”} (P138, Church, Unbiased-CAM). 
Finally, we found that 8) confirmation bias may cause the CAM correctness to be misjudged. For example, P76 first thought a misleading Biased-CAM \textit{“helps make a blurry picture more clear”}, but later realized \textit{“it's in the wrong spot.”} (“Garbage Truck”); in contrast, P24 wrongly accused that an Unbiased-CAM \textit{“was focused on the wrong thing”}, but changed his opinion after seeing the unblurred image, admitting \textit{“Now that I see it's a dog, it is more clear.”}

In summary, these findings explain why truthful Debiased-CAM and Unbiased-CAM helped participants to verify classifications of unblurred or weakly blurred images. For unblurred images, these CAMs: 1) focused user attention to relevant objects to speed up verification, 2) averted attention from irrelevant targets to simplify decision making, and 3) matched user expectations \cite{ross2017right} of the target object shapes. For weakly blurred images, these CAMs: 4) provided hints on which parts to study in blurred images, and 5) supported hypothesis formation and confirmation \cite{wang2019designing, sharpe2000distribution} of suspected objects. For strongly blurred images, participants generally rated all CAMs as unhelpful because: 6) verifying the images was too difficult, 7) they felt misguided to blindly trust CAMs, and 8) they misjudged CAMs based on preconceived notions, i.e., confirmation bias \cite{wang2019designing}.

\section{Discussion and Design Implications}
Our results highlighted issues in explanation faithfulness when CNN models explain their predictions on images subjected to systematic error bias, which
we addressed by with Debiased-CAM to improve explanation truthfulness and helpfulness, and consequently the prediction performance. 
We discuss implications for XAI and HCI researchers and practitioners to: 
1) be wary of how contexts and corruptions can make explanations misleading,
2) support scalable human-centric explanations,
3) extend debiasing to social contexts, and
4) carefully design unconfounded user studies to evaluate XAI.
We also discuss 5) generalizations of our debiasing method to other XAI techniques and data types.

\vspace{-0.1cm} 
\subsection{Physical contextual bias in explanations}
Most AI explanations have been developed to support model debugging, help end-users identify incorrect model reasoning, or trust correct explanations. 
However, we have shown that moderate systematic error (biases) in data, which may seem innocuous, can lead to severe deviations in explanations, despite model fine-tuning. For baseline models, these truly reflect the model reasoning, but they can be incongruent with user expectations, and can harm user trust.

We have investigated three prevalent sources of bias --- blur, color, lighting --- that can plausibly occur in image applications, and were feasibly manipulable to test in evaluations. We showed that CAM deviations are significant in such cases, yet Debiased-CAM can improve them. 
In pilot studies, we have also investigated other bias types listed in~\cite{hendrycks2019benchmarking}. 
We found that environmental biases (e.g., snow, fog, frost) cause moderate CAM deviations that will require debiasing. 
Some biases had very weak CAM deviations (e.g., brightness, contrast), since the image pixel changes were monotonic which does not affect the additive activations in neural networks much. Debiasing may not be needed for such biases.
Finally, we found that biases due to image processing and compression artifacts (e.g., Gaussian noise, JPEG) had small CAM deviations.
We expect other blur biases (e.g., defocus, frosted glass, motion, zoom) to have slightly stronger, but similar effects as Gaussian blur that we evaluated, since the images remain similar to the original. 

However, images subjected to adversarial noise~\cite{dombrowski2019explanations} would be particularly concerning, since an attacker can intelligently and maliciously inject noise to deliberately harm the performance or explanation, such that CAM deviations will be worse. Training Debiased-CAMs under such attacks may be more difficult.

\vspace{-0.1cm} 
\subsection{Scalable human-centric explanations}
The needs for model explainability are diverse. Langer et al. cataloged many desiderata of explanations~\cite{langer2021we}, including several societal objectives such as agreeability, auditability, fairness and privacy.
To support human interpretability at the cognitive level, explanations need to conform to human prior knowledge~\cite{erion2021improving,lage2018human,ross2017right,selvaraju2019taking}, human reasoning processes~\cite{miller2019explanation,wang2019designing}, and human perceptual processes~\cite{zhang2022towards}.
In this work, we focus on improving the \textit{agreeability} of explanations towards human prior knowledge. 
This typically requires manual inspection and annotation by people~\cite{ross2017right,lage2019evaluation}, which is labor-intensive.
Instead, given a clean dataset with agreeable explanations, we train our model to produce debiased explanations. This uses self-supervision, so it is also scalable and not labor-intensive. Users may only need to select images rather than annotate details in each image. For our studies, we had assumed that unbiased images have reliable explanations, but this should be verified by human labelers. An interface to support quick ratings of explanation acceptability would help to accelerate this data curation.
Another scalable strategy involves defining axioms (e.g., attribution priors~\cite{erion2021improving}, psychological preferences~\cite{wang2021show}, or visual cognitive chunks~\cite{abdul2020cogam}) and constraining explanations towards them.
From our qualitative analysis, we identified desiderata for truthful saliency maps (e.g., trace the shape of relevant objects, control the spread or tightness of hot spots) that can be used as general axioms for faithful saliency maps.
This further increases scalability by reducing the dependency on selecting reliable explanation references.

\vspace{-0.1cm} 
\subsection{Debiasing explanations against social bias}
Although we have focused on bias due to physical contexts, bias in social situations also needs debiasing.
People are subjected to \textit{egocentric bias}~\cite{konow2005blind} and \textit{societal discrimination (unfairness)}~\cite{dodge2019explaining}.
With egocentric bias, different stakeholders would prioritize their own objectives~\cite{ehsan2021expanding} and may be ignorant of other viewpoints. Debiased explanations could encode different interpretation preferences (e.g., \cite{lage2018human,wang2021show}) to show how slightly different two stakeholders interpret a decision (e.g., patient and doctor for medical diagnosis).
With social bias, models may predict or reason undesirably for some protected groups of people based on sensitive attributes (race, gender, etc.). For example, saliency maps can detect bias in a model by highlighting a female face for Nurse, but highlighting a stethoscope held by a woman for Doctor~\cite{selvaraju2017grad}. Instead of the current approach to debias models with data balancing, our debias approach can retrain models to de-emphasize focusing on sensitive concepts (e.g., faces).
However, we caution about the dark pattern of debiasing explanations to make an unfair model appear fair by retraining its explanation to appear fair (e.g., ~\cite{dombrowski2019explanations,Dombrowski2022TowardsRE}).

\vspace{-0.1cm} 
\subsection{Sensitive measures for faithful explanation}
Saliency map explanations have mostly been evaluated with simulation metrics and rarely with human subjects~\cite{alqaraawi2020evaluating,kittley2019evaluating}. User studies are important to verify the severity of problems (\textit{perceptually noticeable enough?}) and the efficacy of solutions (\textit{problem no longer perceivable or perceptually forgivable?}). 
% Through our controlled experiments, we validated the severity of CAM deviations, and showed that differences between Unbiased- and Debiased-CAMs are not significant, and from qualitative analysis showed that Debiased-CAMs have acceptable margin of error. 
%
However, designing successful experiments with strong effects and sensitive measures is difficult and many studies fail to find effects~\cite{bansal2021does,kittley2019evaluating,poursabzi2021manipulating}.
To improve the sensitivity, experiments need more sensitive measures and carefully designed participant tasks.

Current user studies use simple true/false or multiple choice responses and confidence ratings, but these measures are prone to lucky guesses, do not capture secondary choices, or suffer from social desirability bias. 
The insensitivity of such methods could have led to null results~\cite{alqaraawi2020evaluating,kittley2019evaluating}.
Instead, we employed more sensitive and objective measures of labeling likelihood ("balls and bins" question).
%
% Instead of asking for ratings subjectively, it is also important to objectively measure explanation agreement, since users tend to over-trust~\cite{Kaur2020InterpretingIU} even with wrong explanations.
We also measured explanation agreement objectively, since users tend to over-trust wrong explanations~\cite{Kaur2020InterpretingIU}, affecting the validity of subjective ratings.
Hence, we employed the grid-selection UI, which is similar to segment selection in~\cite{zhang2019dissonance}.
Another method is to ask participants to write important and ignored features in the free text~\cite{alqaraawi2020evaluating}, but this is difficult to automatically evaluate.

Explanation understanding is typically evaluated with human simulatability tasks~\cite{lim2009and,lipton2018mythos}, where users try to predict what a model would predict. However, participant answers may be confounded by leaking information that participants are tested on.
Zhang et al.~\cite{zhang2019dissonance} evaluated saliency using a reverse-ablation method to incrementally reveal important segments and ask participants to label the image; this avoids the hindsight bias effect~\cite{Roese2012HindsightB}.
In this work, we controlled when to pose questions.
To measure perceived truthfulness, we first measured objective ground truth before showing CAMs to avoid participants copying or being primed. To measure perceived helpfulness in a privacy-preserving application, we posed questions twice, first with blurred images, then unblurred images. This mitigated the hindsight bias effect.
Thus, we add sensitive experiment apparatuses to the literature on evaluating XAI.

% {\color{blue} \textbf{Generalization to other XAI techniques.}} 
\vspace{-0.1cm} 
\subsection{Generalization to other XAI and data types}
Our self-supervised debiasing can apply to other gradient-based explanations \cite{bach2015pixel, simonyan2014deep, sundararajan2017axiomatic} by formulating the activation, gradient or propagated terms as a secondary prediction task. 
However, some saliency explanations, such as Layer-wise Relevance Propagation (LRP) \cite{bach2015pixel} and Integrated Gradients \cite{sundararajan2017axiomatic}, which produce fine-grained “edge detector” heatmaps \cite{adebayo2018sanity} are likely to be more severely degraded with biasing, such as strong blurring. 
Beyond gradient-based explanations, model-agnostic explanations such as LIME \cite{ribeiro2016should} and Kernel SHAP \cite{lundberg2017unified} can be debiased by regularizing on a saliency loss metric. 
Notably, CNN explanation techniques such as feature visualizations \cite{bau2017network, olah2017feature} and neuron attention \cite{li2018tell} have higher dimensionality that requires more sensitivity to debias. Dimensionality reduction with autoencoders or generative adversarial networks (GANs) could provide latent features that are feasible to debias. 
Finally, concept-based explanations such as TCAV \cite{kim2018interpretability} and RexNet~\cite{zhang2022towards} can be debiased to align the generated concept with user expectations.

% {\color{blue} \textbf{Generalization to other bias and data types.}} 
Debiased-CAM can be generalized to other types of data subjected to bias,
particularly those that can be modeled with CNNs, such as audio and time series data.
%Training to debias against these can help to generate explanations that are more robust and interpretable for various applications.
% We have investigated deviations in CAM explanations due to three common image biases, Gaussian blurring, color shifting, and low lighting. 
% Other cases of biasing include images pixelated for privacy protection \cite{padilla2015visual}, noisy images under low light \cite{park2018dual}, ultrasound images \cite{cohen2002new}, and images with motion blur \cite{kupyn2018deblurgan}. Training to debias against these can help to generate explanations which are more robust and interpretable for more contexts of use. 
Other than biases in images, debiasing is also necessary for explaining model predictions of other data types and behaviors, such as audio signals with noise or obfuscation~\cite{mcloughlin2015robust}, and human activity recognition with inertial measurement units (IMU) or other wearable sensors~\cite{ryoo2017privacy}. 
With the prevalence of data bias in the real-world and privacy obfuscation, Debiased-CAM provides a generalizable framework to train robust performance and faithful explanations for responsible AI.

\section{Conclusion}
We highlight issues in explanation faithfulness when CNN models explain their predictions on images that are biased with systematic error, and address this by developing Debiased-CAM to improve the truthfulness of explanations. We achieved these improvements by ensuring that model parameters were learned based on more important attention as identified by unbiased explanations and on more diverse inputs due to data augmentation across multiple bias levels. We also implemented more precise training with multiple prediction tasks and differentiable explanation loss. Our results showed that even when image data were degraded or distorted due to bias, 1) they retained sufficient useful information that DebiasedCNN could learn to recover salient locations of unbiased explanations, and 2) these salient locations were highly relevant to the primary task such that prediction performance could be improved. % Furthermore, our multi-bias, multi-task debiasing training can improve model robustness.

% \section{Acknowledgments}
\begin{acks}
This work was supported in part by the Ministry of Education, Singapore under the grant T2EP20121-0040, and was carried out at the NUS Centre for Research in Privacy Technologies (N-CRiPT) and the NUS Institute for Health Innovation and Technology (iHealthtech).
\end{acks}

%%
%% The next two lines define the bibliography style to be used, and
%% the bibliography file.
\bibliographystyle{ACM-Reference-Format}
\bibliography{main}
\clearpage
%%
%% If your work has an appendix, this is the place to put it.

\appendix
\onecolumn
\captionsetup[figure]{labelfont={small},font={small},name={Supplementary Fig.},labelsep=period}
\captionsetup[table]{labelfont={small},font={small},name={Supplementary Table},labelsep=period}
\setcounter{figure}{0}
\setcounter{table}{0}

\section{Technical Approach Appendix}

\subsection{Datasets and model implementation details}
In simulation studies, we evaluated the models on three datasets for two image tasks (summarized in \ref{table:table-datasetDescription}). 
For Simulation Study 1 (Blur Bias), we used Inception v3 \cite{szegedy2016rethinking} pretrained on ImageNet ILSVRC-2012 \cite{deng2009imagenet} and fine-tuned on blur biased images of ImageNette \cite{imagenette}, which is a subset of ILSVRC-2012.
We only retrained layers from the last two Inception blocks of the Inception v3 model. 
For Simulation Studies 2 and 4 (Blur and Color Temperature Bias on egocentric activity images), we also used Inception v3 pretrained on ILSVRC-2012, and fine-tuned it on the NTCIR-12 \cite{gurrin2016ntcir}.
For Simulation Study 3 (Blur Bias Captioning), we used the Neural Image Captioner (NIC) \cite{vinyals2015show} with Inceptionv3-LSTM model and fine-tuned on blur biased images from COCO \cite{chen2015microsoft}.
We retrained the last two inception blocks of Inception v3 as well as LSTM blocks. 
For Simulation Study 5 (Lighting Bias), we fine-tuned the Inception v3 (pretrained on ILSVRC-2012) on the Transient Attribute database (TransAttr) \cite{laffont2014transient} for multi-label classification. We limited our evaluations to four labels: Snowy, Sunny, Cloudy, Dawn/Dusk.
All model hyperparameters were tuned using the Adam optimizer with batch size 64 and learning rate ${10}^{-5}$. 
% Supplementary material describes further studies with different base CNN models, VGG16 \cite{Simonyan15}, ResNet50 \cite{he2016deep} and Xception \cite{chollet2017xception}.

\begin{table}[h!]
    \centering
    \includegraphics[width=13cm]{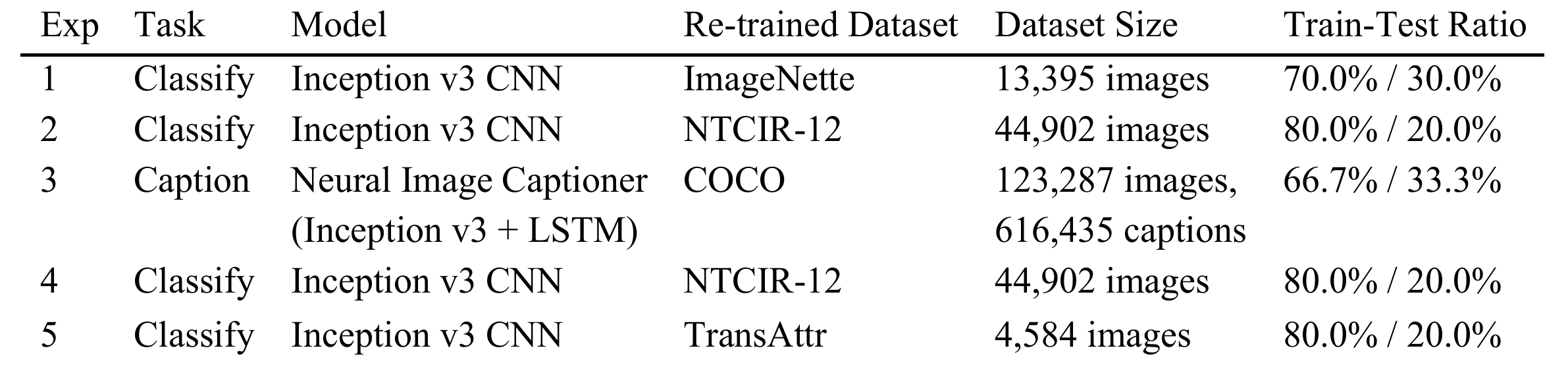}
    \caption{Baseline CNN models trained on training datasets for Simulation Studies. All models were pre-trained on ImageNet ILSVRC-2012 and retrained to fine-tune on respective datasets. Train-test ratios were determined from the original literature 
    % of the models 
    as referenced.} 
    \label{table:table-datasetDescription}
    \Description{
    From left to right, the table indicate the simulation experiment index, the primary prediction task, the backbone CNN model, the dataset used for re-training, the number of instances in dataset, and the split ratio of dataset. 
    }
\end{table}

\clearpage

% \subection{DebiasedCNN and FineTuneCNN Model Variants}
% \subsection{Supplementary Technical Details}
\subsection{Model Variants}

\quad % placeholder to fix the layout issue
\begin{figure}[h!]
    \centering    
    \includegraphics[width=14cm]{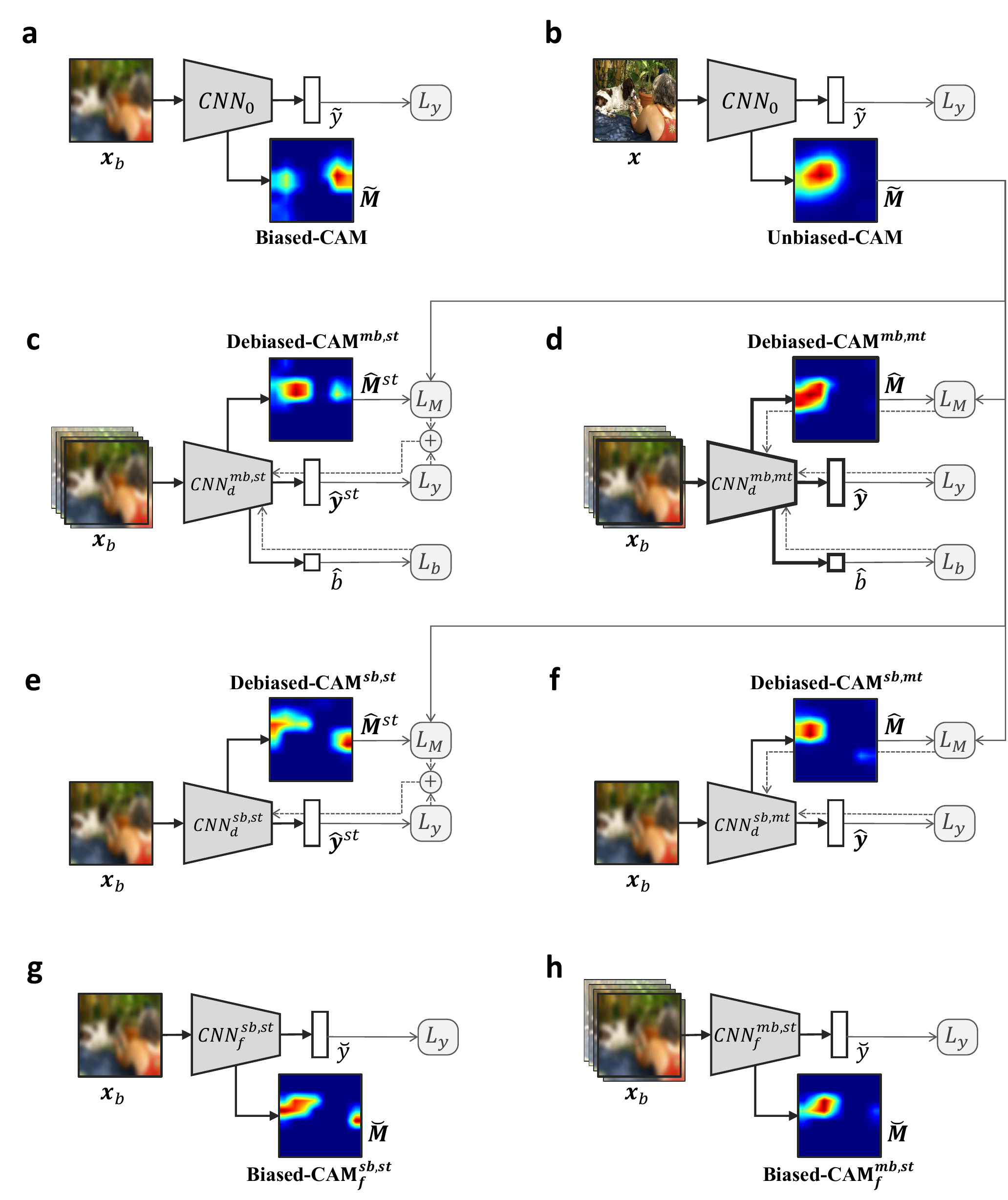}
    \caption{Architectures of self-supervised DebiasedCNN variants and of baseline CNN models and their CAM explanations from a biased “Dog” image blurred at $\sigma=24$. a) RegularCNN on biased image. b) RegularCNN on unbiased image. c) DebiasedCNN (mb, st) with single-task loss as a sum of classification and CAM losses for the classification task, trained on multi-bias images with auxiliary bias level prediction task. d) DebiasedCNN (sb, mt) with multi-task for CAM prediction trained with differentiable CAM loss, and trained on multi-bias images with auxiliary bias level prediction task. e) DebiasedCNN (sb, st) with single-task loss as a sum of classification and CAM losses for the classification task. f) DebiasedCNN (sb, mt) with multi-task for the CAM prediction and differentiable CAM loss. g) FineTunedCNN (sb,st) retrained on images biased at a single-bias level. h) FineTunedCNN (mb,st) retrained on images biased variously at multi-bias levels.}
    \label{sfig:modelVariants}
    \Description{
    Architectures of model variants.
    a) The biased image is fed into a CNN classifier trained on unbiased images to predict the label and generate the biased-CAM.  
    b) The unbiased image is fed into a CNN classifier trained on unbiased images to predict the label and generate the unbiased-CAM. The unbiased-CAM can be further used to train various debiased CNNs.
    c) biased images (from multiple bias levels) are fed into a CNN classifier to train the DebiasedCNN (multi-bias, single-task).
    d) biased images (from multiple bias levels) are fed into a CNN classifier to train the DebiasedCNN (multi-bias, multi-task).
    e) biased images (from a single bias level) are fed into a CNN classifier to train the DebiasedCNN (multi-bias, single-task).
    f) biased images (from a single bias level) are fed into a CNN classifier to train the DebiasedCNN (multi-bias, multi-task).
    g) biased images (from a single bias level) are fed into a CNN classifier to train the BiasedCNN (single-bias, single-task).
    h) biased images (from multiple bias levels) are fed into a CNN classifier to train the BiasedCNN (multi-bias, single-task).
    }
\end{figure}
\clearpage

\begin{table}[h!]
    \centering
    \includegraphics[width=12cm]{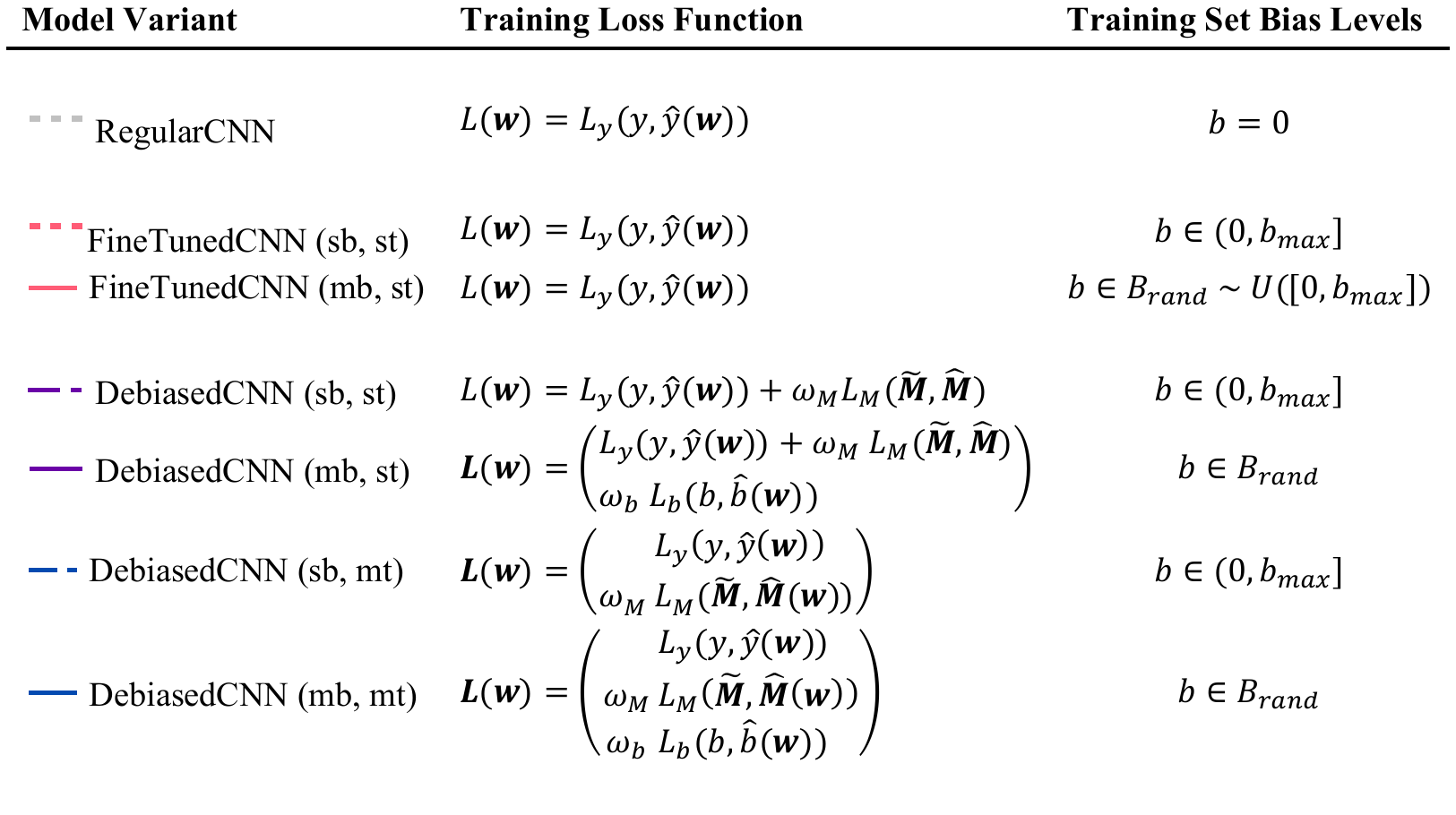}
    \caption{CNN model variants with single-task (st) or multi-task (mt) architectures trained on a specific (sb) or multiple (mb) bias levels. Each training set image $\bm{x}\in\bm{X}$ is preprocessed by a bias operator $\mathfrak{B}$ at a selected level $\mathrm{b}$, i.e., $\bm{x}_b=\mathfrak{B} \left(\bm{x},\left| b \right|>0\right)$, $\forall\bm{x}\in\bm{X}$. $\mathfrak{B}$ depends on the bias type (e.g., blur, color temperature, day-night lighting). For DebiasedCNN, mt refers to including a CAM task with differentiable CAM loss separate from the primary prediction task, while st refers to the primary prediction task with non-differentiable CAM loss. Models trained for single-bias (sb) used training set images biased at a single level $b>0$, while models trained for multi-bias levels (mb) used training datasets with data augmentation where each image is biased to a level that is randomly selected from a uniform probability distribution $B_{rand} \sim U([0,b_{max}])$. Multi-bias DebiasedCNN also adds a task for bias level prediction. Loss functions in vector form specify one loss function per task in a multi-task architecture.}
    \label{table:table-modelVariants}
    \Description{
    From left to right, model variants represented by different colored lines, their corresponding training loss and the bias level setting for training instances. 
    }
\end{table}

\clearpage

\subsection{Debiasing spurious explanations of privacy-preserving AI}
% \quad % placeholder to fix the layout issue
\begin{figure}[!ht]
    \centering    
    \includegraphics[width=7.6cm]{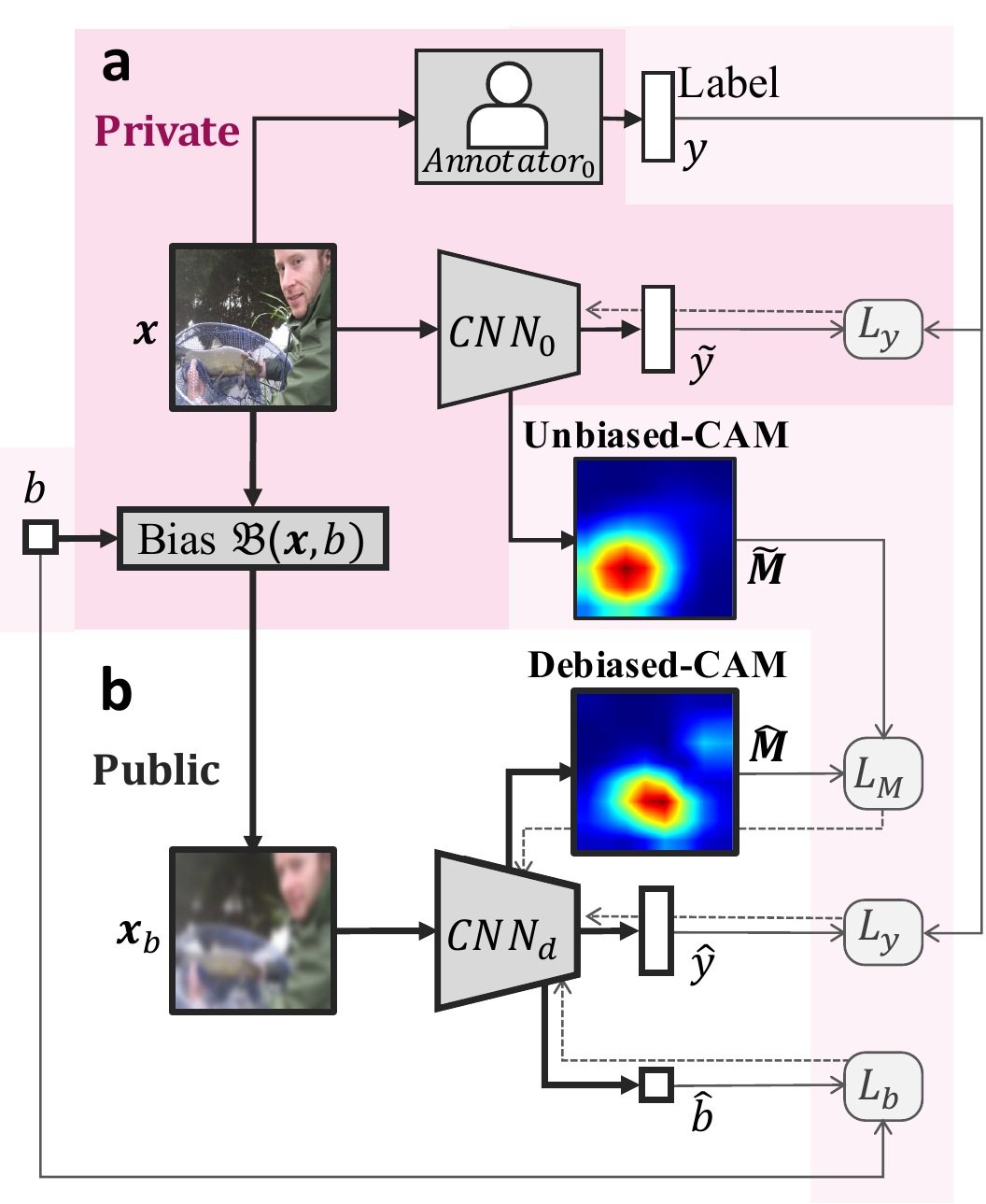}
    \caption{Architecture of multi-task DebiasedCNN model with self-supervised learning from private training data for privacy-preserving machine learning. a) RegularCNN ($CNN_0$) was trained on a private dataset with unblurred image $\pmb{x}$ to generate Unbiased-CAM $\widetilde{\pmb{M}}$. b) DebiasedCNN ($CNN_d$) was trained on the corresponding public (privacy-protected) biased form of the private dataset with blurred image $\pmb{x}_b$ and self-supervised with Unbiased-CAM $\widetilde{\pmb{M}}$ to generate Debiased-CAM $\widehat{\pmb{M}}$. During model training, $CNN_d$ has access to the bias level $b$ of each image $\pmb{x}_b$, Unbiased-CAM $\widetilde{\pmb{M}}$, and actual label $y$, but has no access to them during model inference. $CNN_d$ never has access to any unblurred image $\pmb{x}$. At inference time, DebiasedCNN can generate relevant and faithful Debiased-CAMs from privacy-protected blurred images. }
    \label{sfig:architecturePrivacy}
    \Description{ 
    Meta-architecture of privacy-preserving AI. The unbiased image (considered as private data) passes through the RegularCNN model ($CNN_0$) to get its prediction and Unbiased-CAM. The biased image (public data), which is obtained by applying bias transform on the same unbiased instance, passes through the DebiasedCNN model ($CNN_d$) to get its prediction, bias estimation, and Unbiased-CAM. During this process, Unbiased-CAM (public data) serve as the supervision signal for calibration. }
\end{figure}
\clearpage

\section{Simulation Studies Appendix}

\subsection{Supplemental Method: Calculating Bias Levels}
We provide details to calculate different bias for color temperature and lighting biases.

\subsubsection{Color Temperature Bias}
\label{subsubsection:color_bias}
Color temperature refers to the temperature of an ideal blackbody radiator as if illuminating the scene. 
We biased color temperature as follows. Each pixel in an unbiased image has color $\left(r,g,b\right)^T$, where $R,G,B$ represent the red, green, and blue color values within range 0-255, respectively. Each pixel is biased from neutral temperature $t_0$ by $\Delta t_b$ at bias level $b$ by multiplying a diagonal correction matrix with its color, i.e.,
\begin{equation}
   \left(r_b,g_b,b_b\right)^T=diag\left(255/R_b\ ,255/G_b,255/B_b\right)\left(r,g,b\right)^T,
\label{eq:color_mapping}
\end{equation}
where $\left(R_b,G_b,B_b\right)^T=f_{CT}(T)=f_{CT}(t_0+\Delta t_b)$ are scaling factors obtained from Charity’s color mapping function $f_{CT}$ to map a blackbody temperature to RGB values \cite{blackbody} (Supplementary Fig. \ref{sfig:colorMapping}). We set the neutral color temperature $t_0$ to 6600K, which represents cloudy/overcast daylight. Color temperature biasing is asymmetric about zero bias, because people are more sensitive to perceiving changes in orange than blue colors (Kruithof Curve \cite{davis1990correlated}); and due to the non-linear monotonic relationship between blackbody temperature and modal color frequency (Wien’s Displacement Law). This asymmetry explains why orange biasing led to stronger CAM deviation than blue biasing. 
% For the simulation studies, we varied color temperature bias $\Delta t$ to 7 levels –5400, –3600, –1800, 0, 1800, 3600, 5400 to evaluate Task Performance and CAM Faithfulness, and to random values between –5400 and 5400 to evaluation regression performance.

\setcounter{figure}{2}
\begin{figure}[!ht]
    \centering    
    \includegraphics[width=5.4cm]{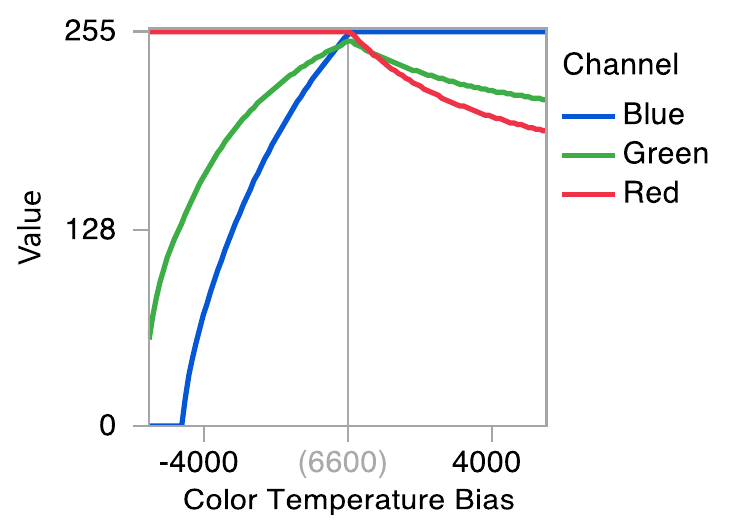}
    \caption{Color mapping function to bias color temperature of images in Simulation Study 4. Changes in Red, Green, Blue values are larger for orange biases (lower color temperature) than blue biases (higher temperature). Neutral color temperature is set to represent shaded/overcast skylight at 6600K.}
    \label{sfig:colorMapping}
    \Description{ 
    A line plot shows how RGB values vary with the color temperature (in the range of 1000K to 11200K).
    }
\end{figure}

\subsubsection{Lighting Bias}
\label{subsubsection:lighting_bias}
Lighting bias occurs when the same scene is lit brightly or dimly. In nature, this occurs as sunlight changes hour-to-hour, or season-to-season. 
The Transient Attributes database~\cite{laffont2014transient} contains photos of scenes from the same camera position taken across different times of the day and year. 
Attribute changes include whether the scene is daytime or nighttime, snowy, foggy, dusk/dawn or not.
We sought to generate images with different degrees of darkness, but the dataset only contained photos that were very bright or very dark. Therefore, we interpolated photos to generate scenes with intermediate darkness.
For each scene, with a daytime image $I_{day}(x, y)$ and nighttime image $I_{night}(x, y)$, we performed the pixel-wise interpolation as,
\begin{equation}
   I_{biased}(x, y) = (1 - \rho) \times I_{day}(x, y) + \rho \times I_{night}(x, y), 
\label{eq:color_mapping-1}
\end{equation} 
where $\rho$ is the night/day ratio. An unbiased image has $\rho = 0$ indicating daytime, and the most biased image has $\rho = 1$ indicating nighttime.

\clearpage

\subsection{Supplemental Results}

\quad % placeholder to fix the layout issue
\begin{figure}[ht]
    \centering    
    \includegraphics[width=17.2cm]{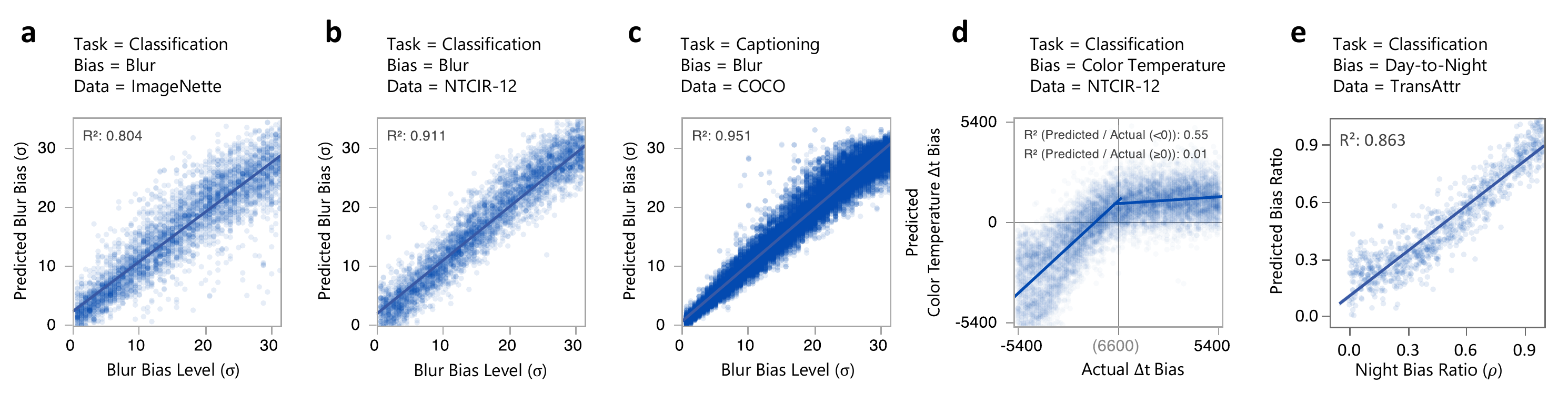}
    \caption{Regression performance for DebiasedCNN (mb, mt) measured as R2 for the bias level prediction task for five simulation studies. 
    % \textbf{a}, Simulation Study 1 (classification with blur-biased ImageNette),  \textbf{b}, Simulation Study 2 (classification with blur-biased NTCIR-12), \textbf{c}, Simulation Study 3 (captioning with blur-biased COCO), \textbf{d}, Simulation Study 4 (classification with color temperature-biased NTCIR-12). {\color{orange} \textbf{e}, SimulationStudy 5 (classification with night-vision biased Day-Night).} \textbf{a-c, e}, 
    Very high $R^2$ values indicate that models trained for Simulation Studies 1-3 and 5 could predict the respective bias levels well. 
    % \textbf{d}, 
    Color temperature bias level prediction depended on whether bias was towards lower (more orange) or higher (more blue) temperatures. Since blue-biased images were less distinguishable, the model was less well-trained to predict the blue color temperature bias level; it was more able to predict orange bias at a reasonable accuracy. }
    \label{sfig:simulationPlot_R2}
    \Description{
    Scatter plots of bias level regression result on different prediction tasks, bias types and datasets. 5 subfigrues share the same layout.
    a) the predicted bias level shows a strong positive correlation with actual bias level (R2 = 0.804) on blur biased ImageNette dataset.
    b) the predicted bias level shows a strong positive correlation with actual bias level (R2 = 0.804) on blur biased NTCIR dataset.
    c) the predicted bias level shows a strong positive correlation with actual bias level (R2 = 0.804) on blur biased COCO dataset.
    d) the predicted bias level shows a strong positive correlation with actual bias level (R2 = 0.804) on color temperature biased NTCIR dataset.
    e) the predicted bias level shows a strong positive correlation with actual bias level (R2 = 0.804) on day-night-conversion biased transAttr dataset.
    }
\end{figure}

\quad % placeholder to fix the layout issue
\begin{figure}[ht]
    \centering    
    \includegraphics[width=14.6cm]{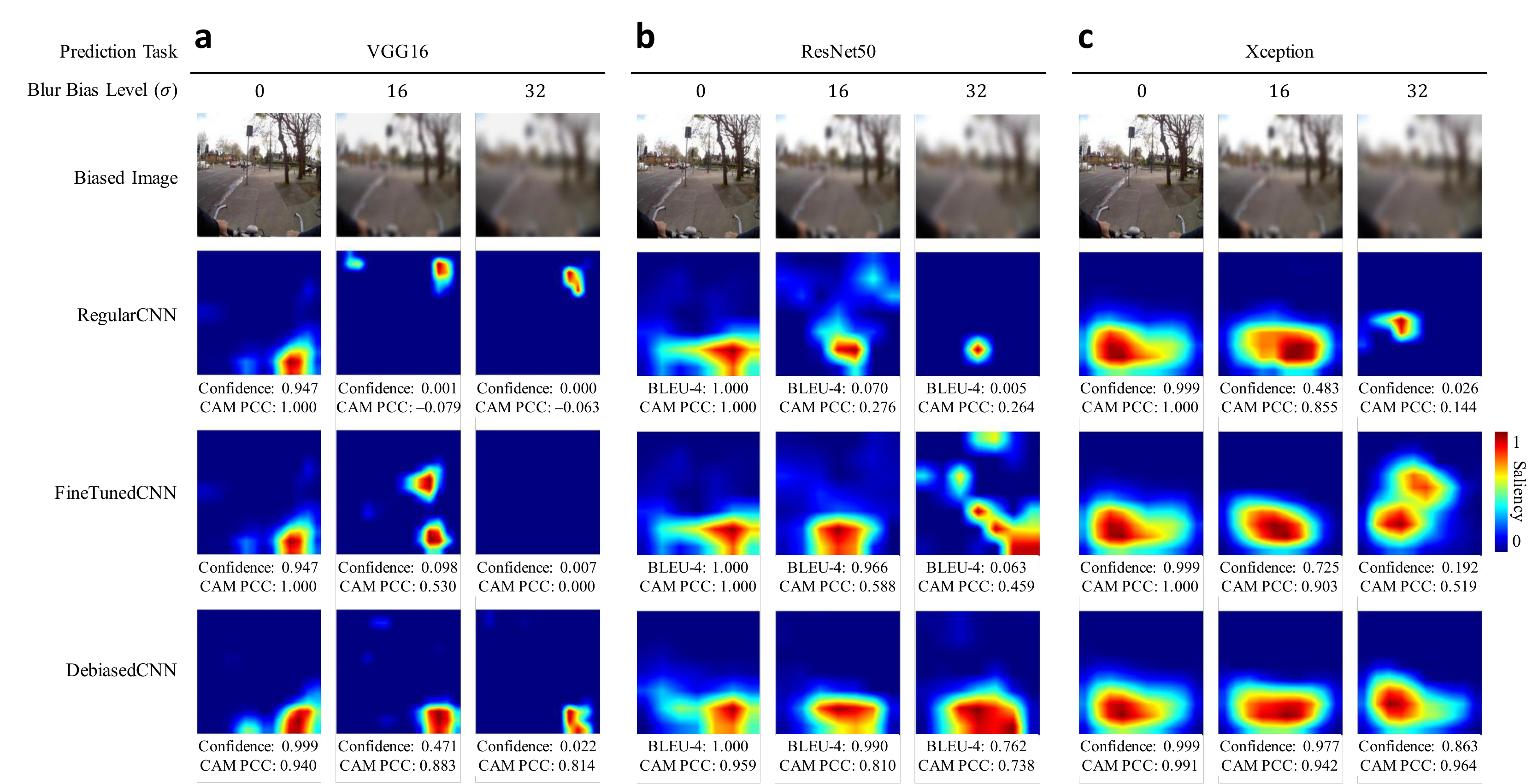}
    \caption{Deviated and debiased CAM explanations from various CNN models at varying bias levels of blur biased image from NTCIR-12 labeled as “Biking”. a) VGG16, b) ResNet50, c) Xception. a)-c), Models arranged in increasing CAM Faithfulness (see Supplementary Fig. \ref{sfig:simulationPlot_acrossBackbones}, second row). CAMs from more performant models were more representative of the image label with higher CAM Faithfulness (PCC). }
    \label{sfig:biasedCAMs_acrossBackbones}
    \Description{
    Demos of images and different types of GradCAMs on different backbone CNN models. Sub-figures a), b), c) share the same layout.
    a) The 1st row shows the same photo (an egocentric photo of biking activity) blurred with 3 different levels (sigma = 0, 16, 32) and GradCAMs from VGG16 model. The 2nd row shows the GradCAM visual explanations of RegularCNN on corresponding blurred images. The red region indicates a higher importance while the blue region indicates a lower importance. The text below the image indicates the confidence score of the image instance and the CAM truthfulness score (via PCC). The 3rd row and 4th row indicate the GradCAM visual explanations of FinetunedCNN and DebiasedCNN respectively.
    b) shows the same photo (an egocentric photo of biking activity) GradCAMs from ResNet50 model.
    c) shows the same photo (an egocentric photo of biking activity) GradCAMs from Xeception model.
    }
\end{figure}

\begin{figure}[ht]
    \centering    
    \includegraphics[width=14cm]{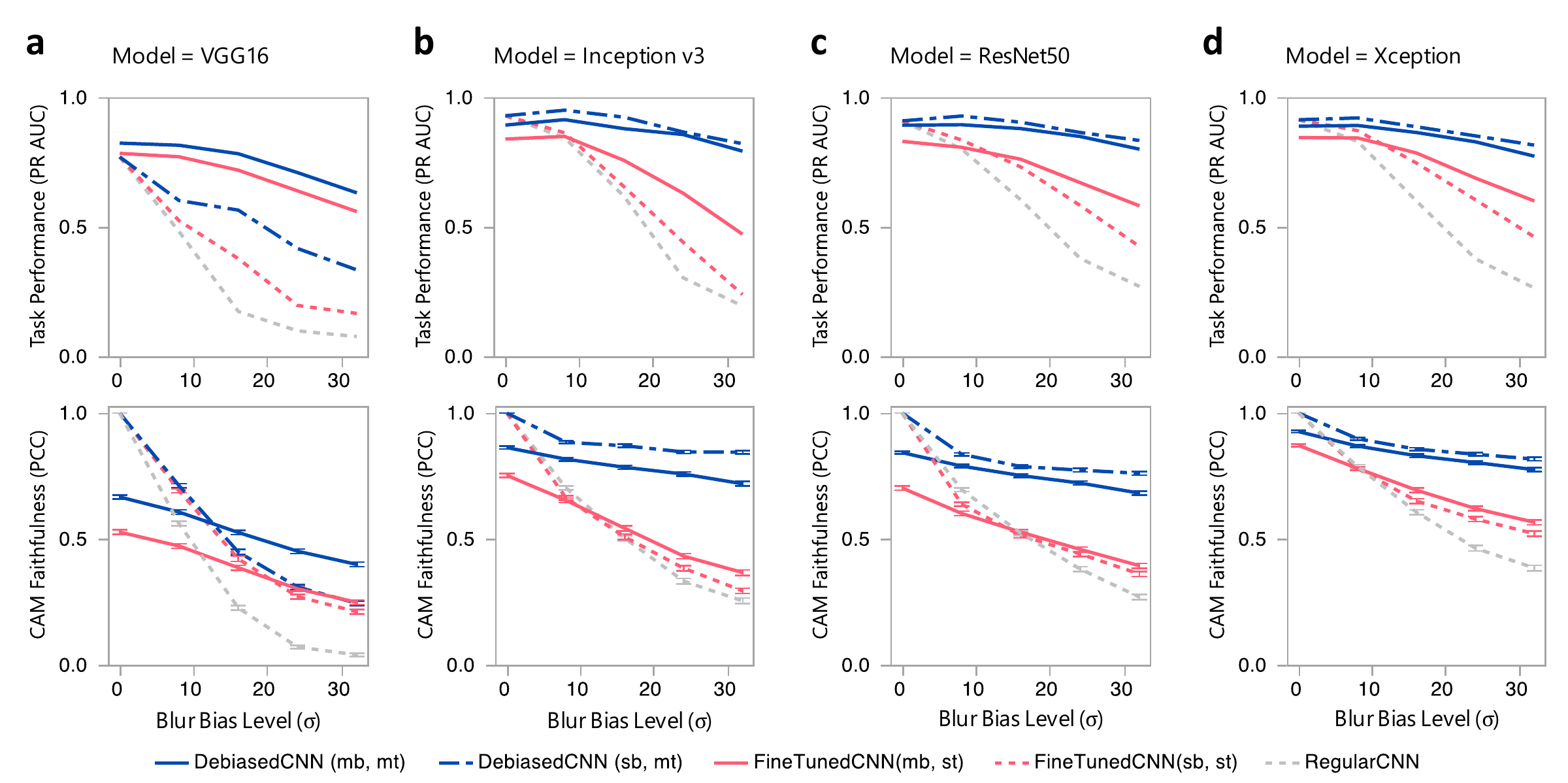}
    \caption{Comparison of model Task Performance and CAM Faithfulness for image classification on NTCIR-12 trained with different CNN models. a) VGG16, b) Inception v3, c) ResNet50, d) Xception. a)-d) Results agreed with Fig. \ref{fig:simulationPlots} that higher bias led to lower Task Performance and CAM Faithfulness, but debiasing improved both. CNN models are arranged in increasing CAM Faithfulness from left to right. All models were pretrained on ImageNet and fine-tune on NTCIR-12. We set the last two layers of VGG16, and last block of ResNet50 and Xception as retrainable. b)-d) Newer base CNN models than VGG16 significantly outperformed it for both Task Performance and CAM Faithfulness. These newer models had similar Task Performance across bias levels, though their CAM Faithfulness differed more notably.  
    }
    \label{sfig:simulationPlot_acrossBackbones}
    \Description{
    Barcharts of evaluation result on different backbone CNN models under blur bias on NTCIR12 dataset. 4 subfigrues share the same layout and the same trend.
    a) Two plots show the PRAUC (for classification performance) and PCC (for CAM faithfulness measurement) across different bias levels for various CAM conditions for VGG16 model. Generally, PRAUC and PCC decrease with a stronger blur bias level, while DebiasedCNN > FinetunedCAM > RegularCNN.
    b) Two plots show the PRAUC (for classification performance) and PCC (for CAM faithfulness measurement) for Inception v3 model.
    c) Two plots show the BLUE-4 (for image captioning performance) and PCC (for CAM faithfulness measurement) for ResNet50 model.
    d) Two plots show the PRAUC (for classification performance) and PCC (for CAM faithfulness measurement) for Xception model. 
    }
\end{figure}
\clearpage

\section{User Studies Appendix}

\subsection{User Studies Image Selection and CAMs}
\label{section:image_selection}
For both user studies, we chose 10 images to select one instance per class label for 10 classes of ImageNette. This balanced between selecting a variety of images for better external validity, and too much workload for participants due to too many trials. CAMs were generated from specific CNN models in Simulation Study 1. At each blur level, Unbiased-CAM and Biased-CAM were generated from RegularCNN, while Debiased-CAM was generated from DebiasedCNN (mb, mt). A key objective of the user studies was to validate the results of the simulation studies regarding CAM types and image blur bias levels, hence, we selected canonical images that:

\begin{enumerate}
\itemsep0em 
    \item Had RegularCNN and DebiasedCNN predict correct labels for unblurred images, since we were not investigating the use of CAMs to debug model errors. CNN predictions on blurred images may be wrong, but we showed the CAM of the correct label.
    \item Were easy to recognize when unblurred, so that users can perceive whether a CAM is representative of a recognizable image. This was validated in our pilot study.
    \item Were somewhat difficult but not impossible to recognize with Weak blur, so that participants can feasibly verify image labels with some help from CAMs.
    \item Were very difficult to recognize with Strong blur, such that about half of pilot participants were unable to recognize the scene, to investigate the upper limits of CAM helpfulness.
    \item Had Unbiased-CAMs that were representative of their labels, to evaluate perceptions with respect to truthful CAMs. Conversely, debiasing towards untruthful CAMs is futile. 
    \item Had Biased-CAMs for Strong blur that were perceptibly deviated and localized irrelevant objects or pixels; otherwise, no difference between Unbiased-CAM and Biased-CAM will lead to no perceived difference between Unbiased-CAM and Debiased-CAM too.
    \item Had Debiased-CAMs that were an approximate interpolation between the Unbiased-CAM and Biased-CAM of each image, to represent the intermediate CAM Faithfulness of Debiased-CAM found in the simulation studies.
\end{enumerate}

These criteria were verified with participants in a pilot study and the selected images had CAM Faithfulness representative of Simulation Study 1 for Debiased-CAM, but with slightly lower CAM Faithfulness for Biased-CAM to represent worse case scenarios. CAMs were %identical or 
different based on CAM type and Blur Bias level. Unbiased-CAMs were the same for all Blur Bias levels, and Unbiased-CAM and Biased-CAM were the same for None blur level. For other conditions, CAMs were deviated and debiased based on CAM type and Blur Bias level.
We chose not to test participants with images in NTCIR-12 due to quality and recognizability issues. Since images were automatically captured at regular time intervals, many images were transitional (e.g., pointing at ceiling while “Watching TV”), which made them unrepresentative of the label. Furthermore, in pilot testing, participants had great difficulty recognizing some scenes (e.g., “Cleaning and Chores”) in images with Strong blur, such that the tasks became too confusing to test. Nevertheless, our results can generalize to wearable camera images with Weak blur, for users who are familiar with or can remember their personal recent or likely activities.

\clearpage

\begin{figure}[H]
    \centering    
    \includegraphics[width=16cm]{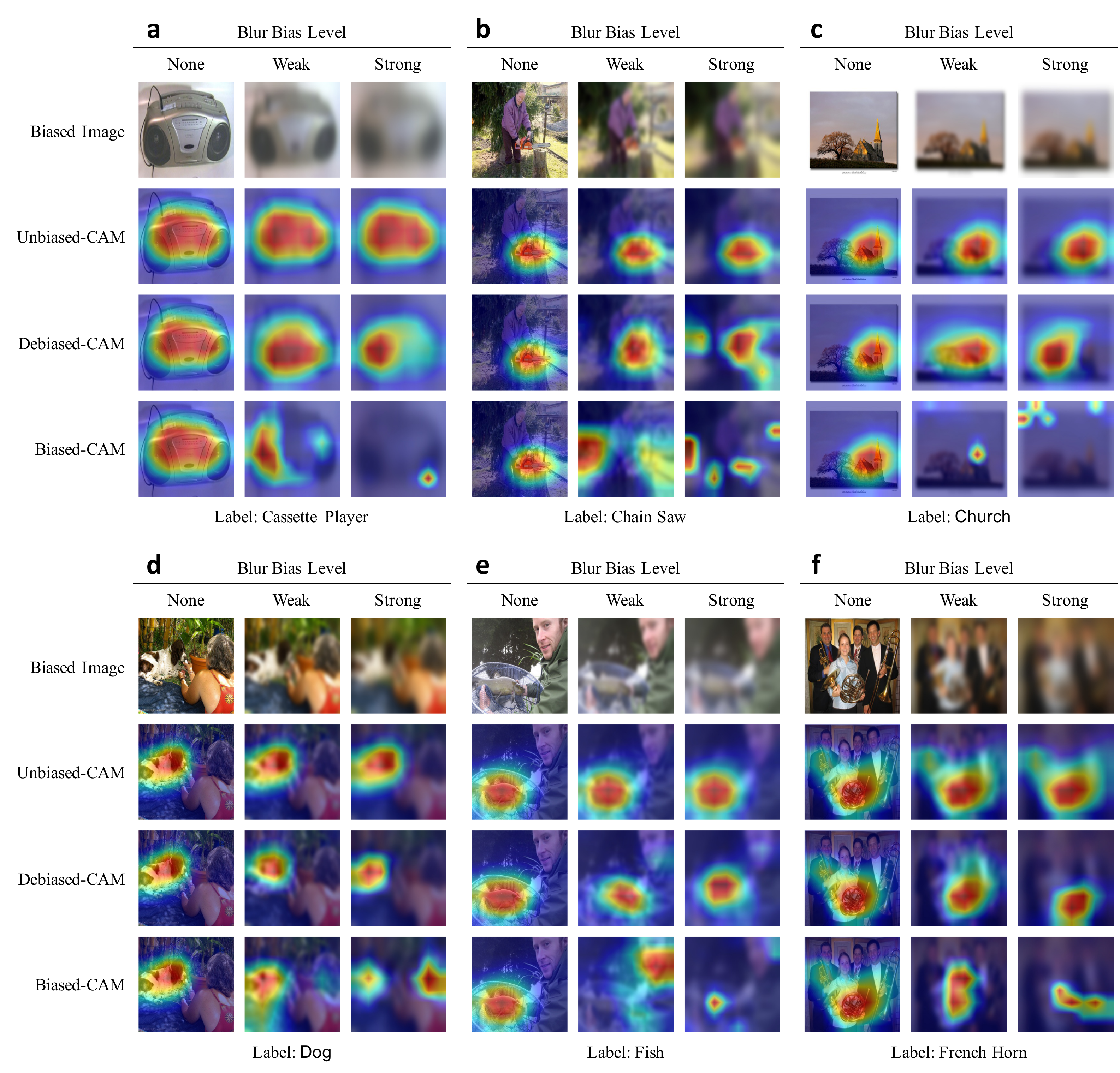}
    \caption{Images and CAMs at various Blur Bias levels and CAM types that participants viewed in both User Studies.} 
    \label{sfig:userStudyInstances1}
    \Description{
    Demos of images and different types of GradCAMs used in user studies. All sub-figures share the same layout.
    a) The 1st row shows the an image instance of Cassette Player blurred with 3 different levels. The 2nd row shows the unbiased-CAM visual explanations on corresponding blurred images. The 3rd row and 4th row indicate the visual explanations of Debiased-CAM and Biased-CAM respectively.
    b) shows the an image instance of Chain Saw and GradCAMs variants.
    c) shows the an image instance of Church and GradCAMs variants.
    d) shows the an image instance of Dog and GradCAMs variants.
    e) shows the an image instance of Fish and GradCAMs variants.
    f) shows the an image instance of French Horn and GradCAMs variants.
    }
    \vspace{-0.3cm}
\end{figure}

\begin{figure}[ht]
    \centering    
    \includegraphics[width=11cm]{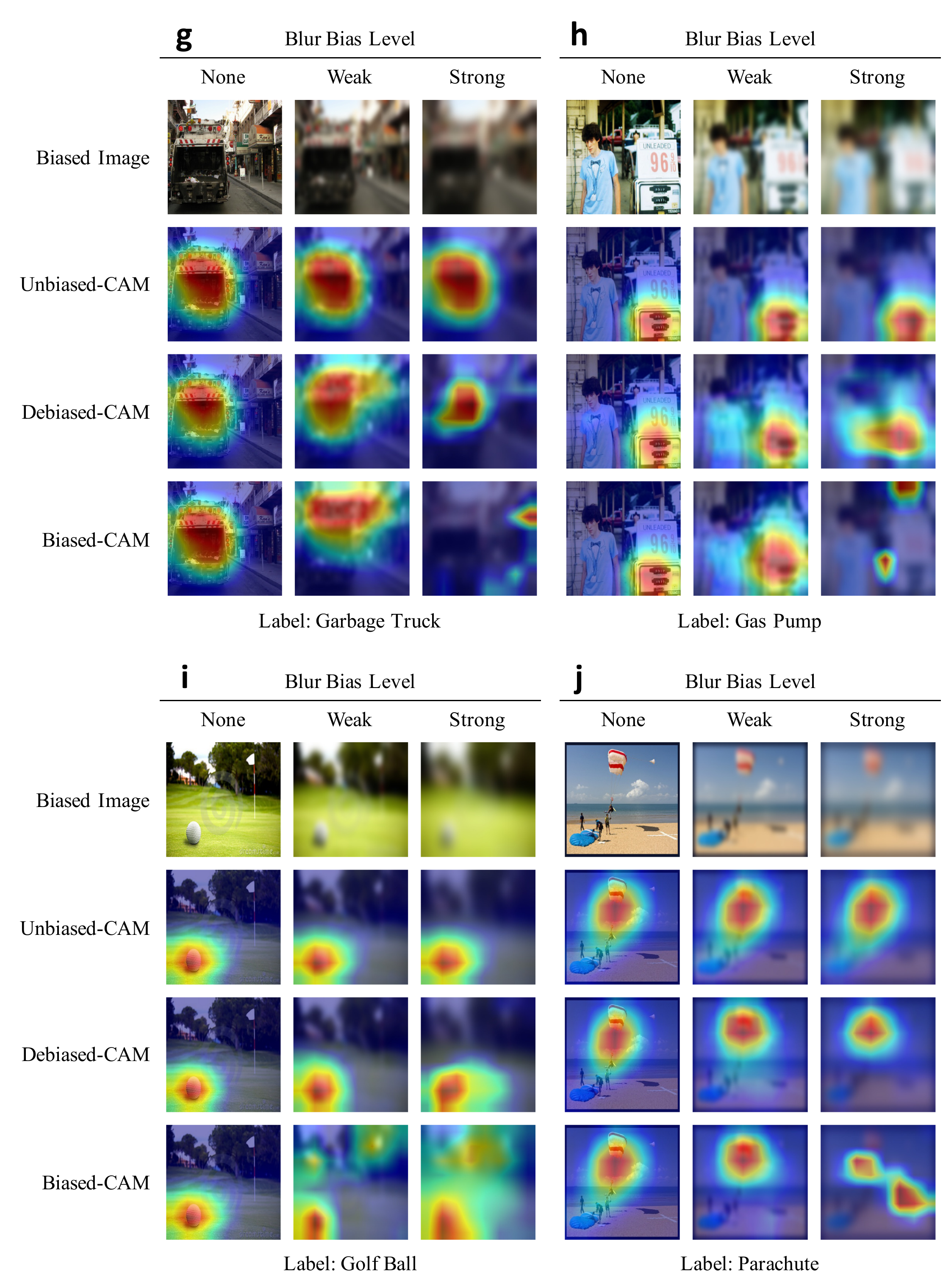}
    \caption{(Continued) Images and CAMs at various Blur Bias levels and CAM types that participants viewed in both User Studies.} 
    \label{sfig:userStudyInstances2}
    \Description{
    Demos of images and different types of GradCAMs used in user studies. All sub-figures share the same layout.
    g) The 1st row shows the an image instance of Garbage Truck blurred with 3 different levels. The 2nd row shows the unbiased-CAM visual explanations on corresponding blurred images. The 3rd row and 4th row indicate the visual explanations of Debiased-CAM and Biased-CAM respectively.
    h) shows the an image instance of Gas Pump and GradCAMs variants.
    i) shows the an image instance of Golf Ball and GradCAMs variants.
    j) shows the an image instance of Parachute and GradCAMs variants.
    }
    \vspace{1.0cm}
\end{figure}

\begin{figure}[h]
    \centering    
    \includegraphics[width=6.0cm]{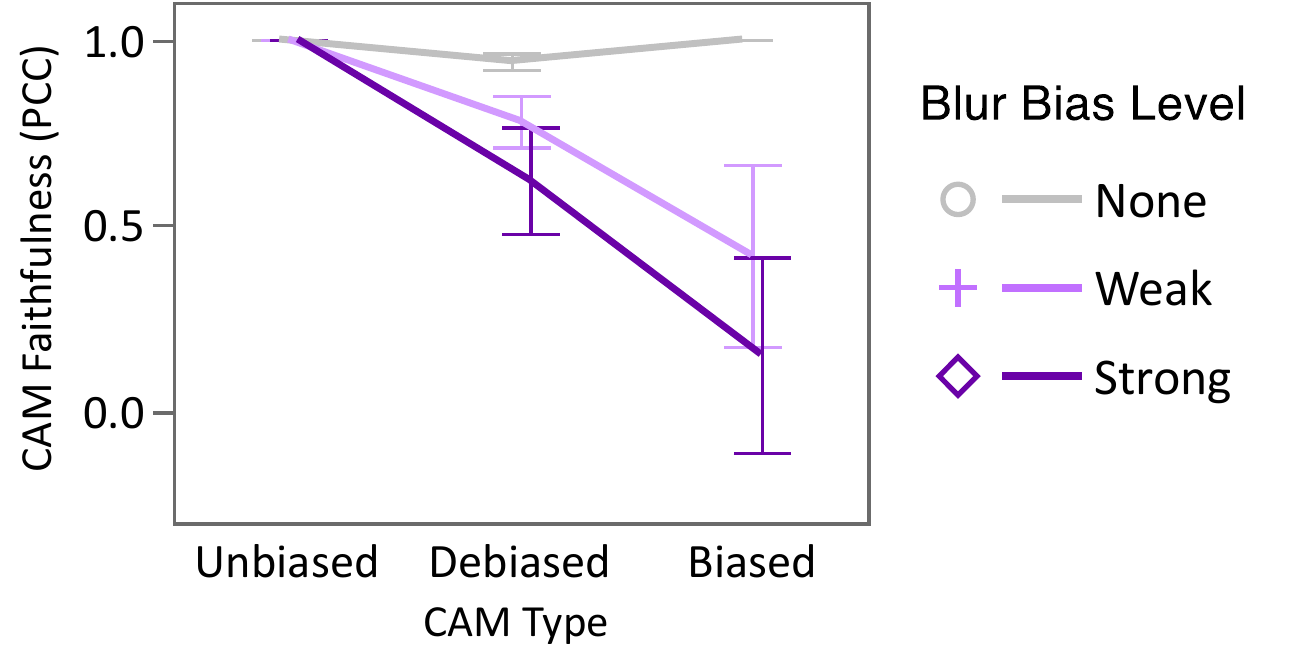}
    % \vspace{-0.3cm}
    \caption{CAM Faithfulness of selected 10 image instances used in user studies.
    Faithfulness decreased as Blur Bias increased, was the highest for Unbiased-CAM, the lowest for Biased-CAM, and improved by Debiased-CAM. Error bars indicate 90\% confidence interval.
    }
    \label{fig:InstanceFaithfulness}
    \Description{
    Barchart of CAM Faithfulness of instances used in user studies across 3 CAM conditions and 3 blur bias levels. Generally, None Blur > Weak Blur > Strong Blur, while Unbiased CAM > Debiased CAM > Biased CAM.
    }
    \vspace{-0.3cm}
\end{figure}

\clearpage

\subsection{User Study 1 and 2 Questionnaires}

We illustrate key sections in the questionnaire for the CAM Truthfulness User Study 1 and CAM Helpfulness User Study 2. Both questionnaires were identical except for the main study section.

\begin{figure}[ht]
    \centering    
    \includegraphics[width=18cm]{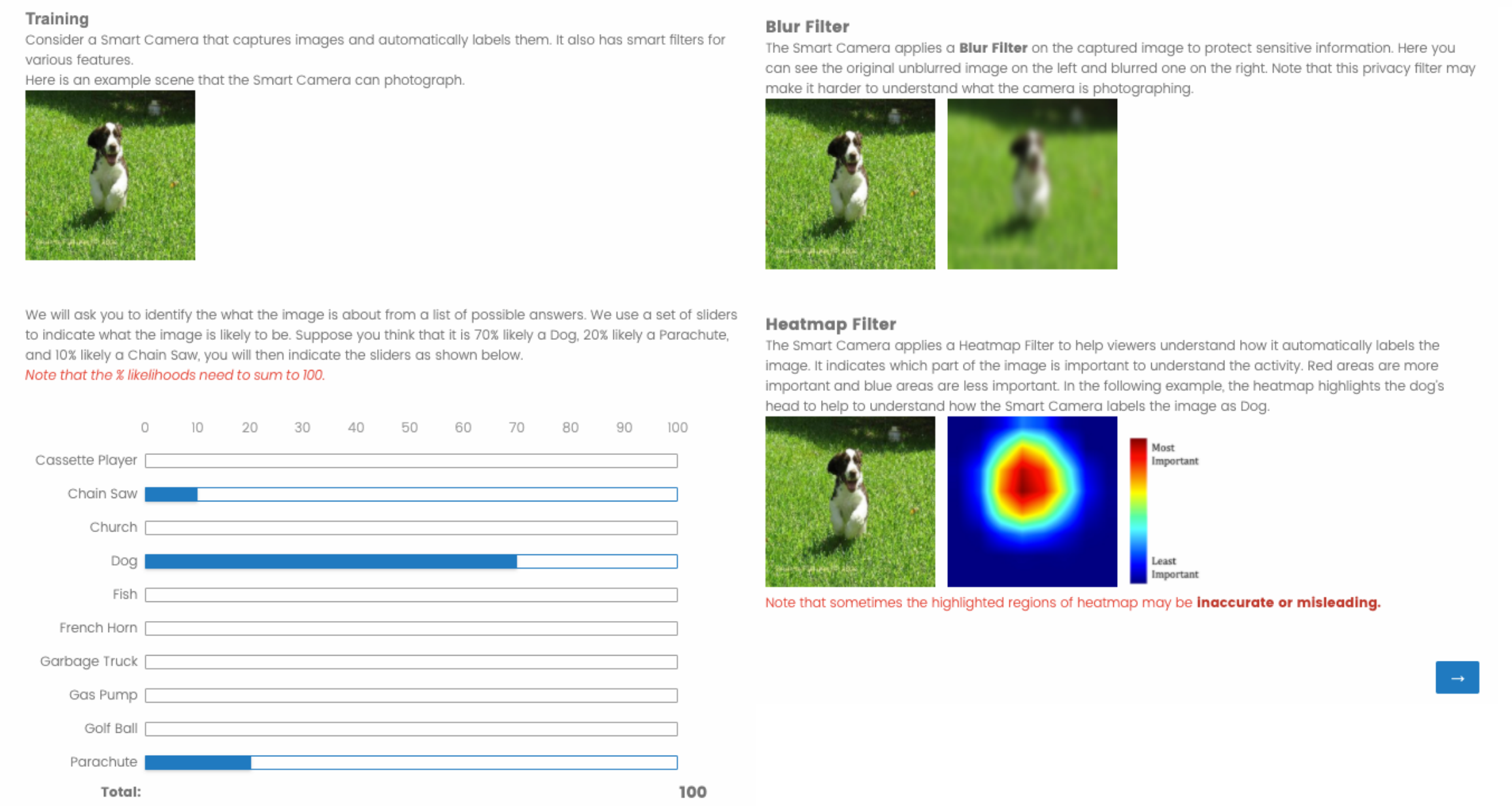}
    % \vspace{-0.2cm}
    \caption{Tutorial to introduce the scenario background of a smart camera with privacy blur and heatmap (CAM) explanation. It taught the participant to i) interpret the “balls and bins” question (\cite{goldstein2014lay}), ii) understand why images were blurred, and iii) interpret the CAM. }
    \label{sfig:userStudy1_tutorial}
    \Description{Tutorial page of our survey. The page consists of the task description, the instruction on the "balls and bins" question, the background about blur filter and heatmap filter.}
\end{figure}

\begin{figure}[ht]
    \centering    
    \includegraphics[width=18cm]{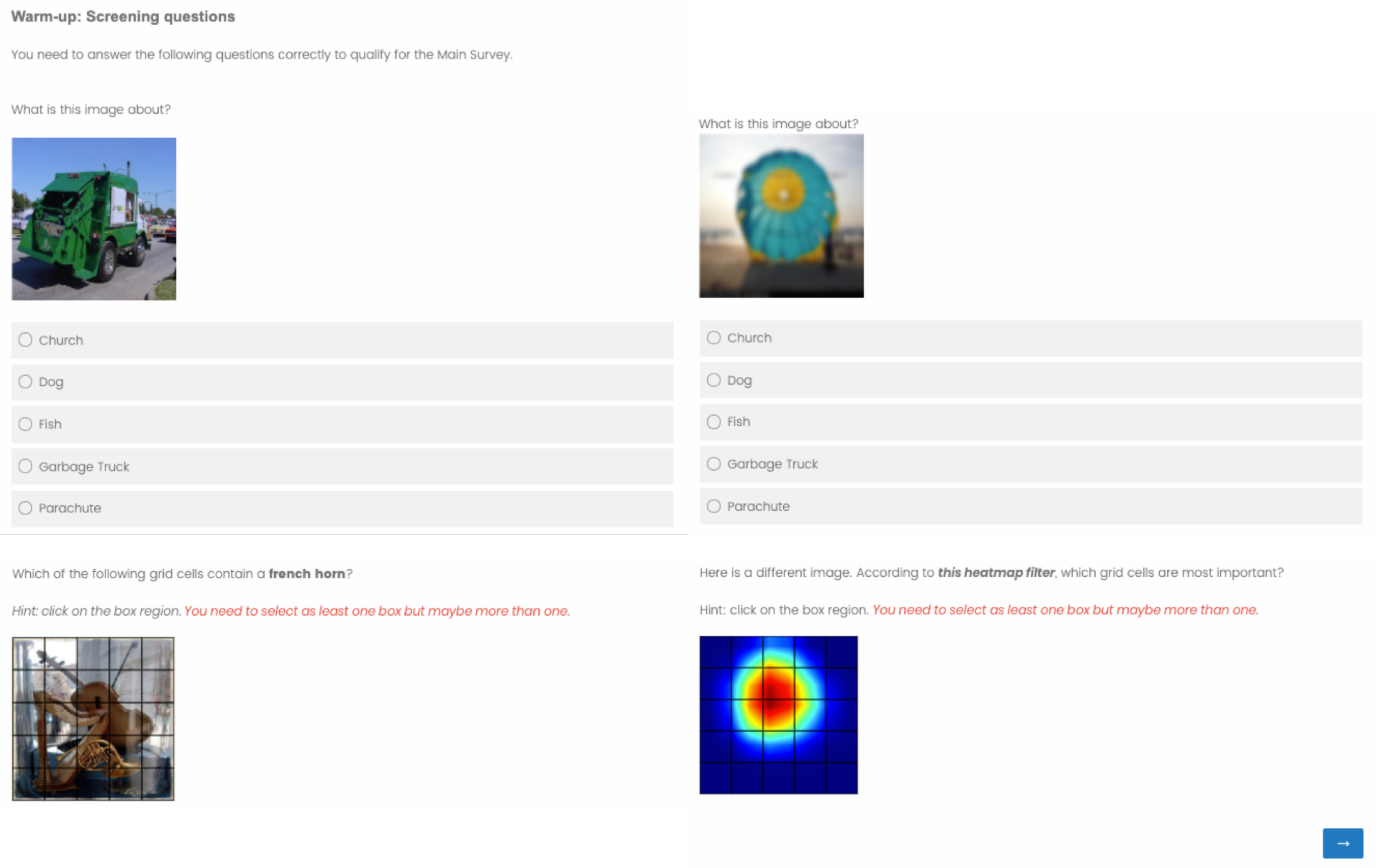}
    % \vspace{-0.2cm}
    \caption{Screening quiz with four questions to test labeling correctness and saliency selection. Questions tested for correct labeling on an unblurred (1) and a weakly blurred (2) photograph image, and correct grid selection of relevant locations in a photograph image (3) and a heatmap (4). The participant is excluded from the study if he answered more than one question wrongly. }
    \label{sfig:userStudy1_screening}
    \Description{Screening test of our survey. The page consists of 4 questions. 
    The 1st question shows an image of Garbage Truck and ask participants to select the correct label.
    The 2nd question shows an blurred image of Parachute and ask participants to select the correct label.
    The 3rd question shows an image of french horn and ask participants to select sub-regions related with french horn.
    The 4th question shows a heatmap and ask participants to select the most salient parts.
    }
\end{figure}
\clearpage

\begin{figure}[h]
    \centering    
    \includegraphics[width=9.5cm]{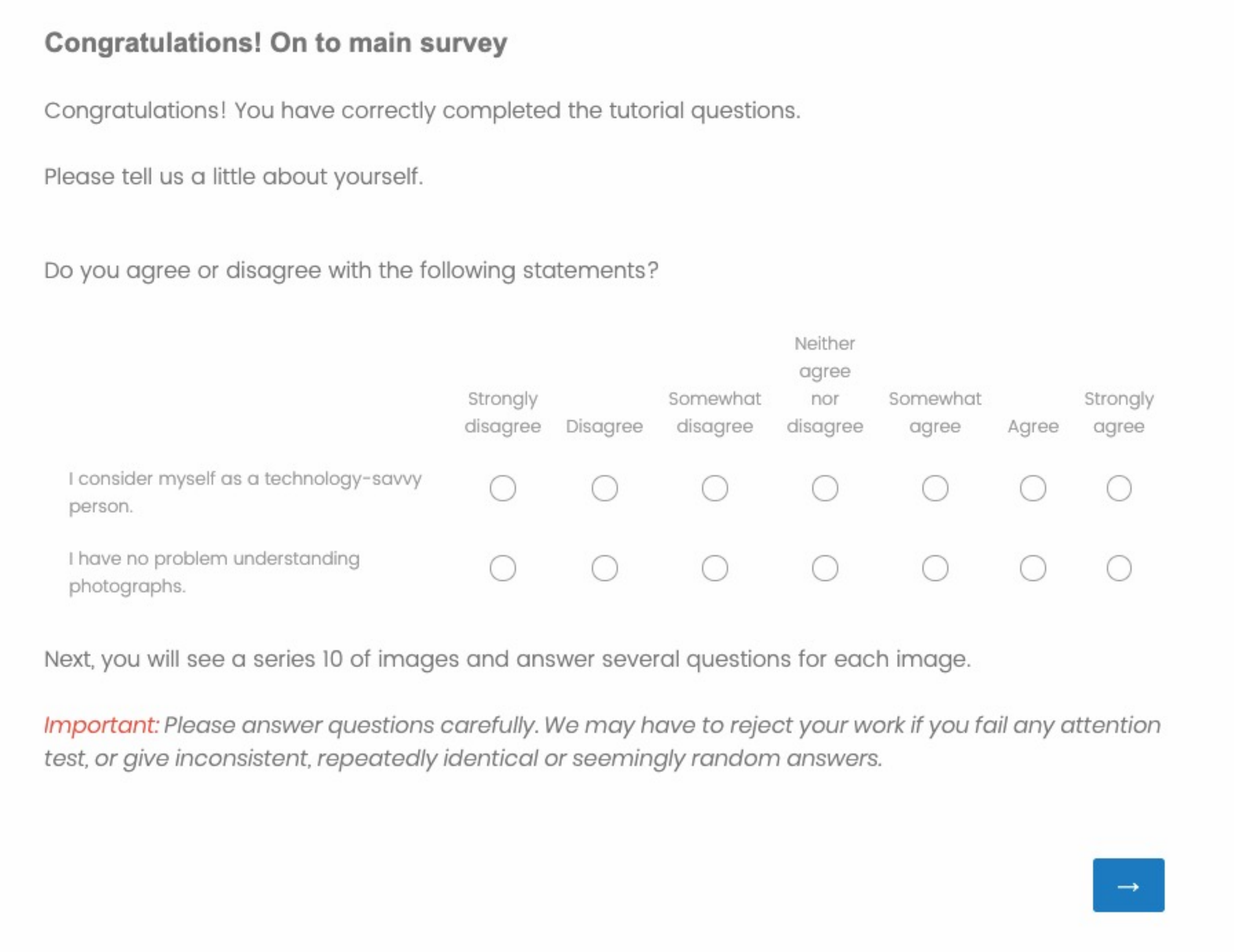}
    \caption{Background questions on participant self-reported technology savviness and photograph comprehension. These questions were posed after passing the screening quiz, and before the main study section to measure the participant’s pre-conceived self-assessment which may be biased after repeatedly viewing variously blurred images and variously biased heatmaps. }
    \label{sfig:userStudy1_background}
    \Description{A few message notifying participants that they successfully pass the screening test, followed by some background questions on self-reported technology savviness and photograph comprehension.
    }
    \vspace{-0.3cm}
\end{figure}
\clearpage

\begin{figure}[h]
    \centering    
    \includegraphics[width=18cm]{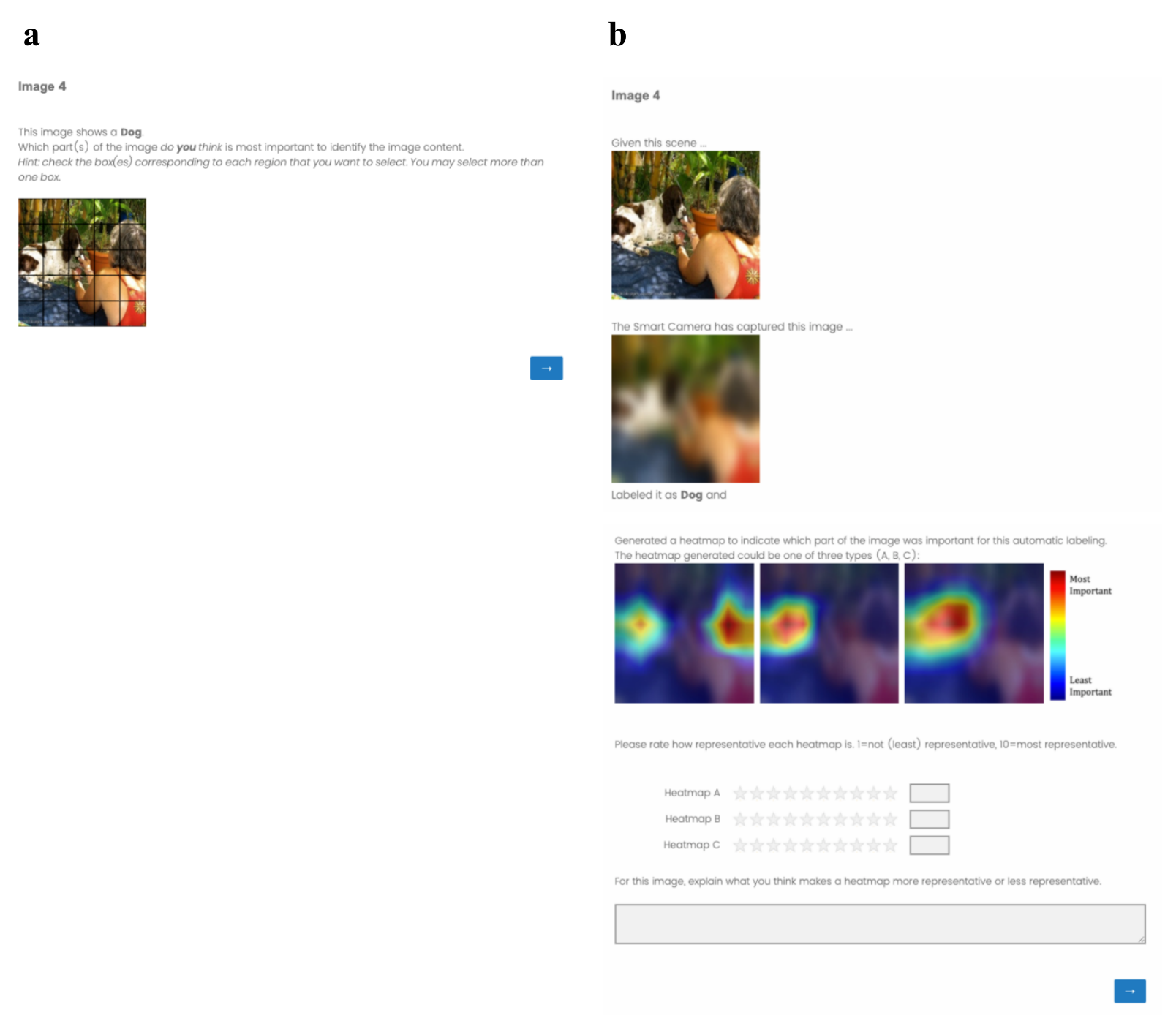}
    \caption{Example main study per-Image Trial for CAM Truthfulness User Study 1. a) The first page asked the participant to q1) select on a grid which locations in an unblurred image are important to identify the image as labeled. b) The second page showed how the smart camera has captured the image (at a randomly selected Blur Bias level), and asked the participant to q2) rate the Truthfulness of all three CAM types (randomly arranged) along a 10-point scale and to q3) explain her rating rationale. }
    \label{sfig:userStudy1_Trial}
    \vspace{-0.3cm}
    \Description{main study per-Image Trial for CAM Truthfulness User Study. In the first page, a grid selection task is shown where the participant need to select important grids to identify a certain concept. In the second page, the unblurred and blurred version of the image instance (a dog play with a man), and 3 different heatmaps are shown, followed by a rating question and a open ended text question.
    }
\end{figure}

\clearpage

\begin{figure}[h]
    \centering    
    \includegraphics[width=18cm]{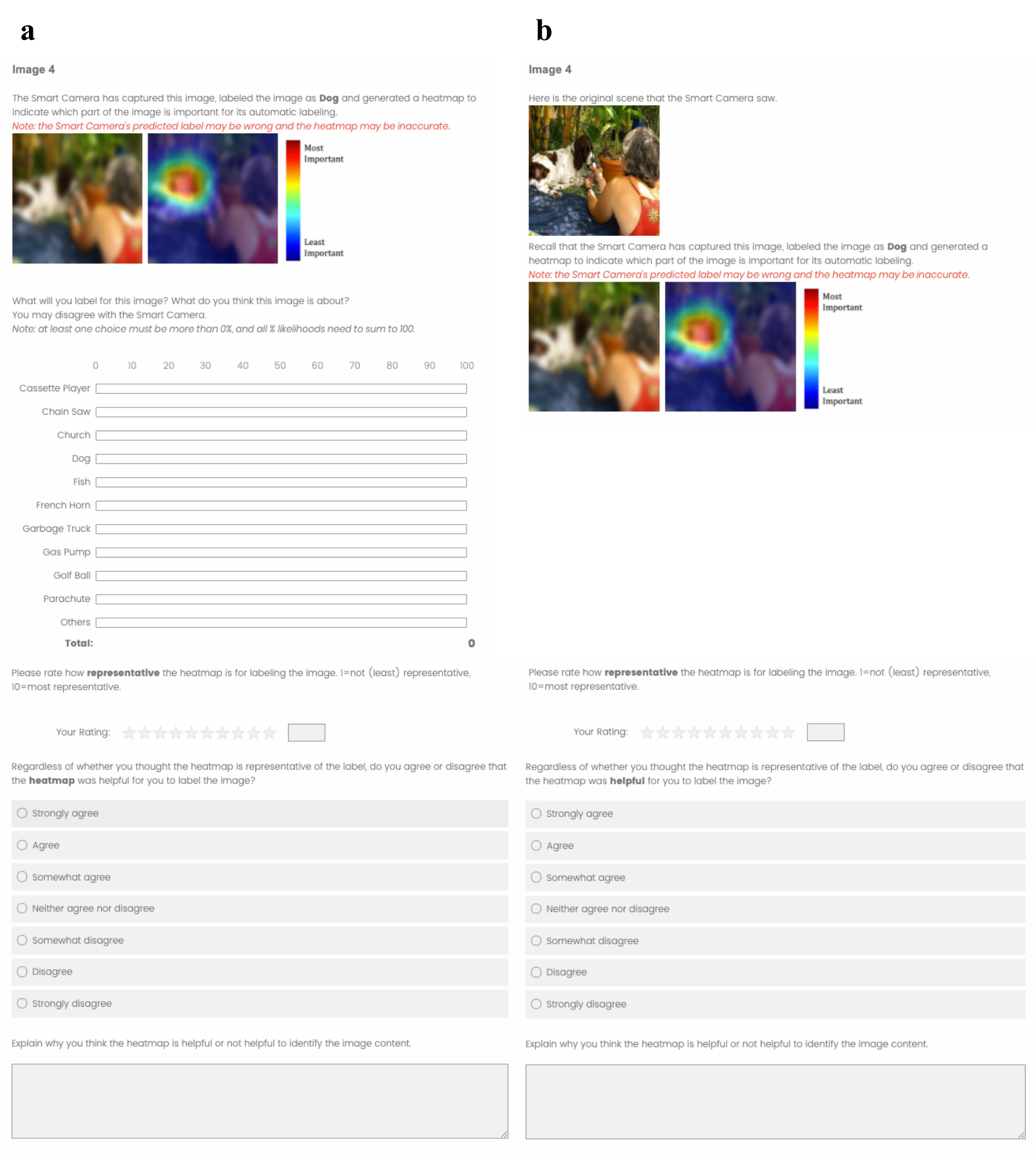}
    \caption{Example main study per-Image Trial for CAM Helpfulness User Study 2. a) The first page showed the smart camera’s captured blur biased image, generated heatmap (CAM) explanation, and predicted label; and asked the participant to q1) indicate the label likelihood with a “balls and bins” question; q2) rate the CAM Truthfulness, q3) rate the CAM Helpfulness and q4) explain his rating rationale. b) The second page showed the image unblurred, redisplayed the blurred image and CAM and repeated the questions for q5) CAM Truthfulness rating, q6) CAM Helpfulness rating and q7) rating rationale; the repeated questions allow the comparison of ratings before (preconceived) and after (consequent) the participant knew about the ground truth image. }
    \label{sfig:userStudy2_Trial}
    \Description{main study per-Image Trial for CAM Helpfulness User Study. In the first page, a blurred image and a heatmap are shown. The participant need to infer the label of the image in a balls and bins question, rate the helpfulness of the heatmap and describe the reason. 
    In the second page, in addition to the blurred image and heatmap, the unblurred version is revealed. The same group of questions were asked again.
    }
    \vspace{-0.3cm}
\end{figure}
\clearpage

% \clearpage

\subsection{Statistical Analyses}

\begin{table}[ht]
    \centering
    \includegraphics[width=14cm]{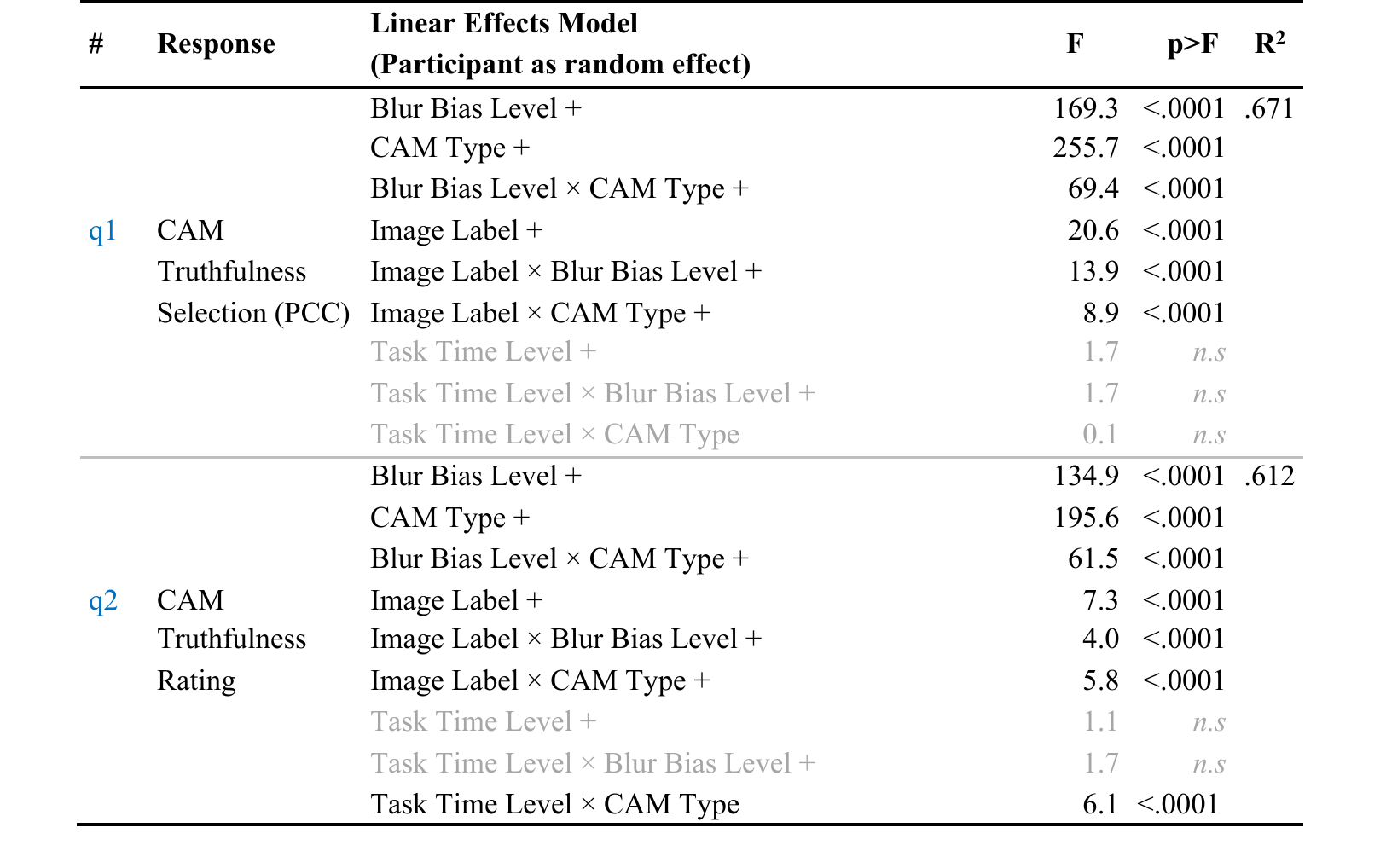}
    \caption{Statistical analysis of responses due to effects as linear mixed effects models for CAM Truthfulness User Study 1. 
    All models had various fixed main and interaction effects (shown as one effect per row) and Participant as a random effect. Rows with grey text indicate non-significant effects. Numbers (blue) correspond to survey questions in each image trial in Fig. \ref{fig:userStudy1Procedure}a.}
    \label{table:table-study1ModelDetails}
    \Description{
    Statistical analysis of responses for CAM Truthfulness user study. From left to right, the table shows the question index, the measured responses (including CAM  Truthfulness Selection (PCC) and CAM Truthfulness Rating), the effects in the linear effects model, the F value, the p value and R2.}
\end{table}

\begin{table}[ht]
    \centering
    \includegraphics[width=14cm]{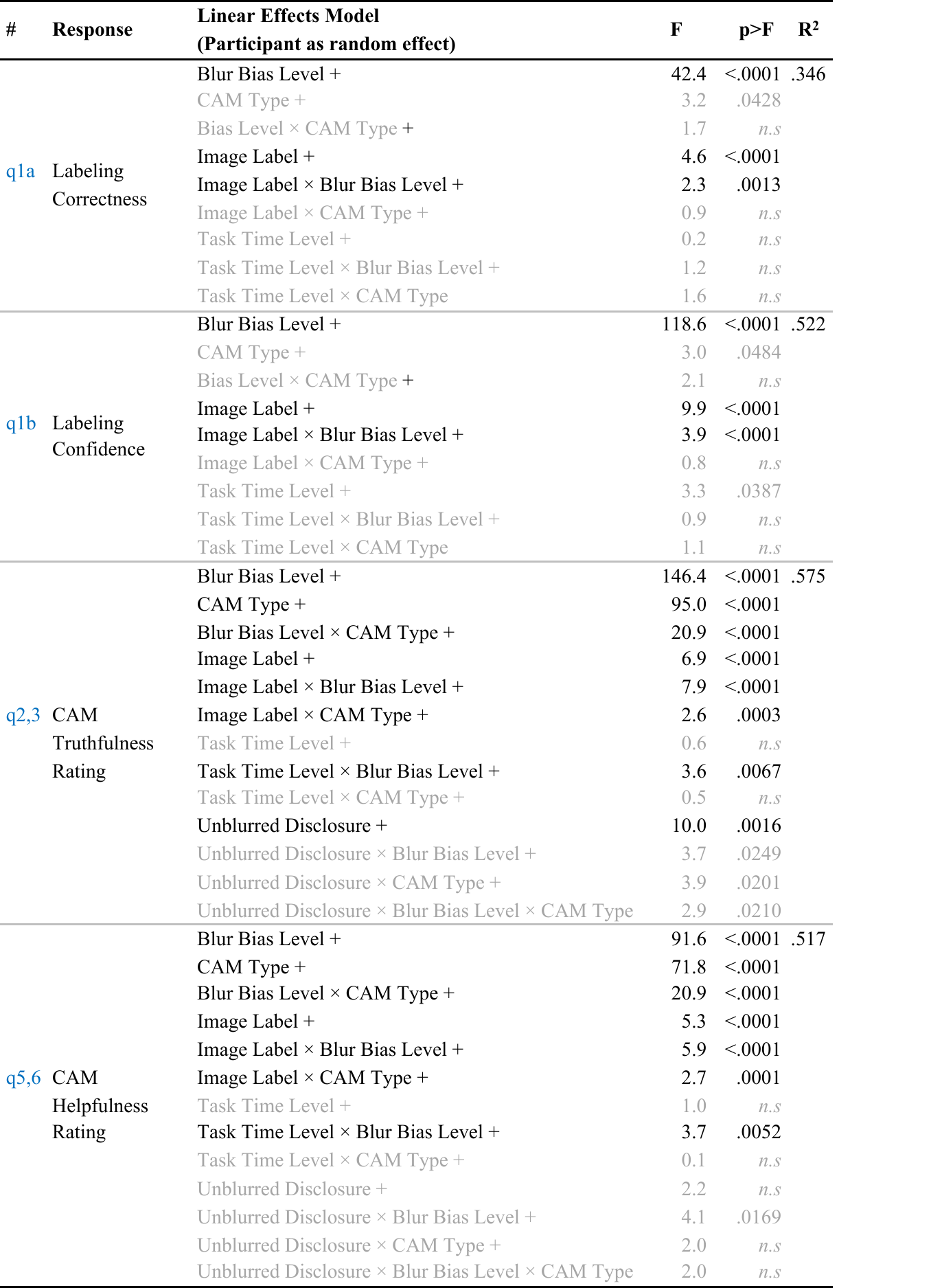}
%    \vspace{0.3cm}
    \caption{Statistical analysis of responses due to effects as linear mixed effects models for CAM Helpfulness User Study 2. 
    All models had various fixed main and interaction effects (shown as one effect per row) and Participant as a random effect. Rows with grey text indicate non-significant effects. Numbers (blue) correspond to survey questions in each image trial in Fig. \ref{fig:userStudy2Procedure}a.}
    \label{table:table-study2ModelDetails}
    \Description{
     Statistical analysis of responses for CAM Helpfulness user study. From left to right, the table shows the question index, the measured responses (including Labeling Correctness, Labeling Confidence, CAM Truthfulness Rating and CAM Helpfulness Rating), the effects in the linear effects model, the F value, the p value and R2.}
\end{table}

\clearpage

\subsection{Supplementary Results}

% \quad % placeholder to fix the layout issue
\begin{figure}[ht]
    \centering    
    \includegraphics[width=11cm]{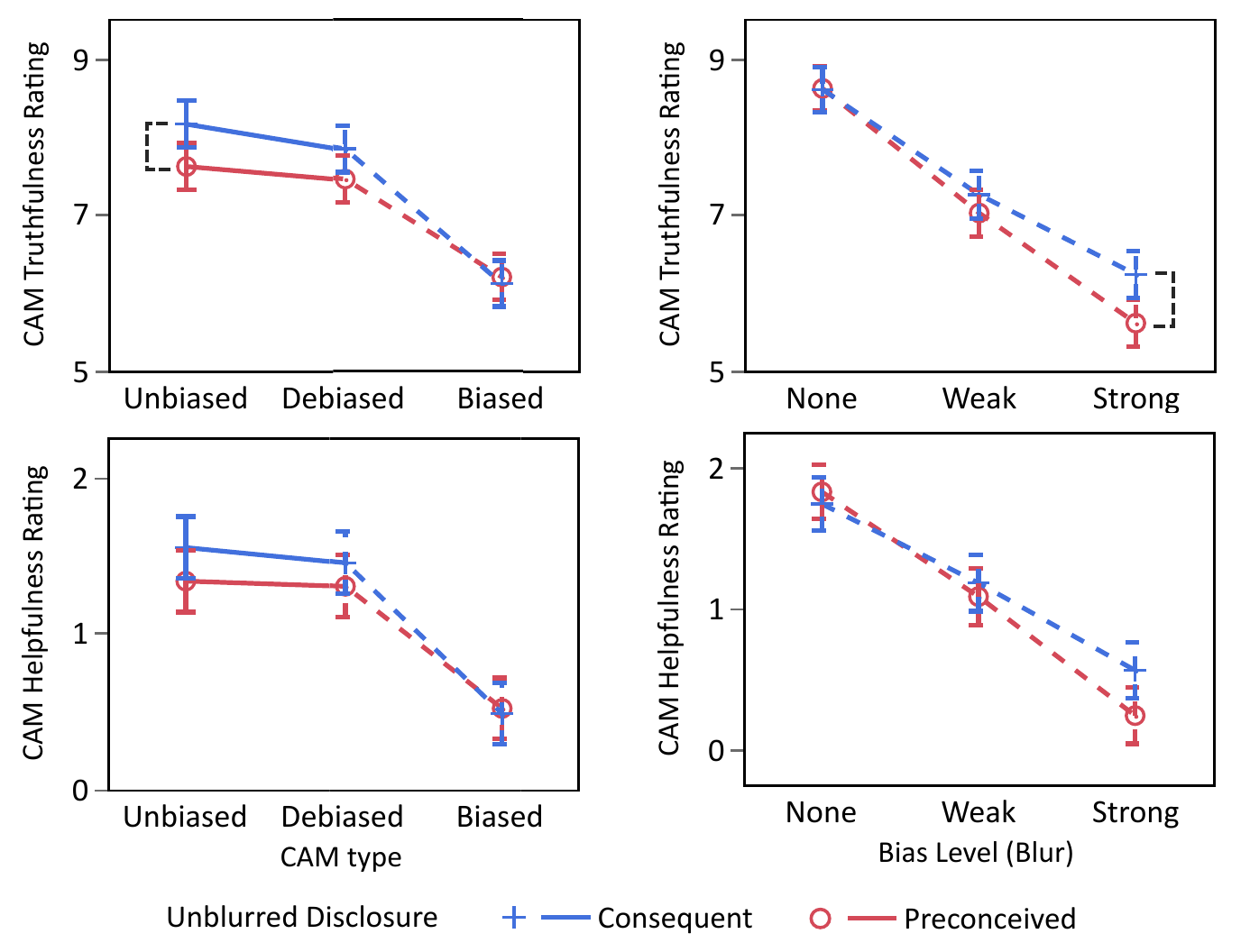}
    \caption{Comparisons of perceived CAM Truthfulness and CAM Helpfulness before (preconceived) and after (consequent) disclosing the unblurred image. There was a significant difference across Unblurred Disclosure for CAM Truthfulness Rating (p = .0013) but not for CAM Helpfulness Rating. 
    Comparing preconceptual to consequent ratings, Unbiased-CAMs were rated as less truthful (M = 7.7 vs. 8.3, p = .0004), Debiased-CAMs were rated marginally less truthful (p = .0212), Biased-CAMs were rated similarly untruthful, and overall, CAMs of Strongly blurred images were rated as less truthful (M = 5.6 vs. 6.3, $p<.0001$). These results suggest that even with the least biased CAM (Unbiased-CAM), the unfamiliarity of unblurred scenes can hurt trust (truthfulness) in the CAM, though there was no change in perceived helpfulness before or after disclosing the unblurred image. CAM Truthfulness Ratings were measured along a 1-10 scale, and CAM Helpfulness Ratings along a 7-point Likert scale (–3 = Strongly Disagree, 0 = Neither, +3 = Strongly Agree). Error bars indicate 90\% confidence interval. Dotted lines indicate extremely significant $p<.0001$ comparisons, and solid lines indicate no significance at $p>.01.$ }
    \label{sfig:userStudyPlots_preconceivedVsConsequent}
    \Description{
    Barcharts of the comparison on perceived ratings (including Truthfulness and Helpfulness) and consequent ratings. Perceived ratings show a similar trend with consequent ratings, where significance is found between DebaisedCAM and BiasedCAM for both truthfulness and helpfulness rating.
    }
\end{figure}

\end{document}